\documentclass[%
reprint,
jmp,
superscriptaddress,
 amsmath,amssymb,
 aps,
 pra,
]{revtex4-2}

\usepackage{graphicx}
\usepackage{dcolumn}
\usepackage{bm}
\usepackage[version=3]{mhchem}
\usepackage{svg}
\usepackage{graphicx} 
\usepackage{multirow}
\usepackage{comment}
\usepackage{tabularx}
\usepackage{xcolor}
\usepackage{array}
\usepackage{makecell}
\usepackage{amssymb}
\usepackage{float}
\usepackage{afterpage}
\usepackage{natbib}
\usepackage{braket}
\usepackage{bm}  
\usepackage{amsmath}
\usepackage{amsfonts}
\usepackage{hyperref}
\usepackage{booktabs} 
\usepackage{titlesec}
\usepackage{dsfont}
\usepackage[normalem]{ulem}

\titleformat*{\paragraph}{\normalfont\bfseries}

\renewcommand{\arraystretch}{1.2}

\begin{document}

\preprint{AIP/123-QED}
%

\author{Ilyes Batatia}
\thanks{These authors contributed equally - a detailed statement of author contributions is given at the end}
\affiliation{Engineering Laboratory, University of Cambridge, Cambridge, CB2 1PZ UK}
\affiliation{Department of Chemistry, ENS Paris-Saclay, Université Paris-Saclay, 91190 Gif-sur-Yvette, France}
\author{Simon Batzner}
\thanks{These authors contributed equally - a detailed statement of author contributions is given at the end}
\affiliation{John A. Paulson School of Engineering and Applied Sciences, Harvard University, Cambridge, MA 02138, USA}
\author{D\'{a}vid P\'{e}ter Kov\'{a}cs}
\affiliation{Engineering Laboratory, University of Cambridge, Cambridge, CB2 1PZ UK}
\author{Albert Musaelian}
\affiliation{John A. Paulson School of Engineering and Applied Sciences, Harvard University, Cambridge, MA 02138, USA}
\author{Gregor N.\ C.\ Simm}
\affiliation{Engineering Laboratory, University of Cambridge, Cambridge, CB2 1PZ UK}
\author{Ralf Drautz}
\affiliation{ICAMS, Ruhr-Universit\"at Bochum, Bochum, Germany}
\author{Christoph Ortner}
\affiliation{Department of Mathematics, University of British Columbia, Vancouver, BC, Canada V6T 1Z2}
\author{Boris Kozinsky}
\affiliation{John A. Paulson School of Engineering and Applied Sciences, Harvard University, Cambridge, MA 02138, USA}
\affiliation{Robert Bosch LLC Research and Technology Center, Cambridge, MA 02139, USA}
\author{G\'abor Cs\'anyi}
\affiliation{Engineering Laboratory, University of Cambridge, Cambridge, CB2 1PZ UK}



\title{The Design Space of E(3)-Equivariant Atom-Centered Interatomic Potentials}



\maketitle
\onecolumngrid

\section*{Abstract}
The rapid progress of machine learning interatomic potentials over the past couple of years produced a number of new architectures.
Particularly notable among these are the Atomic Cluster Expansion (ACE), which unified many of the earlier ideas around atom density-based descriptors, and Neural Equivariant Interatomic Potentials (NequIP), a message passing neural network with equivariant features that showed state of the art accuracy.
In this work, we construct a mathematical framework that unifies these models: ACE is extended and recast as one layer of a multi-layer architecture, while the linearised version of NequIP is understood as a particular sparsification of a much larger polynomial model.
Our framework also provides a practical tool for systematically probing different choices in the unified design space.
We demonstrate this by an ablation study of NequIP 
via a set of experiments looking at in- and out-of-domain accuracy
and smooth extrapolation very far from the training data, 
and shed some light on which design choices are critical for achieving high accuracy.
Finally, we present BOTNet (Body-Ordered-Tensor-Network), a much-simplified version of NequIP, 
which has an interpretable architecture and maintains accuracy on benchmark datasets.
\vspace{2.5em}

\twocolumngrid

\section{Introduction}

Over the past decade, there has been a revolution in atomistic modeling leading to the wide adoption of machine learning interatomic potentials, particularly for materials. There has been a wide range of different model architectures proposed in the literature.
The first class of models was built by constructing a descriptor (an array of numbers) to represent the environment of an atom.
The key to the success of these models was to make this descriptor invariant under the Euclidean symmetries of translation, rotation, and reflection, 
as well as under the permutation of atoms of the same element in the environment.
Two examples of such descriptors are the Atom Centered Symmetry Functions (ACSF)~\cite{Behler2007ACSF} and the Smooth Overlap of Atomic Positions (SOAP) \cite{Csanyi2013SOAP}. Many interatomic potentials have been built using these descriptors and subsequently used to model materials - see the corresponding recent review papers~\cite{Behler2021ChemRev,DeringerCsanyi2021ChemRev}.
More recently, it has been recognized that both of these methods can be understood as special cases of the Atomic Cluster Expansion (ACE)~\cite{ACE_ralf,DUSSON2022}.
The key idea of ACE was to introduce a complete set of basis functions (using spherical harmonics and an orthogonal radial basis) for the atomic environment that is built using the body-order expansion hierarchy.
Indeed, many previously proposed descriptors fit into the ACE framework, 
with the key difference being the maximum order of the body-order expansion (three-body for ACSF and SOAP, four-body for the bispectrum~\cite{GAP2010}, etc.) and specific choices of the radial basis functions~\cite{ACE_ralf, Ceriotti2021Representation}.
ACE naturally extends to equivariant features and to include variables beyond geometry, such as charges or magnetic moments.\cite{ACE_equivariant_ralf}
The Moment Tensor Potentials~\cite{MTP} construct a spanning set for the atomic environment using Cartesian tensors that can be expressed as a linear transformation of the ACE basis.
Given a descriptor, the atomic energy is fitted using a simple linear map~\cite{kovacs2021}, a Gaussian process~\cite{GAP2010}, or a feed-forward neural network~\cite{Behler2007ACSF}.
Other descriptor-based models are built for entire molecules or structures directly, rather than decomposed into atomic contributions~\cite{Tkatchenko2021ChemRev,Faber2018FCHL,Goedecker_OM_fingerprints}. 

In parallel to the development of models using descriptors of atom-centered environments, other groups explored the use of message passing neural networks (MPNNs) to fit interatomic potentials.
These models represent the atomic structure as a graph where an edge connects two nodes (atoms) if their distance is smaller than a fixed cutoff.
The models then apply a series of convolution or message passing operations on this graph to learn a representation of the environment of each atom.
This learned representation is mapped to the site energy via a readout function.
(A more detailed introduction to message passing potentials is given in Section~\ref{sec:MPNN}.)
Early models in this class, such as SchNet~\cite{schnet}, PhysNet~\cite{Physnet}, and DimeNet~\cite{DimeNet}, used internal features that are {\em invariant} under rotations of the input structure.

A key innovation of the Cormorant network~\cite{Cormorant}, Tensor-Field Networks~\cite{TensorField}, 
and steerable 3D-CNN-s~\cite{WeilerGatedNonLinearities2018} was to create \emph{equivariant} internal features that transform, under the symmetry operations of the input, like the irreducible representations of the symmetry group, and only construct invariants at the very last step. 
For example, features inside the network can rotate with the structure just like a Euclidean vector would.
To create these equivariant features inside the network, they introduced a new type of nonlinear operation, an equivariant tensor-product, which couples feature via the Clebsch-Gordan coefficients resulting in output features of a desired symmetry.
Several novel equivariant message-passing models were recently published (e.g. (NequIP)~\cite{nequip}, EGNN~\cite{EGNN}, PaiNN~\cite{Schutt2021Painn}, NewtonNet~\cite{haghighatlari2021newtonnet}, GemNet~\cite{klicpera2022gemnet}, TorchMD-Net~\cite{torchmd-net2022}, and SEGNN~\cite{Brandstetter2021Geometric}). The first to appear among these, NequIP (Neural Equivariant Interatomic Potential), improved on the state-of-the-art accuracy at the time by about a factor of two across multiple data sets. 
An alternative equivariant deep learning interatomic potential was also introduced recently that avoids the use of atom-centered message passing~\cite{Allegro2022}. 

In this paper, we set out a framework called Multi-ACE with the aim to unify the mathematical construction of message passing neural networks and ACE by using the latter in each layer of the network.
We set out a significantly expanded design space for creating machine learning-based interatomic potentials that incorporate many previous models.
Analogous work showing how message passing networks and atomic density-based descriptors can be formally unified was presented recently~\cite{nigam2022unified}.
The connection between MPNNs, ACE, and the expansion of the electronic structure as a power series in the Hamiltonian offers a complimentary view~\cite{Bochkarev_2022}. The connection between body order and MPNNs was first proposed in Ref.~\cite{Risi_N_bodyNet2018}.
Using the Multi-ACE framework, it is possible to probe different modeling choices systematically. We use this to understand the effect of some of the different parts of the design space via the example of NequIP and present a detailed study on what innovations and ``tricks'' of the NequIP model are essential for its remarkable performance.
These numerical experiments have led us to a new model called BOTNet, where we remove less essential design aspects of NequIP, particularly the non-equivariant nonlinear activations, while maintaining high accuracy across various datasets.

The paper is organized as follows: 
In Section~\ref{sec:MPNN}, we introduce the language and notation of MPNNs, the notion of equivariance, and body-ordering.
In Section~\ref{sec:EquiACE}, we review the Atomic Cluster Expansion and present its extension to equivariant properties.
In Section~\ref{sec:MultiACE_general}, we show how the equivariant Atomic Cluster Expansion can be extended to Multi-ACE, the unifying framework of equivariant interatomic potentials.
In Section~\ref{sec:BOTNet}, we briefly introduce the new BOTNet model and its code, which is a new implementation of equivariant interatomic potentials that, 
thanks to its modular design, is well suited for quick experimentation within this framework.
In Section~\ref{sec:Datset}, we describe the datasets used for the study on the components of NequIP and BOTNet. Although we focus on molecular benchmarks here, the models apply to any atomistic system.
In Section~\ref{sec:DesignSpace}, we present a detailed case study using the Multi-ACE framework to explore the design choices of NequIP and BOTNet and examine in detail the importance of each of the components of the models.
In Section~\ref{sec:normalization}, we investigate the importance of normalization.
We discuss the internal normalization of the features and the normalization of the datasets used to parametrize them.
In Section~\ref{sec:Benchmark}, we put the previous results into a broader context showing the excellent accuracy of NequIP and BOTNet on some standard datasets and comparing them to previous approaches.

\section{Message Passing Neural Network Potentials}
\label{sec:MPNN}

In this section, we summarize the message passing neural network framework \cite{GNNBattagial2018,MPNNGilmer2017,bronstein2021geometric} for fitting interatomic potentials.
Later, we will use this framework to elucidate the connections between linear Atomic Cluster Expansion and MPNNs.
The comparison of a wide range of models within this framework helps identify and explain their key similarities and differences.

MPNNs are a class of graph neural networks that can parametrize a mapping from the space of labeled graphs to a vector space of features. They can be used to
parametrize interatomic potentials by making atoms correspond to the nodes of the graph, and an edge connects two nodes if their distance is less than a specified cutoff distance, $r_\text{cut}$. The model maps a set of atoms with element types positioned in the three-dimensional Euclidean space to the total potential energy.
Typically, $r_\text{cut}$ is several times larger than the length of a covalent bond.
Thus, the corresponding graph is quite different from the typically drawn bonding graph of a molecule;
instead, it represents the spatial relationships between atoms on a somewhat larger length scale. We denote the set of neighbors of an atom $i$, i.e., atoms with the cutoff distance, by $\mathcal{N}(i)$.

\subsection{Semi-local states}
\label{sec:semi_loc_states}

We denote the state of an atom $i$ by the tuple $\sigma_{i}^{(t)}$:
\begin{equation}
    \label{eqn:state}
    \sigma_{i}^{(t)} = (\bm{r}_{i},\bm{\theta}_{i},\bm{h}_{i}^{(t)}),
\end{equation}
where
$\bm{r}_i$ denotes the atom's Cartesian position vector,
$\bm{\theta}_i$ a set of its fixed attributes such as the chemical element
(typically represented by a one-hot embedding),
and $\bm{h}_i^{(t)}$ its learnable features.
These features, unlike the attributes, are updated after each message passing iteration, denoted by $t$, based on the states of the atoms connected to atom $i$.
We refer to the states as \emph{semi-local} as the features will ultimately depend on states of atoms far away (around 10 to 40~\AA, depending on the local neighbourhood cutoff and the number of iterations).
A smooth cutoff mechanism is employed such that the updates are continuous when atoms leave or enter each other's local neighborhood.

\subsection{Message passing formalism}
\label{sec:mp_formalism}

We reformulate the original MPNNs equations \cite{MPNNGilmer2017} for atomic states.
In general, an MPNN potential consists of a message passing and a readout phase.
In the message passing phase,
atomic features, $\bm{h}_i^{(t)}$, are updated based on an aggregated \textit{message}, $\bm{m}^{(t)}_i$,
derived from the states of the neighboring atoms, within the set $\mathcal{N}(i)$:
\begin{equation}
    \label{eqn:message_func}
    \bm{m}_i^{(t)} = \frac{1}{\lambda}\bigoplus_{j \in \mathcal{N}(i)} M_t(\sigma_i^{(t)}, \sigma_j^{(t)}),
\end{equation}
where $\bigoplus_{j \in \mathcal{N}(i)}$ refers to a permutation invariant pooling operation over the neighbors of atom $i$
and $\lambda$ is a normalization constant that can correspond to, for example, the average number of neighbors across the training set
(see Section~\ref{sec:internal_normalization} for a detailed analysis of the role of this normalization).
$M_t$ denotes a learnable function acting on the states of atoms $i$ and $j$.
The most widely used permutation invariant pooling operation is the sum over the neighbours.
This operation creates messages, $\bm{m}_i^{(t)}$, that are 2-body in nature, i.e., linear combinations of functions that simultaneously depend on the features of only two atoms.
Then, the message $\bm{m}_i^{(t)}$ may be combined with the features of atom $i$ by a learnable update function, $U_t$:
\begin{equation}
    \label{eqn:update_function}
    \sigma_i^{(t + 1)}  \equiv  (\bm{r}_{i},\bm{\theta}_{i},\bm{h}_{i}^{(t+1)}) = (\bm{r}_{i},\bm{\theta}_{i}, U_t(\sigma_i^{(t)} , \bm{m}_i^{(t)})).
\end{equation}
In $U_t$, it is possible to form higher body-order messages by, for example, applying a square function to the message
to obtain a linear combination of 3-body functions that simultaneously depend on the central atom and \emph{two} of its neighbors.
Both the message function, $M_t$, and the update function, $U_t$, depend on the iteration index $t$.

In the readout phase, learnable functions $\mathcal{R}_{t}$ map the atomic states onto atomic site energies
\begin{equation}
    E_{i} =  \sum_{t}\mathcal{R}_{t}( \sigma_i^{(t)}).
\end{equation}
Here, some models use the atomic states from every iteration, while others use only a single readout function that takes the state after the final iteration and maps it to the site energy.

\subsection{Equivariant messages}
\label{sec:equivariant_message}

Physical properties, such as the energy or the dipole moment, transform in specific ways under the action of certain symmetry operations, such as translations and rotations of the atomic coordinates.
For atomistic modeling, the symmetry group of Euclidean symmetries (translations, rotations, and reflections) of three-dimensional space, $\text{E}(3)$, is of special interest.
For example, if a molecule is rotated in space, the predicted dipole moment should rotate accordingly, whereas the energy should remain unchanged.
Here, we restrict ourselves to rotational and reflectional symmetries, the $\text{O}(3)$ group,
as translation invariance can be ensured by working with interatomic displacement vectors, $\bm{r}_{ji} := \bm{r}_{j} - \bm{r}_{i}$.

A natural and convenient way to ensure that the outputs of models transform correctly is to impose constraints on the internal representations of the model to respect these symmetries.
We categorize the features of an equivariant neural network based on how they transform under the symmetry operations of the inputs.
Formally we can think of a message $\bm{m}^{(t)}_{i,L}$ as a function of the input positions ${\bm r}_{i}$, here suppressing, for the sake of brevity, the dependence on attributes ${\bm \theta}_i$.
Then we say that $\bm{m}^{(t)}_{i,L}$
is rotationally equivariant (with symmetry label $L$) if it transforms according to the irreducible representation $L$ of the symmetry group,
\begin{equation}
    \label{eq:equivariant_const}
    \begin{split}
        \bm{m}^{(t)}_{i,L}\left(Q\cdot({\bm r}_{1},...,{\bm r}_{N})\right) = \bm{D}^{L}(Q)\bm{m}^{(t)}_{i,L}({\bm r}_{1},...,{\bm r}_{N}),& \\
        \forall Q \in \text{O}(3)&
    \end{split}
\end{equation}
where $Q\cdot(\bm{r}_{1},...,\bm{r}_{N})$ denotes the action of an arbitrary rotation on the set of atomic positions $(\bm{r}_{1},...,\bm{r}_{N})$
and $\bm{D}^{L}(Q)$ is the corresponding Wigner D-matrix.
Hence, a message indexed by $L$ transforms like the spherical harmonic $Y_L^M$ under rotation.

An important practical choice for implementing equivariant neural networks is the basis in which features and messages are expressed.
For the rest of the paper, we will assume that they are encoded in spherical coordinates.
This is in line with many equivariant models such as SOAP-GAP~\cite{Csanyi2013SOAP},
SNAP~\cite{THOMPSON2015SNAP},
ACE~\cite{ACE_ralf,ACE_equivariant_ralf} and its recursive implementations such as \cite{lysogorskiy_performant_2021},
NICE~\cite{NICE},
NequIP~\cite{nequip},
equivariant transformer~\cite{torchmd-net2022},
and SEGNNs~\cite{Brandstetter2021Geometric}.
By contrast, some equivariant MPNNs like NewtonNet~\cite{haghighatlari2021newtonnet},
EGNN~\cite{EGNN}, or PaINN~\cite{Schutt2021Painn} express the features in Cartesian coordinates.
Since models in this latter class use Euclidean vectors, which correspond to $L=1$ spherical vectors,
they fit into the same framework through a change of basis.
Spherical vectors transform according to $\bm{D}_{1}(Q)$, which correspond to $3\times3$ rotation matrices.
Some models, such as SchNet~\cite{schnet} and DimeNet~\cite{DimeNet} employ only invariant messages, i.e., $L=0$ equivariance.

\subsection{Body ordered messages}
\label{sec:body_order}

The body-order expansion of a general multivariate function is
\begin{multline}
    \label{eqn:body-order}
    F(\{r_i\}_{i=1}^{N}) =
    f^{(0)}
    + \sum_{i=1}^{N}f^{(1)}(r_{i})
    + \sum_{1 \leq i < j \leq N}^{}f^{(2)}(r_{i},r_{j}) \\
    \dots
    + \sum_{1 \leq i_{1} <...< i_{N} \leq N } f^{(N)}(r_{(i_{1})},r_{(i_{2})},...,r_{(i_{N})}).
\end{multline}
If the magnitude of higher order terms is sufficiently small so that they can be truncated,
this expansion can be a powerful tool for approximating high-dimensional functions.
The concept of body ordering also appears in quantum mechanics~\cite{thomas2021rigorous,Drautz06},
and there is ample empirical evidence that a body ordered expansion of the potential energy converges rapidly for many systems~\cite{DUSSON2022}.

By explicitly controlling the body order, one can efficiently learn low dimensional representations corresponding to low body-order terms.
This is suggested to lead to interatomic potentials with enhanced generalization ability~\cite{aPIPs2020}.
For a message $\bm{m}^{(t)}$, the body order can be defined as the largest integer $\mathcal{T}$ such that
\begin{equation}
    \label{eqn:body-order2}
    \frac{\partial^{\mathcal{T}} \bm{m}^{(t)}}{\partial r_{i_{1}} \dots \partial r_{i_{\mathcal{T}}}} \neq 0, \quad \forall \: \text{distinct} \: (i_{1},..,i_{\mathcal{T}})
\end{equation}
holds, where the elements in the tuple $(i_{1},..,i_{\mathcal{T}})$ are all distinct, and for all $\tau > \mathcal{T}$ the left-hand side of Eq.~\ref{eqn:body-order2} is identically zero~\cite{DUSSON2022,Drautz04}.

We call a model body-ordered if it can be written explicitly in the form of Eq.~\eqref{eqn:body-order} with all terms up to $\mathcal{T}$ present.
This is in contrast to non-body-ordered models in which either the expansion is infinite, or only a subset of terms are present.
To achieve body-ordering in an MPNN model one needs \textit{linear} update and readout functions.
This is because nonlinear activation functions, such as the hyperbolic tangent function (tanh), the exponential function, or a vector normalization, have infinite Taylor-series expansions, 
which make the body-order infinite without all the terms in Eq.~\eqref{eqn:body-order} being explicitly present.
For a more detailed discussion on the effect of body ordering, see Section~\ref{sec:DesignSpace}.

\section {Equivariant Atomic Cluster Expansion with Continuous Embedding and Uncoupled Channels}
\label{sec:EquiACE}

ACE~\cite{ACE_ralf, DUSSON2022} was first proposed as a framework for deriving an efficient body-ordered symmetric polynomial basis to represent functions of atomic neighborhoods.
It has been shown that many of the previously proposed \textit{symmetrized atomic field representations}~\cite{Ceriotti2021Review},
such as the Atom Centered Symmetry Functions~\cite{Behler2007ACSF},
SOAP~\cite{Csanyi2013SOAP},
the Moment Tensor Potential basis functions~\cite{MTP},
and the hyperspherical bispectrum~\cite{GAP2010} used by SNAP~\cite{THOMPSON2015SNAP}
can all be expressed in terms of the ACE basis~\cite{DUSSON2022,ACE_ralf,ACE_equivariant_ralf,lysogorskiy_performant_2021}.

In the following, we present a version of the ACE formalism for deriving $E(3)$-invariant and equivariant basis functions that incorporates a continuous embedding of chemical elements
and will serve as the main building block of the Multi-ACE framework.

\subsection{The one-particle basis}
\label{sec:one_particle_basis1}

The first in constructing the ACE framework is to define the one-particle basis, which is used to describe the spatial arrangement of atoms $j$ around the atom $i$:
\begin{equation}
    \label{eqn:one-particle_simple}
    \phi_{nlm z_i z_j}(\bm{r}_{ji}) = R_{nlz_i z_j}(r_{ji}) Y_l^m(\bm{\hat{r}}_{ji}),
\end{equation}
where the index $z_i$ and $z_j$ refer to the chemical elements of atoms $i$ and $j$.
The one-particle basis functions are formed as the product of a set of orthogonal radial basis functions $R_{nl}$ and spherical harmonics $Y_l^m$.
The positional argument $\bm{r}_{ji}$ in Eq.~\eqref{eqn:one-particle_simple}
can be obtained from $(\sigma_i^{(t)}, \sigma_j^{(t)})$, thus making the value of the one-particle basis function depend on the states of two atoms.

The formulation in Eq.~\ref{eqn:one-particle_simple} uses discrete chemical element labels.
The drawback of this approach is that the number of different basis functions rapidly increases with the number of chemical elements in the system.
Given $S$ different chemical elements and maximum body-order $N$, the number of basis functions is proportional to $S^N$.
By contrast, MPNNs typically leverage a learnable mapping from the discrete chemical element labels to a continuous fixed-length representation. 
Using such an embedding with ACE eliminates the scaling of the number of basis functions with the number of chemical elements.
The one-particle basis can be generalized to allow for this continuous embedding via a set of functions whose two indices we explain below:
\begin{equation}
    \label{eqn:one-particle_general1}
    \phi_{kv}(\sigma_{i},\sigma_{j}) =
    R_{kcl}(r_{ji}) Y^{m}_{l}(\bm{\hat{r}}_{ji}) T_{kc}(\bm{\theta}_{i},\bm{\theta}_{j}),
\end{equation}
where $T_{kc}$ is a generic function of the chemical attributes $\bm{\theta}_{i}$ and $\bm{\theta}_{j}$ and is endowed with two indices, $k$ and $c$, and the radial basis likewise.
Of these, $c$, together with $l$ and $m$, will be {\em coupled} together when we form many-body basis functions (see Eq.~\eqref{eq:product-basis} below).
These coupled indices are collected into a single multi-index $(v \equiv lmc)$ for ease of notation. We refer to $k$ as the {\em uncoupled} index.

Beyond the chemical element labels, $T_{kc}$ can account for the dependence of the one-particle basis functions on other attributes of the atoms, such as the charge, magnetic moment \cite{ACE_equivariant_ralf}, or learnable features.
Furthermore, the output of $T_{kc}$ can be invariant or equivariant to rotations.
In the case of equivariant outputs, the indices $k$ (in the uncoupled case) or $c$ (in the coupled case) will themselves be multi-indices
that contain additional indices (e.g., $l'$ and $m'$) describing the transformation properties of these outputs.

To recover Eq.~\eqref{eqn:one-particle_simple} with the discrete element labels,
we set $\theta_i, \theta_j$ to $z_i, z_j$ and assume $k \equiv 1$, i.e., there are no uncoupled indices.
Further, we choose $c$ to be a multi-index, $(c\equiv{n z_i z_j})$, with $T_{kc}$ being an index selector, $T_{kc} = T_{1 n z_i z_j} = \delta_{z_i \theta_i}\delta_{z_j \theta_j}$.
In this case, the index $n$ of the radial basis $R_{nl z_i z_j}$ in Eq.~\eqref{eqn:one-particle_simple} is also part of the ``coupled'' multi-index $c$.

In the language of MPNNs, the values of the one-particle basis functions would be thought of as edge features of a graph neural network model.
This graph would be directed since the one-particle basis functions are not symmetric with respect to the swapping of the central atom $i$ and the neighbor atom $j$.

\subsection{Higher order basis functions}
\label{sec:higher_order_basis}
%
A key innovation of ACE was the construction of a complete many-body basis, which can be computed at a constant cost per basis function~\cite{2022-recursive}.
The high body-order features can be computed without having to explicitly sum over all triplets, quadruplets, etc., which is achieved by what came to be called the ``density trick''~\cite{Ceriotti2021Representation}, introduced originally for the fast evaluation of high body order descriptors~\cite{GAP2010,Csanyi2013SOAP}.
This allows any $E(3)$-equivariant function of an atomic neighbourhood to be expanded using a systematic body ordered expansion at a low computational cost \cite{DUSSON2022}.

The next step of the ACE construction is analogous to traditional message passing:
we sum the values of the one-particle basis functions evaluated on the neighbors to form the atomic- or $A$-basis. This corresponds to a projection of the one-particle basis on the atomic density.
Therefore, in the atomic environment representation literature, this step is often referred to as the  \emph{density projection}~\cite{Ceriotti2021Review},
\begin{equation}
    \label{eq:atomic-basis}
    A_{i,kv} = \sum_{j \in \mathcal{N}(i)} \phi_{kv}(\sigma_{i}, \sigma_{j}).
\end{equation}
The $A$-basis is invariant with respect to the permutation of the neighbor atoms,
and its elements are 2-body functions in the sense of the definition in Eq.~\eqref{eqn:body-order}.
This means that this basis can represent functions that depend on all neighbors' positions but can be decomposed into a sum of 2-body terms.

Then, to create basis functions with higher body-order, we form products of the $A$-basis functions to obtain the product basis, $\bm{A}_{i,k\bm{v}}$:
\begin{equation}
    \label{eq:product-basis}
    \begin{gathered}
        \bm{A}_{i,k\bm{v}}
        = \prod_{\xi=1}^{\nu} A_{{i},kv_{\xi}},
        \quad\bm{v} = (v_{1},...,v_{\nu}),
    \end{gathered}
\end{equation}
where $\nu$ denotes the correlation order and the array index $\bm{v}$ collects the multi-indices of the individual $A$-basis functions, representing a $\nu$-tuple.
The product basis is a complete basis of permutation-invariant functions of the atomic environment.

Taking the product of $\nu$ $A$-basis functions results in basis functions of correlation order $\nu$, which thus have body-order $\nu + 1$, on account of the central atom. In the language of density-based representations, these tensor products correspond to $\nu$-correlations of the density of atoms in the atomic neighborhood \cite{nigam2022unified}.

For example, the $\nu=3$, four-body basis functions have the form
\begin{equation}
    \boldsymbol{A}_{i, k\boldsymbol{v}} = A_{i,kv_1}A_{i,kv_2}A_{i,kv_3},
\end{equation}
where $\boldsymbol{v}=(v_1v_2v_3)$. This illustrates the difference between the uncoupled $k$ channels and the coupled $v$ channels - we did not form products with respect to the indices collected in $k$.
Note that in linear ACE, as described in Refs.~\cite{ACE_ralf, DUSSON2022, kovacs2021}, the tensor product is taken with respect to all of the indices (radial, angular, and chemical elements) that are in $\bm{v}$, and no uncoupled indices are used.

\subsection{Symmetrization of basis functions}
%
The product basis constructed in the previous section linearly spans the space of permutationally and translationally invariant functions but does not account for rotational invariance or equivariance of the predicted properties or intermediate features.
To create rotationally invariant or equivariant basis functions, the product basis must be symmetrized with respect to O(3).
The symmetrization takes its most general form as an averaging over all possible rotations of the neighborhood.
In the case of rotationally invariant basis functions, this averaging is expressed as an integral of the product basis over rotated local environments,
\begin{equation}
    \label{eq:invariant-overcomplete3}
    B_{i, {k \bm v}} := \int_{O(3)}  {\bm A}_{i, k \bm v}\Big(\big\{ Q \cdot \big(\sigma_i, \sigma_j\big) \big\}_{j \in \mathcal{N}(i)} \Big) \, dQ,
\end{equation}
where we make explicit the dependence of the product basis on the atomic states,
and $Q \cdot (\sigma_i, \sigma_j) = (Q \cdot \sigma_i, Q \cdot \sigma_j)$ denotes the action of the rotation on a pair of atomic states.
The above integral is purely formal.
To explicitly create a spanning set of the symmetric $B$ functions above, one can instead use tensor contractions
as the angular dependence of the product basis is expressed using products of spherical harmonics (see Eq.~\eqref{eq:symmetrized_basis_coupling} below).

The construction of Eq.~\eqref{eq:invariant-overcomplete3} is readily generalized if equivariant features are required~\cite{zhang2022equivariantACE, NICE, ACE_equivariant_ralf}.
If the action of a rotation $Q$ on a feature ${\bm h}$ is represented by a matrix ${\bm D}(Q)$, then we can write the equivariance constraint as
\begin{equation}
    {\bm D}(Q)^{-1} {\bm h}\Big(\big\{ Q \cdot \big(\sigma_i, \sigma_j\big) \big\}_{j \in \mathcal{N}(i)} \Big) = {\bm h}\Big(\big\{\sigma_i, \sigma_j \big\}_{j \in \mathcal{N}(i)} \Big).
\end{equation}
To linearly expand ${\bm h}$, the basis functions must satisfy the same symmetries which is achieved by defining the symmetrized basis as
\begin{equation}
    \label{eq:equivariant-overcomplete}
    \begin{aligned}
         & B_{i, k\bm v, \alpha} =                                                                                                                              \\
         & \int_{O(3)} (\bm{D}(Q)^{-1} e_\alpha) {\bm A}_{i,k\bm v }\Big(\big\{ Q \cdot \big(\sigma_i, \sigma_j\big) \big\}_{j \in \mathcal{N}(i)} \Big) \, dQ,
    \end{aligned}
\end{equation}
where the $e_\alpha$ are a basis of the feature space ${\bm h}$. This approach can be applied to parameterize tensors of any order, both in Cartesian and spherical coordinates.
For instance, if the ${\bm h}$ we are representing a Euclidean 3-vector, the $e_\alpha$ can just be the three Cartesian unit vectors, $\bm{\hat x}$, $\bm{\hat y}$, and $\bm{\hat z}$.

Going forward, we focus on features with spherical $L$-equivariance and label them accordingly as ${\bm h}_{L}$ and the corresponding basis functions as $B_{i, k\bm \alpha LM}$.
The matrices ${\bm D}(Q)$ become the Wigner-D matrices, i.e., ${\bm D}^L(Q)$.

The integration over the rotations can be reduced to recursions of products of Wigner D-matrices and carried out explicitly as a tensor contraction~\cite{DUSSON2022, NICE}.
It is then possible to create a spanning set of $L$-equivariant features of the integrals of the types of Eqs.~\ref{eq:invariant-overcomplete3} and \ref{eq:equivariant-overcomplete} using linear operations.
This can be done by introducing the generalized coupling coefficients:
\begin{equation}
    \begin{aligned}
        \label{eq:symmetrized_basis_coupling}
        B_{i, k\eta, LM}
         & = \sum_{\bm v} \mathcal{C}_{\eta,\bm v}^{LM}
        {\bm{A}}_{i, k\bm{v}},
    \end{aligned}
\end{equation}
where $\mathcal{C}_{\eta,\bm v}^{LM}$ are the coupling coefficients corresponding to correlation order $\nu$ and imposed equivariance $L$.
The output index $\eta$ enumerates the different possible combinations of $\bm A_{i,k\bm v}$ that have equivariance $L$.
For a detailed discussion of the invariant case, see Ref.~\citenum{DUSSON2022}.

\begin{figure}
    \includegraphics[width=\linewidth]{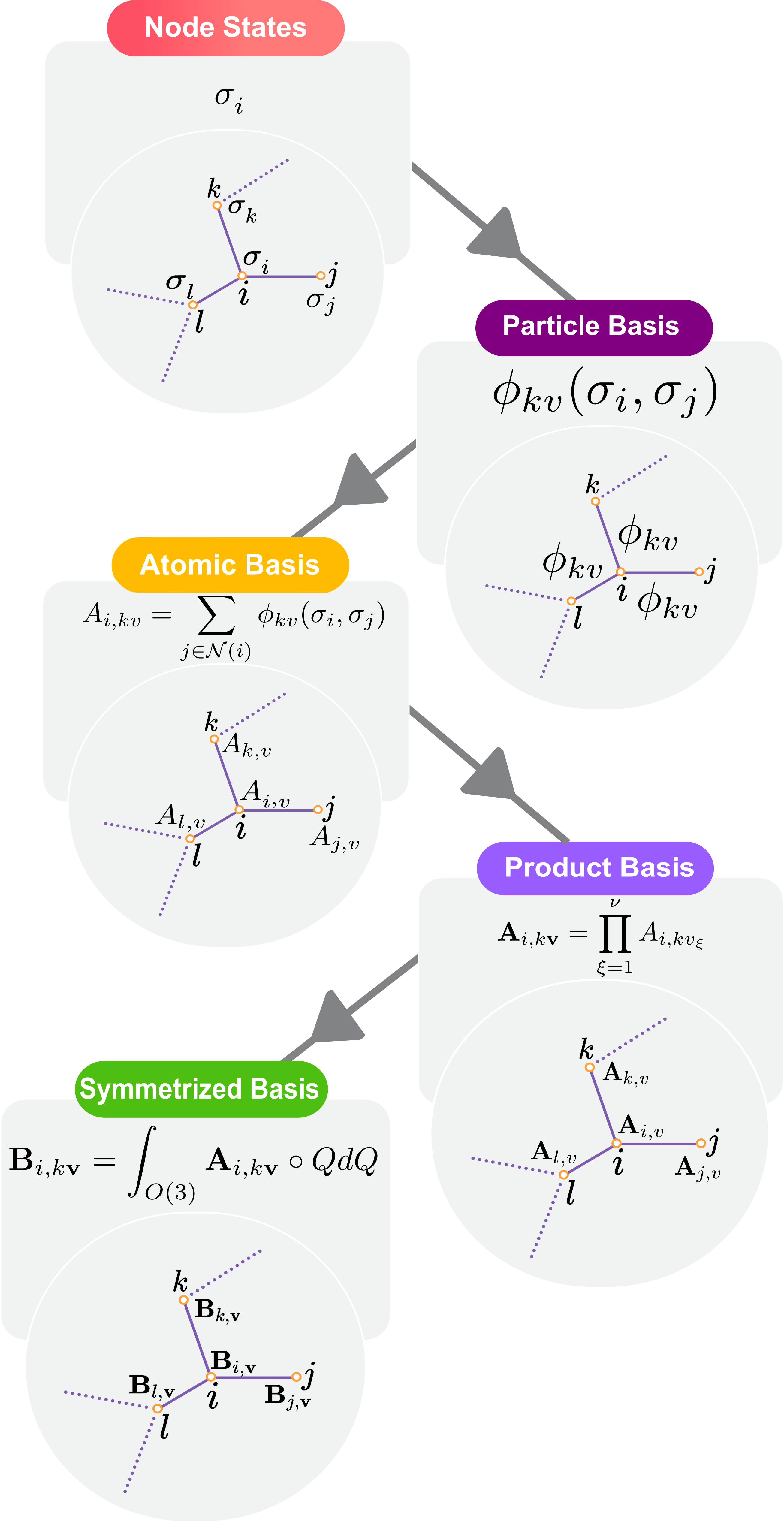}
    \caption{
        Illustration of the construction of high body-order ACE features.
        First, a neighborhood graph is constructed with each node being labeled by its state.
        Then, the one-particle basis is computed for each edge.
        After that, a pooling operation is performed to create permutation invariant $A$-functions of semi-local environments.
        To construct higher body-order features, the product basis is formed by taking tensor-products of all coupled indices of the $A$-functions.
        Finally, to create equivariant messages, the $\bm{B}$ basis is formed by first specifying the required equivariance and evaluating the corresponding symmetrisation integral.
        We illustrate here the case of invariant $\bm{B}$ basis.
    }
    \label{fig:product_symmetrization}
\end{figure}

Using spherical coordinates for the features, $\mathcal{C}_{\eta,\bm v}^{LM}$ corresponds to the generalized Clebsch-Gordan coefficients,
and the symmetry label $L$ corresponds to the usual labeling of the $O(3)$ irreducible representations.
An additional degree of freedom is to have a different ${\bm{A}}_{i, k\bm vL}$ product basis for each symmetry $L$ (e.g., by choosing different one-particle basis functions depending on $L$).
This is a choice made for NequIP and is discussed in more detail in Section~\ref{sec:MultiACE_general}.
Creating symmetric high body-order basis functions is summarized in Figure~\ref{fig:product_symmetrization}.

The functions $B_{i, k\eta, LM}$ form a spanning set, meaning that all $(\nu+1)$-body functions with symmetry $L$ of the atomic environment can be represented as a linear combination of $B$ functions~\cite{DUSSON2022}.
The values of the $B$ functions can be combined into an output $m_{i,kLM}$ on each atom $i$ and each channel $k$ via a learnable linear transformation
\begin{equation}
    \label{eqn:message_ace}
    m_{i,k LM} = \sum_{\eta}w_{k\eta L} B_{i, k\eta, LM}.
\end{equation}

Finally, to generate the target output for atom $i$, the uncoupled channels $k$ can be mixed via a learnable (linear or non-linear) function $\Phi_{i,L} = \mathcal{F}(\bm m_{i,L})$.
For a more detailed discussion on the precise form of non-linearities, see Section~\ref{sec:non_linearity}.

\section{Multi-ACE: A General Framework of Many-Body Equivariant Message Passing Interatomic Potentials}
\label{sec:MultiACE_general}

In this section, we show how multiple equivariant ACE layers introduced in Section~\ref{sec:EquiACE} can be combined to build a message passing model \cite{Bochkarev_2022, MACE2022}.
The resulting framework encompasses most equivariant MPNN-based interatomic potentials. 
In the case of using a single message passing layer, the framework can be reduced to linear ACE or the other atom-centered descriptor-based models. 

To create a Multi-ACE model, we need to specify how the output of one ACE layer is used in the next layer.
This is done by updating the state of the atoms by assigning the output of the previous layer to the feature $\bm h_i^{(t+1)}$:
\begin{equation}
    \label{eq:multiACE_update}
    \begin{split}
        \sigma_i^{(t+1)} &= ({\bm r}_{i}, {\bm \theta}_i, {\bm h}_i^{(t+1)}),  \\
        {\bm h}_{i}^{(t+1)} &= U_{t}( \sigma_i^{(t)},  \bm m^{(t)}_{i}),
    \end{split}
\end{equation}
where $\bm m_{i}^{(t)}$ is a set of messages at iteration $t$ as defined in Eq.~\ref{eqn:message_ace} and $U_t$ is the update function for each layer.
In most MPNNs, the $k$ channel of the message corresponds to the dimension of the learned embedding of the chemical elements~\cite{schnet, nequip}.
We further need to extend Eq.~\eqref{eqn:one-particle_general1} to incorporate the dependence on the output of the previous ACE layer, which can be achieved by making it an argument of the $T_{kc}$ functions
\begin{multline}
    \label{eqn:one-particle_general2}
    \phi_{kvL}^{(t)}(\sigma_{i}^{(t)},\sigma_{j}^{(t)}) = \\
    R^{(t)}_{kcl_1L}(r_{ji}) Y^{m_1}_{l_1}(\bm{\hat{r}}_{ji}) T^{(t)}_{kcL}(\bm{h}_{j}^{(t)},\bm{\theta}_{i},\bm{\theta}_{j}),
\end{multline}
where $(v \equiv l_1m_1 c)$.
We have also added the index $L$ to the one-particle basis to enable having a different set of one-particle basis functions for messages $m_{i,kLM}$ with different symmetry $L$.

\begin{table*}
    \centering
    \label{tab:mpnn_schnet_nequip_ace}
    \caption{
        Different machine learning potentials in the framework of MPNNs.
        We identify SchNet, NequIP, and ACE as examples of MPNNs and exhibit their explicit components in the design space:  the message, symmetric pooling, and update functions.
        Note that in NequIP, the choice of non-linearity is not fixed, and we have chosen a normed activation with $\tanh$ to be shown here.
        In each case, learnable parameters (weights) are shown as $W$ and biases as $b$.
    }
    \setlength{\aboverulesep}{0pt}
    \setlength{\belowrulesep}{0pt}
    \resizebox{\textwidth}{!}{
        \renewcommand{\arraystretch}{2}
        \begin{tabular}{l|c|c|c}
            \toprule
                                                                         & SchNet                                                           & NequIP                                                                                & Linear ACE                                                                                                \\ \midrule
            \strut Message function $M_{t}$                              & $R_{k}^{(t)} \left(||r_{j} - r_{i}|| \right) h_{j,k}^{(t)}$            & $R_{kl_1l_{2}L}^{(t)}(r_{ji})Y^{m_1}_{l_1}(\bm{\hat{r}_{ji}})h^{(t)}_{j,kl_{2}m_{2}}$ & $R_{n}(r_{ji}) Y^{m}_{l}(\bm{\hat{r}}_{ji})\delta_{z_i \theta_i}\delta_{z_j \theta_j}$                    \\
            \strut Symmetric pooling  $\bigoplus_{j \in \mathcal{N}(i)}$ & $\sum_{j \in \mathcal{N}(i)}$                                    & $\sum_{l_{1}m_{1}l_{2}m_{2}} \mathcal{C}_{l_{1}m_{1}l_{2}m_{2}}^{LM}
             \sum_{j \in \mathcal{N}(i)}$               & $\sum_{\eta }w_{ \eta }\ \sum_{\bm{v}} \mathcal{C}_{\eta,\bm v}^{00}\prod_{\xi=1}^{\nu} \sum_{j \in \mathcal{N}(i)}$ \\
            \strut Update function   $U_{t}$                             & $\boldsymbol{h}^{(t)}_{i} + \tanh \left( W^{(t)}\boldsymbol{m}_{i}^{(t)} +\boldsymbol{b}^{(t)} \right)$ & $\boldsymbol{h}^{(t)}_{i} + \tanh \left( \|W^{(t)}\boldsymbol{m}_{i}^{(t)}\|^{2} \right) W^{(t)}\boldsymbol{m}_{i}^{(t)}$    & -                                                                                     \\
            \bottomrule
        \end{tabular}
    }
\end{table*}

We now relate the equations of the MPNN framework (see Section~\ref{sec:MPNN}) to those of the Multi-ACE framework.
First, we identify the message function $M_t$ with the one-particle basis of Eq.~\eqref{eqn:one-particle_general2}:
\begin{equation}
    M_t(\sigma_i^{(t)}, \sigma_j^{(t)}) := M_{kvL}^{(t)}(\sigma_i^{(t)}, \sigma_j^{(t)}) =  \phi_{kvL}^{(t)}(\sigma_{j}^{(t)},\sigma_{i}^{(t)}).
\end{equation}
Next, we define the permutation invariant pooling operation $\bigoplus_{j \in \mathcal{N}(i)}$ of Eq.~\eqref{eqn:message_func}.
To obtain a symmetric many-body message $m_{i,kLM}^{(t)}$ of correlation order $\nu$,
the pooling operation must map the one-particle basis that is two-body to a set of many-body symmetric features that can be combined in a learnable way to form the message on each nodes.
This is what the ACE formalism of Section~\ref{sec:EquiACE} achieves.
This way, we obtain the central equation of Multi-ACE:
\begin{multline}
    \label{eq:multi_ace_message}
    m_{i,kLM}^{(t)} = \bigoplus_{j \in \mathcal{N}(i)} M_t(\sigma_i^{(t)}, \sigma_j^{(t)}) = \\
    \sum_{\eta }w_{i, k\eta L}^{(t)}\ \sum_{\bm{v}} \mathcal{C}_{\eta,\bm v}^{LM}
    \prod_{\xi=1}^{\nu} \sum_{j \in \mathcal{N}(i)} \phi_{kv_{\xi}L}^{(t)}(\sigma_{i}^{(t)},\sigma_{j}^{(t)}),
\end{multline}
where $w_{i, k\eta L}^{(t)}$ are learnable weights, and $\nu$ is the maximum correlation order, which equals to the body-order minus 1.
$\mathcal{C}_{\eta,\bm v}^{LM}$ denotes the generalized Clebsch-Gordan coefficients defined in Eq.~\eqref{eq:symmetrized_basis_coupling}.
The general scheme of higher order message passing is illustrated in Figure~\ref{fig:product_symmetrization}.

\begin{figure}
    \centering
    \includegraphics[width=0.25\textwidth]{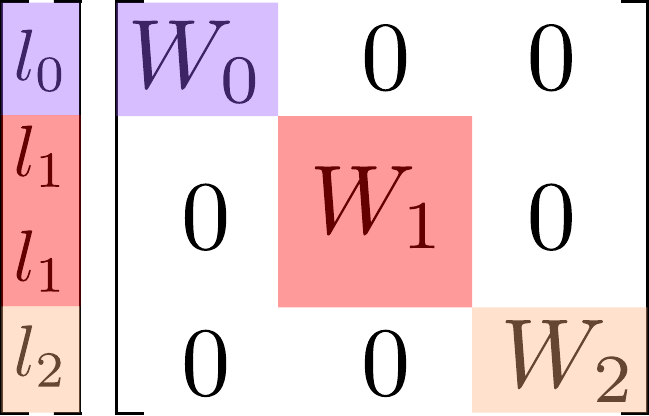}
    \caption{
    Block structure of weight matrices for an equivariant linear operation.
    As only linear combinations of features of the same representations (here $l_{0},l_{1},l_{2}$) are allowed to interact, the weight matrix is block diagonal.
    }
    \label{fig:blockdiag}
\end{figure}

The update function $U_{t}$ from Eq.~\eqref{eqn:update_function} corresponds to a learnable linear combination of the uncoupled channels of the symmetrized message.
$U_{t}$ can be written as
\begin{equation}
    \label{eqn:ACE_update}
    h^{(t+1)}_{i,kLM} = U_{t}(\sigma_{i}^{(t)},\bm{m}_{i}^{(t)}) = \sum_{\tilde{k}} W^{(t)}_{k\tilde{k}L}m_{i,\tilde{k}LM}^{(t)}
\end{equation}
with $W^{(t)}$ being a block diagonal weight array (cf. Figure~\ref{fig:blockdiag}) of dimension $[N_\text{channels} \times N_\text{channels} \times L_\text{max}]$,
$N_\text{channels}$ is the number of uncoupled $k$ channels in the message and $L_\text{max}$ is the maximum order of symmetry in the message that is passed from one layer to the next.
$U_{t}$ can also depend on the attributes (e.g., the chemical element) of the central atom via a so-called ``self-connection'' (see Section~\ref{sec:DesignSpace} for details).
The update functions acting on equivariant features can also be non-linear, but for that, it has to have a particular form (See Ref.~\cite{WeilerGatedNonLinearities2018} and Appendix~\ref{sec:non_linear_equivariant}).

After the $T$-th layer, a learnable (linear or non-linear) readout function that can depend on the final message or all previous ones gives the site energy of atom $i$.

\begin{table*}
    \centering
    \caption{
        Different choices in the Multi-ACE formalism lead to different models in the literature.
        The internal $l_\text{max}$ specifies the angular information contained on the messaging function $M_{t}$ indexed by the highest weights of the irreducible representations of $O(3)$.
        The update $L_\text{max}$ specifies the angular information in the update function.
        The local correlation order is the correlation order of the first message $m_{i}^{(0)}$.
        The total correlation order corresponds to the correlation order of the entire model as a function of individual atoms.
        The models above the separation line correspond to spherical equivariant interatomic potentials and the models under to Cartesian equivariant interatomic potentials.
    }
    \resizebox{\textwidth}{!}{
        \begin{tabular}{lcccccccccc}
            \toprule
                                                                 & $l_\text{max}$ & Update $L_\text{max}$ & Local correlation order ($\nu$) & Number of layers ($T$) & Total correlation order & $T^{(t)}_{kc} \left( \bm{h}_{j}^{(t)},\theta_{i},\theta_{j} \right)$ & Coupling ($v$) \\ \midrule
            \textbf{SOAP} \cite{Csanyi2013SOAP}                  & $\geq 3$       & 0                     & 2                               & 1                      & $\geq 3 $               & $\delta_{z_i \theta_i}\delta_{z_j \theta_j}$                         & nlm            \\
            \textbf{Linear ACE} \cite{kovacs2021}                & $\geq 1$       & 0                     & $\geq 1 $                       & 1                      & $\geq 3$                & $\delta_{z_i \theta_i}\delta_{z_j \theta_j}$                         & nlm            \\
            \textbf{SchNet} \cite{schnet}                        & 0              & 0                     & 1                               & T $ \geq 2$            & T                       & $h_{j,kl=0}^{(t)} $ (Scalars)                                        & $\emptyset$    \\
            \textbf{DimeNet} \cite{DimeNet}                      & 0              & 0                     & 2                               & T $ \geq 2$            & 2T                      & $h_{j,l=0}^{(t)}$ (Scalars)                                          & $\emptyset$    \\
            \textbf{Cormorant} \cite{Anderson2019CormorantCM}    & $\geq 1$       & $\geq 1$              & 1                               & T $\geq 2$             & T                       & $h_{j,klm}^{(t)}$  (Spherical Vec.)                                  & lm             \\
            \textbf{NequIP} \cite{nequip}                        & $\geq 1$       & $\geq 1$              & 1                               & T $\geq 2$             & T                       & $h_{j,klm}^{(t)}$  (Spherical Vec. )                                 & $l_1m_1l_2m_2$ \\
            \textbf{GemNet} \cite{klicpera2022gemnet}            & $\geq 1$       & $\geq 1$              &  3                               & T $\geq 2 $            & T                       & $h_{j,klm}^{(t)}$ (Spherical Vec.)                                   & $l_1m_1l_2m_2$ \\
            \midrule
            \textbf{NewtonNet} \cite{haghighatlari2021newtonnet} & 1              & 1                     & 1                               & T $\geq 2$             & T                       & Cartesian Vectors                                                    & -              \\
            \textbf{EGNN} \cite{EGNN}                            & 1              & 1                     & 1                               & T $\geq 2$             & T                       & Cartesian Vectors                                                    & -              \\
            \textbf{PaINN} \cite{Schutt2021Painn}                 & 1              & 1                     & 1                               & T $ \geq 2$            & T                       & Cartesian Vectors                                                    & -              \\
            \textbf{TorchMD-Net} \cite{torchmd-net2022}          & 1              & 1                     & 1                               & T $\geq 2 $            & T                       & Cartesian Vectors                                                    & -              \\
            \bottomrule
        \end{tabular}}
    \label{tab:different_models}
\end{table*}

\subsection{Coupling of channels}
\label{sec:coupling}

An important design choice of ACE models is how channels interact when forming the product basis.
This choice significantly affects the scaling of the number of features, and hence, it is an essential part of the design space.
This is best illustrated by considering the degree of freedom regarding the handling of different chemical elements.
In the case of general linear ACE and other similar descriptors like SOAP, the element channel of the one-particle basis is a discrete index.
When forming the higher order many-body basis functions that will produce the features, these channels are coupled, forming all possible combinations.
For example, if there are four different chemical elements, the number of 3-body basis functions will be proportional to $4^3$.
The alternative approach, employed by most MPNNs, is to map the chemical elements to a set of fixed-length vectors via a learnable transformation.
When the higher-order features are formed during the message passing phase, these channels do not get coupled; hence the number of features does not depend on the number of chemical elements.
Instead, the channels are mixed during the update phase.

Similar choices can be made for the radial basis functions.
Linear ACE uses orthonormal radial basis functions and forms all possible combinations (up to truncation by maximum polynomial degree) for the higher order features.
For example, for the 3-body functions, the radial part has the form $R_1(\bm{r}_{ij})R_2(\bm{r}_{ik})$ for all allowed combinations of $R$'s. 
By contrast, NequIP learns a separate (non-linear) combination of radial features for each one-particle basis, as shown in Equation~\eqref{eq:NequipConv}.
Therefore, there is a single learnable radial basis function $R_{kl_1l_2L}^{(t)}$ for each channel $k$, spherical harmonic $l_1$, neighbour feature symmetry $l_2$, and output symmetry $L$.  
The uncoupled channels $k$ only get mixed during the update phase. 

The analysis within the design space leads to the question of the optimal amount of coupling within the product basis in the spectrum between the full coupling of linear ACE and no coupling in NequIP.

\subsection{Interpreting models as Multi-ACE}
\label{sec:interpreting_models_as_multiACE}

The Multi-ACE framework includes many of the previously published equivariant message passing networks.
The most basic specification of a multi-ACE model considers the number of layers $T$, 
the correlation order of each layer $\nu$, 
the internal order of the spherical harmonic expansion within the layer in the one-particle basis $l_\text{max}$,
and the order of the spherical harmonics in the message passing phase after symmetrization, $L_\text{max}$.
Other choices include the type of features (Cartesian or spherical basis) and the type of dependence of the radial basis on the indices $kcl$ in Eq.~\eqref{eqn:one-particle_general2}.
Note that the pointwise non-linearities present in some of those models affect both the local correlation and the total correlation, as discussed in Section~\ref{sec:DesignSpace}.
For simplicity, we chose not to consider them for the following discussion.
A comparison of different models' design choices are summarized in Table~\ref{tab:different_models}.

The convolution of the SchNet network can be obtained by considering $T \geq 2$, $\nu=1$, $L=0$, and $l_\text{max}=0$.
The DimeNet invariant message passing network includes higher correlation order messages (more precisely, 3-body messages by incorporating angular information),
meaning that $T \geq 2$, $\nu = 2 $, $L_\text{max}=0$, and $l_\text{max} = 5$.
NequIP corresponds to $T \geq 2$, $\nu = 1 $ and $L_\text{max} \geq 1$, and $l_\text{max} = L_\text{max}$,
where the symmetrization of Eq.~\eqref{eq:multi_ace_message} can be simplified:
\begin{equation}
    \label{eq:NequipConv}
    \begin{aligned}
         & m_{i,kLM}^{(t)} =                                                                                                                                               \\
         & \sum_{l_1 m_1 l_2 m_2} C^{LM}_{l_1 m_1 ,l_2 m_2} \sum_{j \in \mathcal{N}(i)} R_{k l_1 l_2L}^{(t)}(r_{ji}) Y^{m_1}_{l_1} (\bm{\hat{r}}_{ji}) h^{(t)}_{j,kl_2m_2}
    \end{aligned}
\end{equation}

The models in the lower part of the table do not use a spherical harmonics expansion but work with Cartesian tensors.
Nonetheless, they fit into this framework by considering the equivalence of vectors and $l=1$ spherical tensors.
The coordinate displacements present in, for example, EGNN~\cite{EGNN} and NewtonNet~\cite{haghighatlari2021newtonnet} can thus be rewritten as an $l=1$ spherical expansion of the environment via a change of basis.

Based on the models presented in Table~\ref{tab:different_models}, the Multi-ACE framework lets us identify two main routes that have been taken thus far in building interatomic potentials. The models have either few layers and high local correlation order, like linear ACE (and other descriptor-based models), or many layers and low local correlation order, such as NequIP. 

\subsection{Message passing as a chemically inspired sparsification}
\label{sec:sparsification}

A central aspect of message passing models is the treatment of semi-local information: 
while in approaches such as ACE, the atomic energy is only influenced by neighboring atoms within the local cutoff sphere, the message passing formalism iteratively propagates information, allowing for semi-local information to be communicated.
Equivariant MPNNs like NequIP update atom states based on a tensor product between edge features and neighboring atoms' states, which leads to ``chain-like'' information propagation.

In particular, consider a much-simplified message passing architecture with a single channel $k$ and an update $U$ which is just the identity:
\begin{equation}
    \begin{aligned}
         & h_{i,LM}^{(t+1)} = \\
         & \sum_{l_1m_1l_2m_2}C^{LM}_{l_1m_1,l_2m_2}\sum_{j \in \mathcal{N}(i) } R_{l_1l_2L}^{(t)}(r_{ji})Y^{m_1}_{l_1}(\bm{\hat{r}}_{ji}) h^{(t)}_{j,l_2m_2}
    \end{aligned}
\end{equation}
We can write out the simple example of a two-layer update explicitly:
\begin{widetext}
    \begin{equation}
        \begin{aligned}
             h_{i,LM}^{(2)} &= \sum_{l_1m_1l_2m_2}C^{LM}_{l_1m_1,l_2m_2}\sum_{j_{1} \in \mathcal{N}(i) } R_{l_1l_2L}^{(t)}(r_{ji})Y^{m_1}_{l_1}(\bm{\hat{r}}_{ji}) h^{(1)}_{j_{1},l_2m_2}  \\ &= \sum_{l_1m_1l_2m_2}C^{LM}_{l_1m_1,l_2m_2}\sum_{j_{1} \in \mathcal{N}(i) } R_{l_1l_2L}^{(t)}(r_{ji})Y^{m_1}_{l_1}(\bm{\hat{r}}_{ji})    \sum_{j_{2} \in \mathcal{N}(j_{1}) } R_{l_2}^{(t)}(r_{ji})Y^{m_2}_{l_2}(\bm{\hat{r}}_{ji})h^{(0)}_{j_{2}}
        \end{aligned}
    \end{equation}
\end{widetext}
where we have assumed that $h^{(0)}_{j_{2}}$ is a scalar, learnable embedding of the chemical elements, such that it doesn't possess the $l$ index.

\begin{figure}[!ht]
    \centering
    \includegraphics[width=0.9\columnwidth]{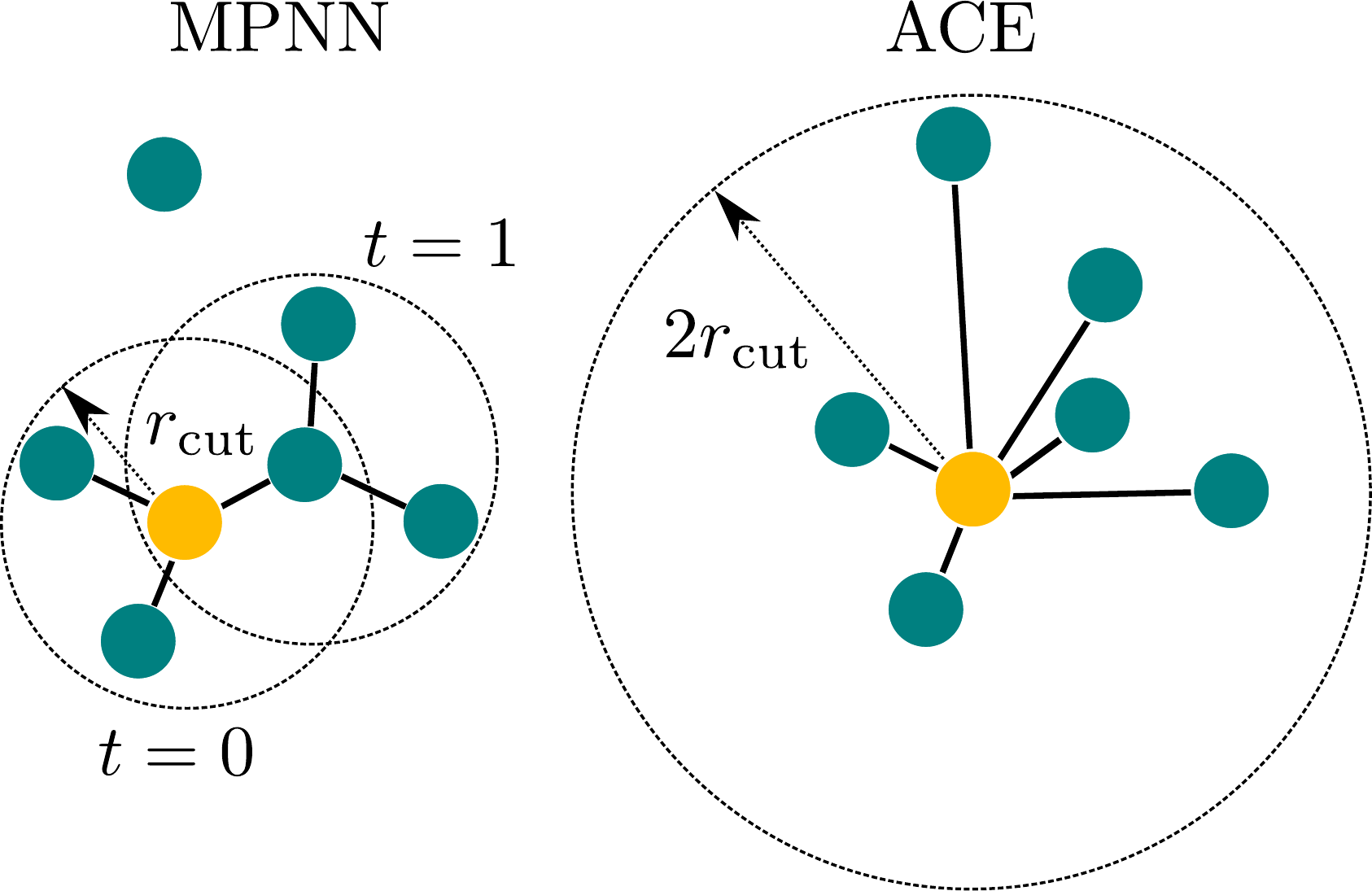}
    \caption{
        Comparison of the clusters formed by two iterations of message passing with cutoff $r_\text{cut}$ at each iteration on the left and the clusters formed by ACE with cutoff $2 r_\text{cut}$ on the right.
        In principle, both methods incorporate information from a distance of up to $2 r_\text{cut}$, but in the case of the MPNN, only atoms that can be reached through a chain of closer intermediates contribute.
    }
    \label{fig:message-passing}
\end{figure}

This defines a pattern of information flow in which the state of $j_{2}$ is first passed onto atom $j_{1}$, resulting in the $(j_{2}, j_{1})$-correlation being captured.
This is then passed onto atom $i$, which encodes the 3-body interaction between atoms $(i, j_{1}, j_{2})$ on atom $i$. 
This scheme induces a chain-wise propagation mechanism ($j_{2} \rightarrow j_{1} \rightarrow i$), which is different from the local models like ACE,
in which the three-body correlation on atom $i$ stems from an interaction between $(i, j_{1})$ and $(i, j_{2})$.

One can then, under the assumption of linearity,
view equivariant MPNNs as a \emph{sparsification} of an equivalent one-layer ACE model but which has a larger cutoff radius $r_\text{cut,ACE} = T \times r_\text{cut,MPNN}$,
where $T$ denotes the number of message passing steps and $r_\text{cut,ACE}$ is the maximal distance of atoms that can see each other in a $T$ layer MPNN.
While in a one-layer ACE, all clusters with central atom $i$ would be considered,
the MPNN formalism sparsifies this to only include walks along the graph (the topology of which is induced by local cutoffs) of length $T$ that end on atom $i$.

In practice, for typical settings of $T$, $r_{\rm cut}$, and $\nu$, a local model like ACE with a cutoff of $T \times r_{\rm cut}$ would be impractical due to the large number of atoms in the neighborhood.
Moreover, the clusters created by atom-centered representations for an equivalent cutoff to MPNNs are less physical, as illustrated in Figure~\ref{fig:message-passing}.
Most physical interactions in chemistry are short-ranged and semi-local information propagate in a chain-like mechanism, thus making the message passing \emph{sparsification} correspond to chemical bond topology.
A more in-depth discussion on the relationship between message passing and semi-local information can be found in \cite{nigam2022unified,Bochkarev_2022}.

\section{BOTNet: Body Ordered Equivariant Network}
\label{sec:BOTNet}

The design space of the Multi-ACE framework provides a setting to study the choices made by different approaches.
The most accurate model published to date is NequIP, which uses an equivariant 2-body message passing scheme.
In Section~\ref{sec:DesignSpace}, we probe the NequIP architecture to understand which parts are crucial for its success and study how changing different parts of the architecture affects the properties of the fitted potential energy surface, including smoothness and out-of-domain extrapolation.
The new model introduced in this section, BOTNet, is a simplified, body-ordered version of NequIP.
We keep the two-body interactions of NequIP within each layer, and the body-order is increased by one in each iteration of the message passing.
This is made possible by removing all pointwise non-linearities in the update, except in the last layer. The different body-ordered contributions to the total energy are predicted as a sum of functions of the learnable $t+1$-body features $h_{i}^{(t)}$ at each iteration.
Note that the BOTnet model is still a nonlinear function of its parameters due to the tensor product operation in the message block.

The final energy expression of BOTnet can be written as a body-ordered energy expansion,
\begin{equation}
    E_{i} = E^{(0)}_{i} + \sum_{t}^{T-1}W^{(t)}\bm h^{(t)}_{i} + \mathcal{F} \left( W^{(T)} \bm{h}^{(T)}_{i} \right)
\end{equation}
where $W^{(t)}$ are learnable weights representing a linear combination of the features giving the body-ordered energy terms and $\mathcal{F} \left( W^{(T)} \bm{h}^{(T)}_{i} \right)$ is a generic nonlinear function accounting for the residual higher order terms in the truncated expansion.
The terms $\bm h^{(t)}_{i}$ have exactly correlation order $t$ (body-order $t+1$).

\begin{figure*}
    \centering
    \includegraphics[width=\linewidth]{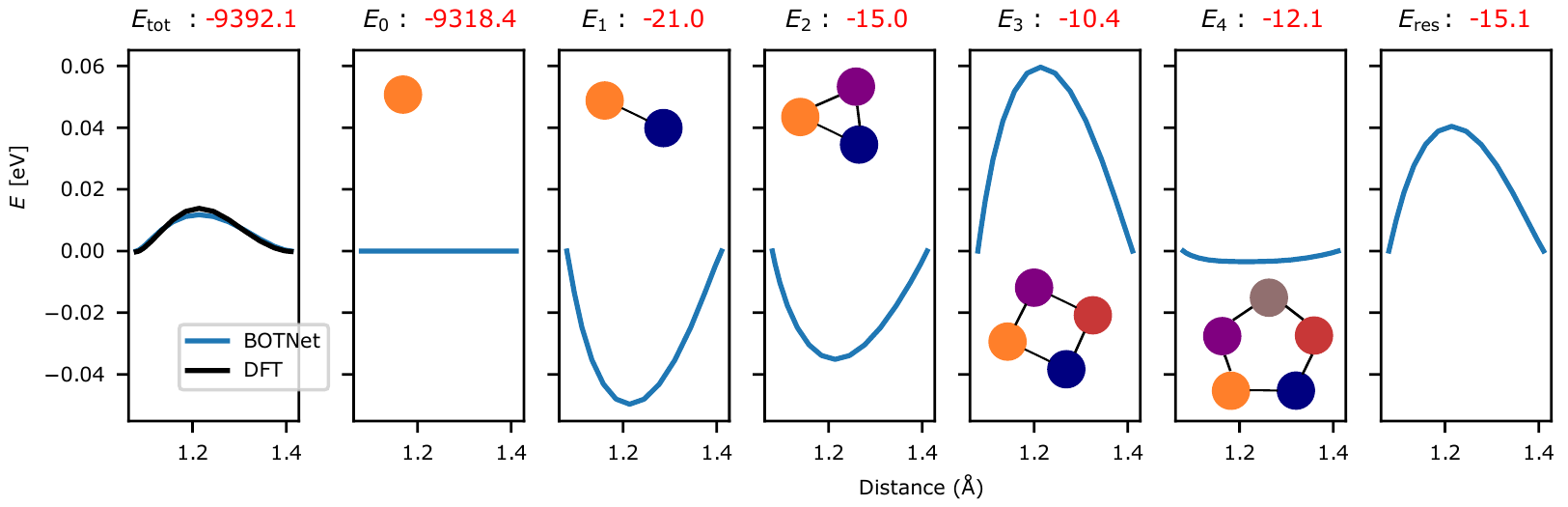}
    \caption{
        Decomposition of the total energy predicted by BOTNet for the H-transfer pathway in Acetylacetone.
        $E_{0}$ corresponds to the ``1-body'' atomic energies.
        The contributions from $E_{1}$ to $E_{4}$ represent energies of increasing body order (2-body to 5-body, respectively).
        The curves are shifted by the energy of the last configuration in the transfer path,
        which is annotated in red above each plot.
        All energies are in eV.
    }
    \label{fig:decomposition_botnet}
\end{figure*}

In Figure \ref{fig:decomposition_botnet}, we illustrate the hierarchical energy decomposition learned by a BOTNet model for the intramolecular hydrogen transfer reaction of acetylacetone.
In Eq.~\eqref{eqn:body-order}, the number of terms summed over grows with the correlation order, but for an efficient expansion the total size of each contribution should decrease.
In the case of BOTNet, we can observe that the terms are decreasing in absolute values (even after summing them for each correlation order over all centers).
The last term is bigger than the correlation order three and four since it is accounting for all higher-order terms in the truncated expansion.

\begin{figure*}
    \centering
    \includegraphics{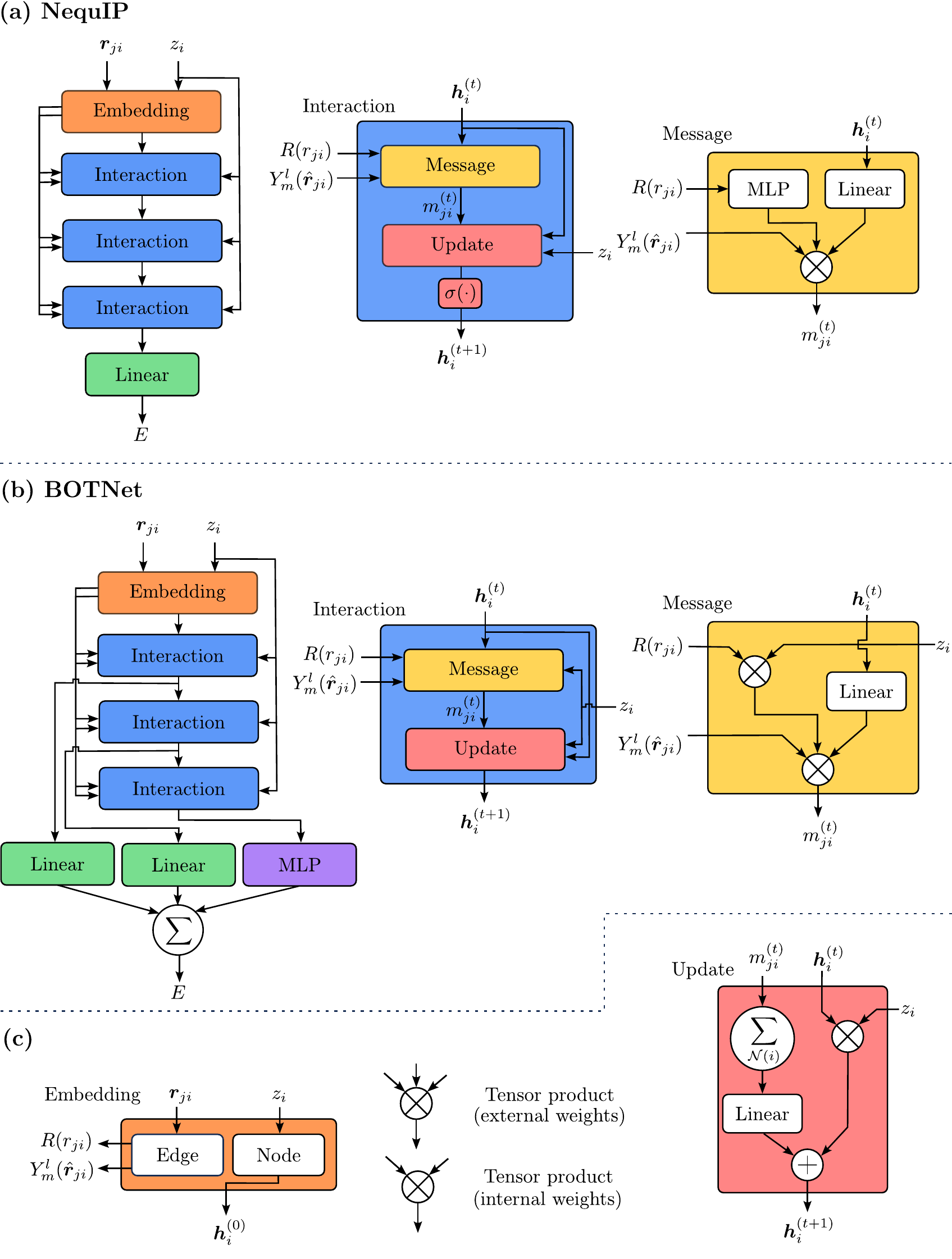}
    \caption{Illustration of the architectures of NequIP (a) and BOTNet (b).
    Panel (c) contains illustrations of components the architectures have in common.
    }
    \label{fig:architecture_figure}
\end{figure*}

Below, we give a detailed description of the architecture of BOTNet. It retains the most crucial elements of NequIP while introducing some new architectural features. 
The details of the two architectures are compared in Figure~\ref{fig:architecture_figure}.

\paragraph{Chemical embedding block}
The first block of both BOTNet and NequIP is the chemical embedding. The elements $z_{i}$ are mapped to vectors of lengths equal to the number of elements via one-hot embedding.
The one-hot vectors are multiplied with a learnable weight matrix of size $ N_\text{elements} \times N_\text{channels}$ outputting a learnable invariant feature vector for each atom corresponding to $\bm h^{(0)}_{i}$.

\paragraph{Radial embedding block}
The radial features are composed of a Bessel basis multiplied by a smooth polynomial cutoff denoted by $R_{n}(r_{ji})$.  The radial embedding block outputs an array of size $N_{n}$ for each edge corresponding to the values of Bessel functions of different frequencies. 

\paragraph{Interaction blocks}
The interaction block takes as input the node features $\bm h^{(t)}_{i}$, the radial features $R_{n}$, the spherical harmonics $Y_{l_1}^{m_1}$, and the node attributes $\bm \theta_{i}$. 
First, the node features $\bm h^{(t)}_{i}$ undergo a learnable linear transformation to mix the $k$ channels.
Then, the radial features $R_{n}$ are transformed together with the one-hot embedding $\bm \theta_i$ by a learnable bi-linear operation into the learnable radial basis, $R^{(t)}_{kl_{1}l_{2}L}(r_{ji})$.
The interaction block of the first layer differs from the rest of the layers in a simple way:
In the first layer, we use a standard MLP having $R_{n}$ as input and outputting $R^{(0)}_{kl_{1}l_{2}L}(r_{ji})$. 
In all subsequent layers we use a bilinear map combining the radial features $R_{n}$ and the chemical attribute $\bm \theta_{j}$ outputting the radial basis $R^{(t)}_{kl_{1}l_{2}L}(r_{ji})$ (see Section~\ref{subsub:radial_basis}).
The reason for having two different radial features (for the first and all subsequent layers) is that at the first layer, 
the attributes $\bm \theta_{j}$ are already present in the feature $\bm h^{(0)}_{j}$,
which makes the element-dependent radial basis redundant.

A symmetrized tensor product operation forms the edges features (one-particle basis) between the learnable radial basis, the spherical harmonics, and the node features. The symmetrized tensor product operation first makes the tensor product between the different elements and then decomposes it into irreducible representations using Clebsch-Gordan coefficients. The edges features are summed over the neighborhood of each atom to produce a message on each atom.

For the update phase of the message passing, we use a linear transformation followed by a simplified self-connection for the first layer and a residual self-connection for all the subsequent ones  (see Section~\ref{subsub:radial_basis}).

\paragraph{Readout blocks} 
After each update, a linear readout maps the invariant part of the learnable feature to the local state energy $E_{i}^{(t)} = \mathcal{R}^{(t)} (\bm h^{(t)}_{i})$. The last readout at iteration $T$, uses a nonlinear MLP to account for the higher orders terms in the truncated order expansion $E_{\text{res}}$.

In summary, there are no non-linearities present in the network to preserve body ordering, except at the last readout (see Section~\ref{sec:DesignSpace}). We show below that the inclusion of the higher-order term in the expansion results is sufficient to recover the accuracy of a fully nonlinear model. We introduce a new type of self-connection to preserve body order at the first round of message passing, ensuring that the network does not learn any non-body-ordered energy shift. We will also show that careful normalization induces a dramatic impact on extrapolation and is crucial for its in-domain accuracy as well.

To facilitate further exploration of the design space, we have implemented our model to allow for easy switching between the different design choices in the architecture of equivariant message passing models.
Due to the code's modular design, most of the different models mentioned in the analysis of the design space (Section~\ref{sec:DesignSpace}), including NequIP and BOTNet, can be accessed from the command line. 
The BOTNet code provides a modular framework for fast experimentation on the state-of-the-art equivariant message passing inter-atomic potentials and is available at \url{https://github.com/gncs/botnet}.

\section{Datasets}
\label{sec:Datset}

In this section, we briefly describe the datasets that were used in the computational experiments. The datasets are available at \url{https://github.com/davkovacs/BOTNet-datasets}.

\subsection{Ethanol and Methanol}

The ethanol and methanol dataset has two training sets. 
The first is taken from the revMD17 dataset~\cite{Christensen2020} and was sampled from a long 500 K \textit{Ab Initio} molecular dynamics trajectory. The models trained on this dataset can be evaluated on an independent test set coming from the same distribution, i.e., decorrelated parts of the same molecular dynamics trajectory.
Further, they can be tested for bond breaking extrapolation, by removing the hydrogen of the alcohol group and keeping the rest of the molecule fixed. Finally, we can test extrapolation by computing the energy change as atoms are displaced along a high- and a low-frequency normal mode from the optimal geometry.

The second training set contains the 1000 ethanol geometries of the first training set but is augmented by 300 methanol geometries also sampled from 500 K \textit{Ab Initio} molecular dynamics simulation.
The models trained using this mixed dataset can be used to analyse the 2-body component of the potentials.
Having two different molecules is required for this analysis because it eliminates the possibility for the models to distribute the total energy amongst the bonds arbitrarily 
by having two molecules where the ratio of the number of bonds between given element pairs is different.

\subsection{3BPA}

The 3BPA dataset contains snapshots of a large flexible drug-like organic molecule sampled from different temperature molecular dynamics trajectories~\cite{kovacs2021}. The models can be trained either on 300 K snapshots or on mixed T snapshots sampled from 300 K, 600 K, and 1200 K. There are three independent test sets for each temperature. The models can also be tested on the challenging task of computing the energy along dihedral rotations of the molecule. This test directly probes the smoothness and accuracy of the part of PES that determines which conformers are present in a simulation, and hence has a direct influence on properties of interest such as binding free energies to protein targets. In the following experiments, we train models on 500 configurations sampled at 300K only and test on the three temperatures.

\subsection{Acetylacetone}

The potential energy surface of acetylacetone has been studied exhaustively in the past due to its many interesting properties, such as the tunneling splitting of the intramolecular hydrogen transfer~\cite{BowmanAcAc2021}. In this paper, we are not trying to create the most accurate PES of this molecule but deliberately use a small training set of 500 configurations making the task particularly challenging. This helps us see the distinctions between the different models. To prepare the training set, we ran a long molecular dynamics simulation at 300 K using a Langevin thermostat at the semi-empirical GFN2-xTB level of theory~\cite{GFN2-xTB}.
We sampled configurations at an interval of 1 ps and re-computed the resulting set of configurations with density functional theory using the PBE exchange-correlation functional with D3 dispersion correction and def2-SVP basis set and \texttt{VeryTightSCF} convergence settings using the ORCA electronic structure package. To test the models, we measure extrapolation both in temperature and along two internal coordinates of the molecule, the hydrogen transfer path and a partially conjugated double bond rotation, which has a very high barrier for rotation.

\section{Software}
\label{sec:software}

Two different codes were used to conduct the experiments in Section~\ref{sec:DesignSpace}. In the tables below, the row labelled ``code'' indicates which was use to conduct each of the experiments.
All ``nequip'' labelled experiments were conducted with the NequIP software in version 0.5.4, which is available at \url{https://github.com/mir-group/nequip}. In addition, the \texttt{e3nn} library \cite{e3nn} was used under version 0.4.4, and PyTorch under version \texttt{1.10.0} \cite{paszke2019pytorch}.\\
All ``botnet'' labelled experiments were conducted with BOTNet software, which is availabe at \url{https://github.com/gncs/botnet}. In addition, version 0.3.2 of the \texttt{e3nn} library \cite{e3nn}and version \texttt{1.8.0} of the PyTorch~\cite{paszke2019pytorch} was used.
If the code is not explicitly specified (for example in the benchmark section),
we use the eponymous code for each model.

\section{Choices in the Equivariant Interatomic Potential Design Space}
\label{sec:DesignSpace}

To make the theory set out in Section~\ref{sec:EquiACE} and Section~\ref{sec:MultiACE_general} practically useful here, we analyse the design space of E(3)-equivariant interatomic potentials. 
Our discussion makes use of the Multi-ACE framework and provides a thorough analysis of the effects of the different design choices that can go into defining an equivariant interatomic potential. We unpack the most crucial ingredients of NequIP and BOTNet, show how the particular choices affect the models' performance in terms of in-domain accuracy and smooth extrapolation, and compare them to linear ACE, which is at a very different point of the framework.

\subsection{One-particle basis}
\label{sec:one_particle_basis2}

The one-particle basis is at the core of any message passing inter-atomic potential and was introduced in Section~\ref{sec:one_particle_basis1}.
In the most general case, the one-particle basis is denoted by $\phi_{kvL}^{(t)}(\sigma_{i}^{(t)},\sigma_{j}^{(t)})$ and was introduced in Eq.~\eqref{eqn:one-particle_general2}.
Below, we analyse some of the choices that can be made regarding the treatment of the chemical elements via the $T^{(t)}_{kcL}(\bm{h}_{j}^{(t)}, \bm{\theta_{i}},\bm{\theta_{j}})$ functions and the treatment of the radial basis $R^{(t)}_{kcl_1L}(r_{ji})$.

\subsubsection*{Treatment of the chemical elements}

The continuous embedding used in MPNNs is analogous to having \texttt{c} separate Linear ACE's, which are all sensitive to the chemical elements in a different learnable way. After each message passing step, the chemical element channels are mixed via a learnable transformation. It is interesting to note that the chemical identity of the neighboring atom (the sender) only enters directly at iteration $t=0$ when $\bm{h_j}^{(0)}$ is the one-hot embedding of the chemical elements afterward it is only indirectly dependent on the sender element via the output of the previous layer. 

In this section, we analyse the effect of increasing the number of uncoupled channels $k$, which corresponds to the dimension of the chemical element embedding. Table \ref{tab:chemical_embedding} compares NequIP models with increasing size of element embedding  $k$. The number of uncoupled (chemical) channels substantially affects the number of parameters. However, the scaling is nearly linear with the number of channels rather than power (equal to the correlation order) with the number of different elements, which would be the case if a discrete chemical element index and the complete linear ACE basis are used. It is also interesting to note that, as usual in deep-learning, over-parametrized models often achieve better results~\cite{OverparametrisedNN2020} not only in-domain (at low temperature) but also extrapolating out-of-domain (at high temperature).

\begin{table*}
    \caption{
        Root-mean-square error on 3BPA dataset for NequIP networks of different chemical embedding size.
        Energy (E, meV) and force (F, meV/\AA) errors of NequIP networks of increasing feature vector size, trained and tested on configurations of the flexible drug-like molecule 3-(benzyloxy)pyridin-2-amine (3BPA).
        All models were trained on 300K.
        All results were generated with the \texttt{nequip} code base.
    }
    \label{tab:chemical_embedding}
    \centering
    {%
        \begin{tabular}{llcccc}
            \toprule
            $N_\text{channels}$ &   & 16      & 32        & 64         & 128           \\
            \midrule
            No. of Parameters   &   & 437,336 & 1,130,648 & 3,415,832  & 11,580,440    \\
            \midrule
            \multirow{2}{*}{300 K}
                                & E & 3.7     & 3.1       & 3.0 (0.2)  & \textbf{2.9}  \\
                                & F & 12.9    & 11.9      & 11.6 (0.2) & \textbf{10.6} \\\hline
            \multirow{2}{*}{600 K}
                                & E & 12.9    & 12.7      & 11.9 (1.1) & \textbf{10.7} \\
                                & F & 32.1    & 30.3      & 29.4 (0.8) & \textbf{26.9} \\ \hline
            \multirow{2}{*}{1200 K}
                                & E & 48.6    & 49.5      & 49.8 (4.0) & \textbf{46.0} \\
                                & F & 104.2   & 101.6     & 97.1 (5.6) & \textbf{86.6} \\

            \bottomrule
        \end{tabular}
    }
\end{table*}

\begin{figure*}
    \centering
    \includegraphics[width=1.07\linewidth]{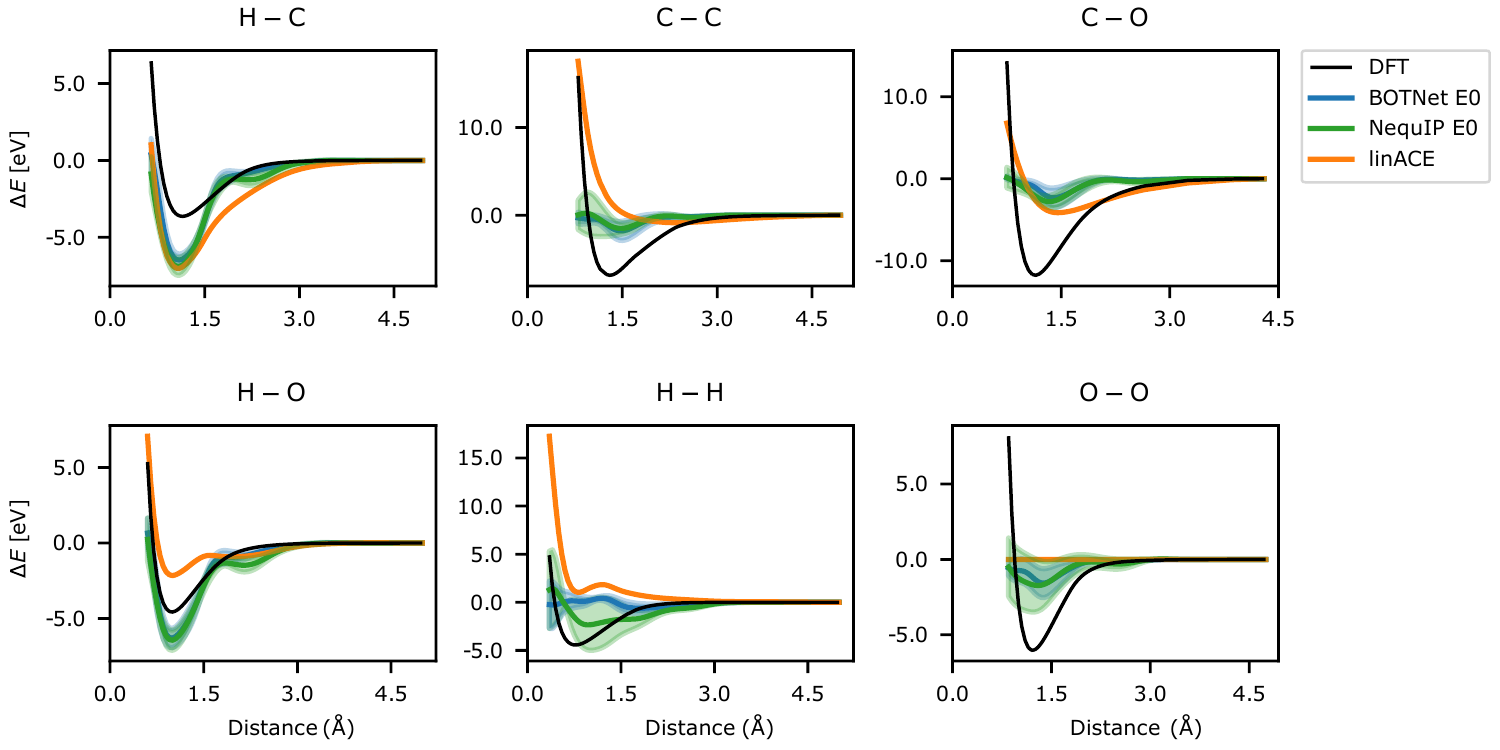}
    \caption{
        Dissociation of dimers from chemical elements presents in ethanol and methanol.
        All dimers (combinations) apart from O--O dimers are present in the dataset. 
        Shaded areas indicate one standard deviation computed over three runs.
    }
    \label{fig:dimers}
\end{figure*}

A further advantage of the element embedding approach is that it allows for some alchemical learning. The embeddings can learn a latent representation of the chemical elements and give meaningful predictions on combinations of elements that do not appear simultaneously in the training set. We have tested this alchemical learning by plotting the 2-body dimer dissociation of different chemical element combinations, as inferred from the ethanol and methanol dataset. This dataset contains complete molecules of ethanol and methanol. Both have a single oxygen atom; thus, the training set does not contain any configurations with two or more oxygen atoms. The dimers curves of linear ACE, NequIP, and BOTNet are shown in Figure~\ref{fig:dimers}. Our linear ACE implementation has no chemical embedding, and thus the O-O dissociation curve is identically zero. In contrast, NequIP and BOTNet predict the shape of the curves and the position of the minimum in a chemically sensible way. Since no dimers were in the training set, we do not expect to recover these dissociation curves with accuracy. However, the general shape and particularly the repulsive interaction for small interatomic distances is essential for obtaining stable molecular dynamics.

\subsubsection*{Radial basis}
\label{subsub:radial_basis}

There is much freedom in choosing a functional form for the radial basis $ R^{(t)}_{kcl_1L}(r_{ji})$.
In the context of atom density-based atomic environment representations such as SOAP~\cite{Csanyi2013SOAP}, ACSF~\cite{Behler2007ACSF} and the bispectrum (SNAP)~\cite{THOMPSON2015SNAP}.
The importance of the radial basis has been long known, and many strategies for improving it have been developed.
It has been shown that most of these representations only differ by their choice of radial basis \cite{ACE_ralf}.
Adopting the best radial basis has been a continuous source of improvement for models in the past. For example, in the case of SOAP, improving the radial basis leads to more efficient, smoother, and faster models~\cite{Miguel2019TurboSOAP, Ceriotti2021Librascal, Foster2020Dscribe}.

The most straightforward choice for a radial basis, used, for example, by linear ACE, is a set of fixed orthogonal polynomial basis functions that are the same for each chemical element and do not depend on $l$ of the spherical harmonics.
The dependence on the atom types enters only via the distance transform.
This distance transform scales the interatomic distances to be in the domain of the orthogonal radial basis. Its form can be dependent on the chemical elements of the two atoms accounting for the differences in atomic radii.

Recently, much work has shown that it can be advantageous to optimize the radial basis in a data-driven way.
This can be done \emph{a priory}~\cite{CeriottiOptRadBasis2021}, or can be optimized during the training of the model \cite{bochkarev_efficient_2022}.

NequIP for example uses a multi-layer perceptron to implement a learnable radial basis that is dependent on the tuple $(k,l_{1},l_{2},L)$,
where $k$ corresponds to the uncoupled channel index,
$l_{1}$ to the representation of the spherical harmonics $Y^{m_1}_{l_1}(\bm{\hat{r}}_{ji})$,
$l_{2}$ to the representation of the equivariant message $h^{(t)}_{j,l_{2}m_{2}}$,
and there is a different radial basis for each output symmetry $L$:
\begin{equation}
    R^{(t)}_{kl_{1}l_{2}L}(r_{ji})  = \text{MLP} \left( R_{n}(r_{ji})f_\text{cut}(r_{ji}) \right),
\end{equation}
where MLP is a multi-layer perceptron.
Typically, the number of layers used in this MLP is three.
$R_n$ are a set of Bessel basis polynomials and $f_\text{cut}(r_{ji})$ is a cutoff function such that
$\lim_{r_{ji} \to 0} f_\text{cut}(r_{ji}) = 0$,
but orthogonality of the different basis functions is not enforced.
This type of radial basis allows for improved flexibility in spatial resolution when combining features of different symmetries.
We refer to this radial basis as \textit{element agnostic radial basis} as it is independent of the chemical elements.

BOTNet uses a similar learnable radial basis but it is also dependent on the sender atom chemical element. This is achieved by forming radial basis functions with the $k$ multi-index running over $N_\text{channels}=N_\text{embedding}\times N_\text{elements}$. This means that BOTNet will have a separate radial basis in each chemical embedding channel for each neighbour chemical element, and the $T_{kc}$ function will pick up the appropriate one via its dependence on $\theta_{j}$ (see \ref{eqn:one-particle_general2}). This radial basis can be written as
\begin{equation}
    \label{eq:BOTNet_radial}
    R^{(t)}_{kl_{1}l_{2}L}(r_{ji})  = \sum_{n}W_{kn(l_1l_2L)}^{(t)} R_{n}(r_{ji}) f_\text{cut}(r_{ji}),
\end{equation}
where $W_{kn(l_1l_2L)}^{(t)}$ is an array of weights of dimensions $[N_\text{ channels},N_\text{basis},N_\text{paths}]$, 
with $N_\text{basis}$ being the number of Bessel basis function $R_n$
and $N_\text{paths}$ 
being number of combination of products of a given symmetry between the equivariant feature $h_{j,l_2m_2}$ and the spherical harmonics.
We refer to this type of radial basis as \emph{element dependent radial basis} because it explicitly depends on the chemical element of atom $j$ via the weight array.

We have observed that \textit{element dependent radial basis} gives better training and validation accuracy.
However, for extreme extrapolation like bond breaking, we have found that the \emph{agnostic radial basis} is a better choice in particular with the correct normalization,
as discussed in Section \ref{sec:normalization}.

\subsection{Non-linear Activations}
\label{sec:non_linearity}

The body ordering, as defined in Section~\ref{sec:body_order}, is a central property of classical force fields and has proven to be a very successful approximation of quantum mechanical systems~\cite{DUSSON2022}. The Linear version of ACE is body-ordered by construction, but most other ML approaches do not have this structure. Having body-ordered models was thought to be beneficial because it enforces the learning of low-dimensional representations of the data, which is an excellent inductive bias for better extrapolation. In the following, we analyse the effect of different nonlinear activations and their effect on body ordering.

The ACE message passing equation on Eq.~\eqref{eq:multi_ace_message} is a nonlinear operation and is fundamentally related to the tensor product of the $O(3)$ group. The effect of this tensor-product non-linearity is to increase the body-order of each layer by $\nu$. Most previously published MPNN architectures have $\nu=1$. Beyond the tensor-product, it is possible to include other types of non-linearities in the update function $U_{t}$ of Eq.~\eqref{eq:multiACE_update} by taking $U_t = \left( \sigma_{i}^{(t)},\bm{m}_{i}^{(t)} \right) = \mathcal{F} \left( W^{(t)}\bm{m}_{i}^{(t)} \right)$ where  $\mathcal{F}$ is a generic nonlinear function and $W^{(t)}$ is a $[N_\text{channels} \times N_\text{channels} \times (L_\text{max} + 1)]$ learnable weight matrix linearly mixing the uncoupled channels $k$. It is important to note that a general nonlinear function $\mathcal{F}$ when applied to equivariant features does not preserve equivariance. A common strategy is to use gated equivariant non-linearities which are summarized in Appendix~\ref{sec:non_linear_equivariant}. In the following when we compare non-linearities the models only differ in the choice of the non-linearities applied to the invariant parts of the models, the equivariant non-linearities are always kept the same.

Suppose the model is explicitly body-ordered and equivariant, then only a smaller subset of non-linearities can be used that preserve the equivariance. The central remark is that a non-linearity preserves body ordering if it admits a finite Taylor expansion. A detailed example showing how the SiLU non-linearity destroys the body-ordered structure is in Appendix~\ref{sec:SiLU_body_order}. Two types of non-linearities preserve the body-ordered structure; the first is previously known as the kernel trick and consists of using non-linearities with a finite Taylor expansion such as the squared-norm to raise the body order of the representation~\cite{Csanyi2013SOAP}.

The approach taken in designing BOTNet was to create a body-ordered model during the first five message passing layers by removing all nonlinear activations from the update but making the last readout nonlinear with an infinite body order. This way, the last readout function is responsible for representing the residual of the body-order expansion not captured by the first five layers. This energy decomposition enforces the learning of low-dimensional structures because the low body-order part of the energy appears explicitly. The corresponding energy expansion of BOTNet is:
\begin{equation}
    E =   E^{(0)} + \sum_{i=1}^{N}E_{i}^{(0)}(\bm r_{i}) + \sum_{1 \leq i < j \leq N}^{N}E_{i,j}^{(1)}(\bm r_{i},\bm r_{j}) + \dots + E_{\text{res}},
\end{equation}
where $E_{\text{res}} = \mathcal{F}(\bm m^{(T)}(\bm r_{i_{1}},...,\bm r_{i_{n}}))$ is a general nonlinear term that accounts for all the missing contributions not captured by the previous body-ordered layers.

The models using different non-linearities are compared in Table~\ref{tab:non_linear_3BPA}. It is clear from the table that in the case of NequIP, the choice of non-linearity is crucial; using \texttt{tanh} instead of \texttt{SiLU} makes the results significantly worse, probably because of the \texttt{tanh} function having 0 gradient for large positive and negative inputs which makes the optimization difficult due to vanishing gradients~\cite{SiLU2017}. This makes models with \texttt{tanh} non-linearity even worse than not using any non-linearities at all (other than the tensor-product). In the case of BOTNet, we can see that adding a nonlinear layer to a strictly body-ordered model to account for the higher-order terms in the truncated body ordered expansion significantly improves the results. The Normalization row indicates the type of data normalization used for the experiments. For further details, refer to the Section \ref{sec:normalization} on normalization.

\begin{table*}
    \caption{
        Root-mean-square Energy (E, meV) and force (F, meV/\AA) error on 3BPA dataset for different choice of nonlinear and linear models. 
        Version of NequIP coded in BOTNet. Models on the same side of the vertical line use equivalent internal normalization. Models on the left use $\lambda = \sqrt{\big \langle \#\mathcal{N}(k) \big \rangle_k}$ and models on the right use $\lambda = \big \langle \#\mathcal{N}(k) \big \rangle_k$ (cf \ref{sec:internal_normalization}) . Linear models refer to model without any nonlinear activation.
        }
    \label{tab:non_linear_3BPA}
    \centering
    {%
        \begin{tabular}{llcc | ccc}
            \toprule
            Model         &   & NequIP Tanh    & NequIP Silu        & NequIP Linear  & BOTNet Linear  & BOTNet               \\
            \midrule
            Code          &   & botnet         & nequip             & botnet         & botnet         & botnet               \\
            \midrule
            Normalization &   & SSH forces-rms & SSH forces-rms     & SSH forces-rms & SSH forces-rms & SSH forces-rms       \\
            \midrule
            \multirow{2}{*}{300 K}
                          & E & 4.8            & \textbf{3.0} (0.2) & 3.7            & 3.3            & 3.1 (0.13)           \\
                          & F & 18.5           & 11.6 (0.2)         & 13.9           & 12.0           & \textbf{11.0} (0.14) \\ \hline
            \multirow{2}{*}{600 K}
                          & E & 20.1           & 11.9 (1.1)         & 15.4           & 11.8           & \textbf{11.5} (0.6)  \\
                          & F & 42.5           & 29.4 (0.8)         & 34.1           & 30.0           & \textbf{26.7} (0.29) \\ \hline
            \multirow{2}{*}{1200 K}
                          & E & 75.7           & 49.8 (4.0)         & 61.92          & 53.7           & \textbf{39.1} (1.1)  \\
                          & F & 156.1          & 97.1 (5.6)         & 109.5          & 97.8           & \textbf{81.1} (1.5)  \\

            \bottomrule
        \end{tabular}
    }
\end{table*}

\subsection{Self-Connection}
\label{sec:self_connection}

An essential and often neglected part of MPNN models is the self-connection. It is a mechanism used to mix information from the previous layer with the output of the current layer in a learnable way. The self-connection mechanism is fundamentally related to the residual architecture of convolutional neural networks~\cite{resnet2015}.

In NequIP the general message passing operation of Eq.~\eqref{eq:multi_ace_message} is chosen to be independent of the receiver (central) atom chemical element $\theta_{i}$, as shown in Eq.~\eqref{eq:NequipConv}.
The effect of this is that the successive message passing iterations ``dilute'' the chemical information of the central atom. NequIP has introduced a self-connection that re-injects chemical information about the central atom after each message passing step to overcome this issue. It is part of the update and has the form reminiscent of residual neural networks:
\begin{equation}
    \label{eq:NequIP_self_connection}
    h^{(t+1)}_{i,kLM} =  h^{(t + 1)}_{ikLM} + \sum_{a\tilde{k}} W_{k\tilde{k}La}\theta_{i,a}h^{(t)}_{i,\tilde{k}LM} ,
\end{equation}
where $W_{k\tilde{k}La}$ is a learnable weight matrix of size $[N_\text{channels} \times N_\text{channels} \times L_\text{max} \times N_\text{elements}]$ of the attribute $\theta_{i}$ which is in the case of NequIP the one hot encoding of the chemical type $z_{i}$ of the central (receiver) atom.

When the residual update of Eq.~\eqref{eq:NequIP_self_connection} is applied after the very first message passing iteration, during the training, the initial feature $h^{(0)}_{i}$ which is independent of the atomic environment, gets updated. This is because the network can learn a shift to the potential energy that is only dependent on the central atom. If one wants to ensure that the model has the correct limit for isolated atoms, this self-connection cannot be applied at the first update. An alternative simplified self-connection, implemented as a bi-linear map, serves the purpose of reinjecting chemical information but does not have the residual connection:
\begin{equation}
    h^{(t+1)}_{i,kLM} =   \sum_{a\tilde{k}} W_{k\tilde{k}La}\theta_{i,a}h^{(t)}_{i,\tilde{k}LM}
\end{equation}

This simplified self-connection has the advantage that the features at the $t=0$ layer do not enter the energy expression removing the learnable shift. This is advantageous if it is necessary to enforce that the model predicts the correct energy for isolated atoms. Table~\ref{tab:sc_3BPA} shows a comparison of NequIP and BOTNet models with the residual and simplified self-connection. It appears that the self-connection plays a crucial role in message-passing accuracy. Moreover, using the residual self-connection, the models can perform significantly better than no residual architecture. The \texttt{mix sc} BOTNet model has the simplified self-connection in the first update followed by the residual one. This architecture does not have an internal learnable shift and can match the performance of the entirely residual architecture closely.
The issue of self connections does not arise in the case of linear ACE because the chemical elements explicitly index the basis functions, and there is only a single message passing operation.

\begin{table*}
    \caption{
        Root-mean-square Energy (E, meV) and force (F, meV/\AA) error on 3BPA dataset of NequIP without self connection (``no sc'') and with a fully residual self connection (``residual sc''), and BOTNet model with a fully residual self connection, only the simplified self connection (``simplified sc'') and with the use of a simplified self connection at the first layer and fully residual self connection in subsequent layers (``mixed sc''). As BOTNet and NequIP differ in many training settings, we bold both sides.
    }
    \label{tab:sc_3BPA}
    \centering
    {%
        \begin{tabular}{llcc|ccc}
            \toprule
            Model         &   & NequIP residual sc   & NequIP no sc   & BOTNet residual sc & BOTNet mixed sc      & BOTNet simplified sc \\
            \midrule
            Code          &   & nequip               & nequip         & botnet             & botnet               & botnet               \\
            \midrule
            Normalization &   & SSH forces rms       & SSH forces rms & SSH forces rms     & SSH forces rms       & SSH forces rms       \\
            \midrule
            \multirow{2}{*}{300 K}
                          & E & \textbf{3.0} (0.2)   & 3.8 (0.1)      & \textbf{3.02}      & 3.1 (0.13)           & 3.7                  \\
                          & F & \textbf{11.6} (0.2)  & 15.8 (0.6)     & 11.7               & \textbf{11.0} (0.14) & 13.7                 \\ \hline
            \multirow{2}{*}{600 K}
                          & E & \textbf{11.9}  (1.1) & 17.1 (1.1)     & 12.3               & \textbf{11.5} (0.6)  & 14.8                 \\
                          & F & \textbf{29.4} (0.8)  & 47.8 (2.9)     & 27.4               & \textbf{26.7} (0.29) & 37.1                 \\ \hline
            \multirow{2}{*}{1200 K}
                          & E & \textbf{49.8} (4.0)  & 108.5 (5.8)    & 43.5               & \textbf{39.1} (1.1)  & 81.4                 \\
                          & F & \textbf{97.1} (5.6)  & 225.5 (14.0)   & \textbf{79.9}      & \textbf{81.1} (1.5)  & 126.93               \\

            \bottomrule
        \end{tabular}
    }
\end{table*}

\subsection{Numerical stability}
\label{sec:num_stability}

\begin{figure}[!ht]
    \centering
    \includegraphics[width=0.8\linewidth]{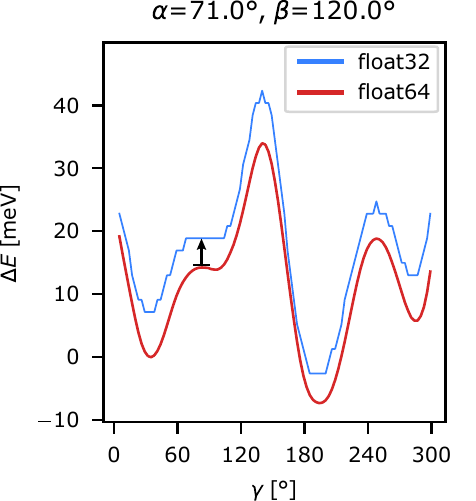}
    \caption{Comparison of cuts of the 3BPA potential energy surface between NequIP models trained using 32 bit floats (red) and 64 bit floats (blue). The 32 bit curve was shifted upwards as indicated by the black arrow for better visibility. 
    }
    \label{fig:float32vsfloat64}
\end{figure}

Numerical stability is of significant importance for computations involving interatomic potentials. It affects the smoothness of the PES and, consequently, the stability of geometry optimisation and accuracy of molecular dynamics simulations. In Figure~\ref{fig:float32vsfloat64} we show a potential energy slice as one of the bond-angles is varied in the 3BPA molecule. The figure shows NequIP models trained using 32 and 64-bit floats. Using the lower precision results in a piecewise linear unsmooth potential energy surface. By using higher precision, the smoothness of the potential energy surface is significantly improved. The same phenomenon was observed with BOTNet.

\section{Normalization}
\label{sec:normalization}

This section inspects the influence of normalization both inside the network and on the data. We show that normalization plays a significant role in converging these over-parametrized models that rely on stochastic gradient estimation. However, non-physical normalization can hurt the extrapolation of the models far away from the training set.

\subsection{Internal Normalization}
\label{sec:internal_normalization}

Internal normalization refers to all procedures applied to internal features and weights to make them respect some statistical properties.
It is of crucial importance in the convergence of stochastic gradient-based optimization,
and one of the most used examples is batch-normalization \cite{ioffe2015batch}.
The first type of internal normalization is that of learnable features.
In NequIP and BOTNet, spherical harmonics are normalized such that the second moment of features inside the network is close to $1$:
\begin{equation}
    || {\bm Y}_{l}(x) ||^{2} = 2l+1, \quad x \in S^{2}.
\end{equation}

The learnable features at each stage should also follow the same statistical property at initialization
\begin{equation}
    \left\langle h^{(t)}_{i,kLM} \right\rangle^{2} \approx 1.
\end{equation}
The underlying motivation for this normalization \cite{e3nn} comes from the assumption that the weights obey
\begin{align}
    \left\langle w_{j} \right\rangle      & = 0                       \\
    \left\langle w_{j}w_{k} \right\rangle & = \sigma^{2} \delta_{ij},
\end{align}
such that the two first moments of the product $h^{(t)}_{i} \cdot w$
are functions of $\left\langle h^{(t)}_{i,kLM} \right\rangle ^{2}$ only,
\begin{align}
     & \left\langle h^{(t)}_{i} \cdot w \right\rangle = \sum_{kLM} \left\langle h^{(t)}_{i,kLM} \right\rangle \left\langle w_{kLM} \right\rangle = 0 \\
     & \left\langle \left(h^{(t)}_{i} \cdot w \right)^{2} \right\rangle = \sigma^{2} \sum_{kLM} \left\langle h^{(t)}_{i,kLM} \right\rangle^{2}.
\end{align}

Another crucial normalization is the message-normalization.
As the message uses a sum operation, it gathers an average of neighbors' features.
NequIP \cite{nequip} proposed to normalise the sum by the square root of the average number of neighbors,
so that $\lambda = \sqrt{\big \langle \#\mathcal{N}(k) \big \rangle_k}$ in Eq.~\eqref{eqn:message_func}.
We found that, for BOTNet, dividing the sum by the average number of neighbors $\left\langle \#\mathcal{N}(k) \right\rangle_k$ across the training dataset yields the best results.

In Table \ref{tab:message_norm}, we compare NequIP and BOTNet models with and without the message normalization. We observe a significant effect of the message normalization on the performance,
especially at high temperatures, being responsible for a decrease in the error of over $30\%$.
As models with and without this normalization have the same expressiveness, they only differ in their learning dynamics during optimization.
These results highlight how crucial internal normalization is for the convergence of stochastic gradient optimization.

\begin{table*}
    \caption{
        Root-mean-square Energy (E, meV) and force (F, meV/\AA) error on 3BPA dataset of NequIP and BOTNet models with different internal normalization.}
    \label{tab:message_norm}
    \centering
    {%
        \begin{tabular}{llccc | cc}
            \toprule
            Model         &   & NequIP no internal & NequIP $\sqrt{\big \langle \#\mathcal{N}(k) \big \rangle_k}$ & NequIP $\big \langle \#\mathcal{N}(k) \big \rangle_k$ & BOTNet no internal & BOTNet  $ \big \langle \#\mathcal{N}(k) \big \rangle_k$ \\
            \midrule
            Code          &   & botnet             & nequip                                                       & botnet                                                & botnet             & botnet                                                  \\
            \midrule
            Normalization &   & SSH forces rms     & SSH forces rms                                               & SSH forces rms                                        & SSH forces rms     & SSH forces rms                                          \\
            \midrule
            \multirow{2}{*}{300 K}
                          & E & 3.3                & 3.0 (0.2)                                                    & \textbf{2.8}                                          & 3.5                & 3.1 (0.13)                                              \\
                          & F & 12.4               & 11.6 (0.2)                                                   & \textbf{10.8}                                         & 13.2               & 11.0 (0.14)                                             \\ \hline
            \multirow{2}{*}{600 K}
                          & E & 12.6               & 11.9 (1.1)                                                   & \textbf{10.6}                                         & 15.0               & 11.5 (0.6)                                              \\
                          & F & 33.3               & 29.4 (0.8)                                                   & 26.8                                                  & 38.8               & \textbf{26.7} (0.29)                                    \\ \hline
            \multirow{2}{*}{1200 K}
                          & E & 54.6               & 49.8 (4.0)                                                   & 43.1                                                  & 89.6               & \textbf{39.1} (1.1)                                     \\
                          & F & 117.6              & 97.1 (5.6)                                                   & 85.5                                                  & 138.5              & \textbf{81.5} (1.5)                                     \\

            \bottomrule
        \end{tabular}
    }
\end{table*}

\subsection{Data Normalization}
\label{sec:data_normalization}

Data normalization is widely used in many areas of deep learning to accelerate the convergence of the optimization \cite{Lecun98efficientbackprop}. We define data normalization as a general transformation of the data prior to training. In the context of machine learning interatomic potentials, normalization can also play a unique role by constraining the data to obey correct physical limits, for example in the case of dissociation to atoms.

Given a data set of energies $\mathcal{D_{E}} = \left\{ E^{i} \right\}^{N}_{i=1}$
and forces $\mathcal{D_{F}} = \{F^{i,j}\}^{N,K}_{i=1}$, where $j$ is a multi-index 
running over $K = {N_\text{atoms} \times N_\text{coordinates}}$,
the normalization operation is a function $\Phi: E \mapsto \hat{E} $ and by the conservation principle $\Phi': F \mapsto \hat{F}$, 
which ensures that the transformed data $\mathcal{\hat{D}_{E}}$ and $\mathcal{\hat{D}_{F}}$ has some statistical properties (statistical normalization),
or in the case of interatomic potentials one might want these transformations to obey certain physical properties (physical normalization) such as correct limit for isolated atoms. 

The most widely used normalization schemes is standardization that we refer to as scale shifting (SSH),
transforming the data as,
\begin{align}
    \label{eq:ssh}
     & \hat{E}  = \frac{1}{\alpha}  \left( E - \mathbb{E}_{\mathcal{D_{E}}}(E) \right) \\
     & \hat{F}  = \frac{1}{\alpha} F,
\end{align}
with $\mathbb{E}_{\mathcal{D_{E}}}(E)$ is the average of the energies across the training set and $\alpha$ can be chosen to be either the root mean square of the forces across the dataset or the standard deviation of the energies across the dataset.
This ensures that the target energies have zero-mean and unit variance meaning that $\mathbb{E}_{\mathcal{\hat{D}_{E}}}(E) = 0$.
This normalization scheme has the property that the models have a non-physical offset of the potential energy surface.
This means that the arbitrary shift of the potential energy does not correspond to the energy of the isolated atoms.
This does not affect the simulations as long as no dissociation to atoms is involved, for example, in bulk simulations, but can be problematic for reactive force fields.

The physical normalization can be written as:
\begin{align}
     & \hat{E}  = \frac{1}{\alpha} \left( E - \sum_{i = 1}^{N} E_{0, Z_i} \right) \\
     & \hat{F}  = \frac{1}{\alpha} F,
\end{align}
where $N$ is the number of atoms in the molecule,
$\alpha$ is a scaling factor (can be interpreted as a change of units),
and $E_{0,Z_i}$ is the atomic energy of the chemical element $Z_i$.
This approach ensures that the dissociated limit with no-interaction energy is correct.

\begin{table*}
    \caption{
        Root-mean-square Energy (E, meV) and force (F, meV/\AA) error on 3BPA dataset of NequIP and BOTNet models with different data normalization.
    }
    \label{tab:data_normalization_3BPA}
    \centering
    {%
        \begin{tabular}{llcc|ccc}
            \toprule
            Model         &   & NequIP  & NequIP              & BOTNet (Element Dependent) & BOTNet (Agnostic) & BOTNet               \\
            \midrule
            Code          &   & nequip  & nequip              & botnet                     & botnet            & botnet               \\
            \midrule
            Normalization &   & $E_{0}$ & SSH forces rms      & $E_{0}$                    & $E_{0}$           & SSH forces rms       \\
            \midrule
            \multirow{2}{*}{300 K}
                          & E & 3.5     & \textbf{3.0} (0.2)  & 3.6                        & 3.5               & \textbf{3.1} (0.13)  \\
                          & F & 13.0    & \textbf{11.6} (0.2) & 14.1                       & 13.4              & \textbf{11.0} (0.14) \\ \hline
            \multirow{2}{*}{600 K}
                          & E & 13.4    & \textbf{11.9} (1.1) & 14.6                       & 15.7              & \textbf{11.5} (0.6)  \\
                          & F & 33.7    & \textbf{29.4} (0.8) & 35.7                       & 33.5              & \textbf{26.7} (0.29) \\ \hline
            \multirow{2}{*}{1200 K}
                          & E & 55.1    & \textbf{49.8} (4.0) & 53.7                       & 44.03             & \textbf{39.1} (1.1)  \\
                          & F & 107.9   & \textbf{97.1} (5.6) & 111.4                      & 101.7             & \textbf{81.5} (1.5)  \\

            \bottomrule
        \end{tabular}
    }
\end{table*}

\begin{figure*}[t]
    \centering
    \includegraphics[width=1.05\linewidth]{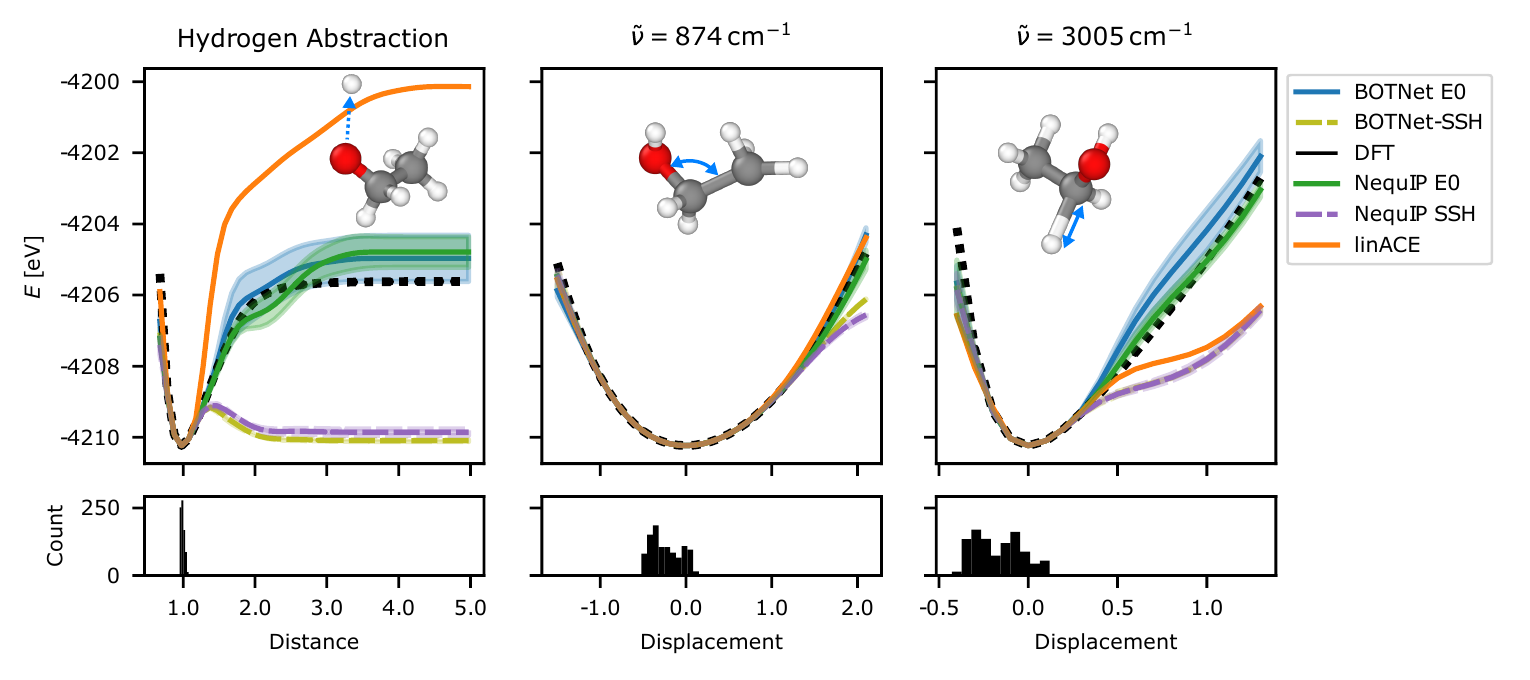}
    \caption{
        Extrapolation capabilities of multiple models trained on the ethanol subset of the rMD17 dataset.
        Models that predict standardized energies (and forces) are shown with dashed lines. Note that linear ACE predicts non-standardized energies and forces.
        The ground-truth DFT energy is shown in black.
        \textbf{(a)}:
        predicted energy for a range of O--H bond lengths (cf. dashed blue arrow) while the remainder of the ethanol molecule is kept fixed.
        The bottom panel shows the distribution over the O--H bond lengths in the training dataset.
        \textbf{(b)} and \textbf{(c)}:
        predicted energy for molecular structures corresponding to displacements along two normal modes starting from the equilibrium structure (cf. blue arrows).
    }
    \label{fig:ethanol}
\end{figure*}

In the following, we are testing the effect of data normalization on both accuracy and extrapolation capabilities. We compare the performance of models with different data normalization in table \ref{tab:data_normalization_3BPA} on the 3BPA data set. We observe that models learning from scale-shifted data achieve the best accuracy, including at higher temperatures. This difference stems from the radically different learning tasks between the scale-shifted and the physically normalized models. The SSH models are learning to reproduce a narrow part of the potential energy surface near equilibrium. In contrast, the physically normalized models are constrained to obey limits far from the data distribution.

The right panel of Figure~\ref{fig:ethanol} shows the energy predicted by each of the models and the ground truth DFT as the O-H bond distance is varied with the position of the other atoms being fixed. The configuration space sampled in the training set along the O-H bond is very narrow, making this a challenging extrapolation task. As the H is moved far from the oxygen, the model has to predict the energy of the ethyl radical, which is not in the training set, and therefore predicting the exact energy is almost impossible. Nonetheless, it is still valuable to compare how smooth and physical the shape of the PES is.
Overall, the models reach good fidelity to the DFT results near the equilibrium where most of the training data is. As expected, the scale-shifted models (NequIP SSH and BOTNet SSH) give nonphysical results far from the training set. The barrier height is about 1~eV which is largely underestimated compared to the DFT, which has a barrier of 4eV. The models with correct atomic energy at the limit (NequIP E0, BOTNet E0, and linACE) give a much better barrier and can predict remarkably accurately the energy of the radical. NequIP E0 and BOTNet E0 reproduce the potential energy surface between 1~\AA{} and 2~\AA{ } with very high fidelity while there is no data in this area.

The two right panels of Figure~\ref{fig:ethanol} show the potential energy surface of ethanol as the atoms are moved along a low frequency ($874 cm^{-1}$) and a high frequency ($3005 cm^{-1}$). The low-frequency mode probes a C-C bending mode with no bond breaking. We see that both scale-shifted and E0 models do equally well at the task, likely due to the absence of bond breaking.
The high-frequency mode probes a C-H stretching mode, eventually getting close to bond breaking. We observe that scale-shifted models with the wrong limit predict the C-H stretching less accurately. In contrast, BOTNet and NequIP models with the correct limit can accurately predict the potential energy surface up to bond breaking, confirming that inputting the correct limit is crucial for reactive interatomic potentials.

\section{Benchmark Experiments}
\label{sec:Benchmark}

This section shows the performance of the equivariant graph neural network models in the broader context, comparing them to the earlier approaches. The results show that this family of methods is, on average, at least a factor of two more accurate than the kernel, linear, or feed-forward neural network methods when applied to the potential energy surface of organic molecules. Moreover, we show that BOTNet and NequIP achieve similar accuracy on a wide range of benchmarks.

\subsection{rMD17: Small molecules benchmark}
The revMD17 dataset contains five different train test splits of 10 different small organic molecules~\cite{Christensen2020}. Each of the splits comprises 1000 configurations for each molecule sampled randomly from a long \textit{Ab initio} molecular dynamics simulation carried out at 500 K. The task is to fit a force field model on the 1000 examples and report the mean absolute error (MAE) of the total energy and the force components. Table~\ref{tab:md17} on the left of the vertical solid line shows the MAE of different models from the literature trained on precisely this dataset. We show the models on the right of the solid vertical line for completeness, but it is important to note that they were trained on an earlier version of this dataset which proved to be noisy, with different train-test splits and DFT settings. Both BOTNet and NequIP models reach state-of-the-art accuracy, far outperforming the other approaches in all the molecules. We observe similar accuracy between BOTNet and NequIP across most of the molecules.

\begin{table*}
    \caption{
        Mean absolute error on rMD17 dataset.
        Energy (E, meV) and force (F, meV/\AA{}) errors of different models trained on 1,000 samples.
        The models on the left were trained and tested using the same train-test splits of rMD17 except for NequIP which uses 1,000 configuration sampled from the full rMD17 dataset~\cite{Christensen2020}, whereas models on the right use the original MD17~\cite{chmiela2017machine}.
        The best model for each molecule (on the left and the right) are shown in bold font.
        For reference 43 meV = 1 kcal / mol.
    }
    \centering
    \label{tab:md17}
    \resizebox{\textwidth}{!}{
        \begin{tabular}{@{}llccccccc | ccccc@{}}
            \toprule
             &   & BOTNet & NequIP & Linear ACE~\cite{kovacs2021} & sGDML~\cite{kovacs2021} & FCHL~\cite{Faber2018FCHL} & GAP~\cite{DeringerCsanyi2021ChemRev} & ANI~\cite{Isayev2020ANI} & PaiNN~\cite{Schutt2021Painn} & GMsNN~\cite{zaverkin2020} & DimeNet~\cite{DimeNet} & NewtonNet~\cite{haghighatlari2021newtonnet} & SchNet~\cite{schnet} \\
            \midrule
            \multirow{2}{*}{\textbf{Aspirin}}
             & E & \textbf{2.3}    & \textbf{2.3}    & 6.1                                   & 7.2            & 6.2                               & 17.7                        & 16.6                                      & 6.9                                  & 16.5                              & 8.8                            & 7.3 & 16.0                                                                                  \\
             & F & 8.5             & \textbf{8.2}    & 17.9                                  & 31.8           & 20.9                              & 44.9                        & 40.6                                       & 16.1                                 & 29.9                              & 21.6                           & 15.1    & 58.5                                                                           \\
            \midrule
            \multirow{2}{*}{\textbf{Azobenzene}}
             & E & \textbf{0.7}    & \textbf{0.7}    & 3.6                                   & 4.3            & 2.8                               & 8.5                         & 15.9                                      & -                                    & -                                 & -                              & 6.1 & 3.5                                                                             \\
             & F & 3.3             & \textbf{2.9}    & 10.9                                  & 19.2           & 10.8                              & 24.5                        & 35.4                                        & -                                    & -                                 & -                              & 5.9  & 16.9                                                                                \\
            \midrule
            \multirow{2}{*}{\textbf{Benzene}}
             & E & \textbf{0.03}   & 0.04            & 0.04                                  & 0.06           & 0.35                              & 0.75                        & 3.3                                       & -                                    & 3.5                               & 3.4                            & -   & -                                                                                 \\
             & F & \textbf{0.3}    & \textbf{0.3}    & 0.5                                   & 0.8            & 2.6                               & 6.0                         & 10.0                                       & -                                    & 9.1                               & 8.1                            & -    & -                                                                                \\
            \midrule
            \multirow{2}{*}{\textbf{Ethanol}}
             & E & \textbf{0.4}    & \textbf{0.4}    & 1.2                                   & 2.4            & 0.9                               & 3.5                         & 2.5                                        & 2.7                                  & 4.3                               & 2.8                            & 2.6    &          3.5                                                               \\
             & F & 3.2             & \textbf{2.8}    & 7.3                                   & 16.0           & 6.2                               & 18.1                        & 13.4                                        & 10.0                                 & 14.3                              & 10.0                           & 9.1          & 16.9                                                                   \\
            \midrule
            \multirow{2}{*}{\textbf{Malonaldehyde}}
             & E & \textbf{0.8}    & \textbf{0.8}    & 1.7                                   & 3.1            & 1.5                               & 4.8                         & 4.6                                       & 3.9                                  & 5.2                               & 4.5                            & 4.1        & 5.6                                                                         \\
             & F & 5.8             & \textbf{5.1}    & 11.1                                  & 18.8           & 10.3                              & 26.4                        & 24.5                                        & 13.8                                 & 19.5                              & 16.6                           & 14.0        & 28.6                                                                        \\
            \midrule
            \multirow{2}{*}{\textbf{Naphthalene}}
             & E & \textbf{0.2}    & \textbf{0.2}    & 0.9                                   & 0.8            & 1.2                               & 3.8                         & 11.3                                       & 5.1                                  & 7.4                               & 5.3                            & 5.2         & 6.9                                                                         \\
             & F & 1.8             & \textbf{1.3}    & 5.1                                   & 5.4            & 6.5                               & 16.5                        & 29.2                                        & 3.6                                  & 15.6                              & 9.3                            & 3.6        &   25.2                                                                        \\
            \midrule
            \multirow{2}{*}{\textbf{Paracetamol}}
             & E & \textbf{1.3}    & 1.4             & 4.0                                   & 5.0            & 2.9                               & 8.5                         & 11.5                                       & -                                    & -                                 & -                              & 6.1         & -                                                                         \\
             & F & \textbf{5.8}    & 5.9             & 12.7                                  & 23.3           & 12.3                              & 28.9                        & 30.4                                        & -                                    & -                                 & -                              & 11.4       & -                                                                          \\
            \midrule
            \multirow{2}{*}{\textbf{Salicylic acid}}
             & E & 0.8             & \textbf{0.7}    & 1.8                                   & 2.1            & 1.8                               & 5.6                         & 9.2                                       & 4.9                                  & 8.2                               & 5.8                            & 4.9     & 8.7                                                                             \\
             & F & 4.3             & \textbf{4.0}    & 9.3                                   & 12.8           & 9.5                               & 24.7                        & 29.7                                        & 9.1                                  & 21.2                              & 16.2                           & 8.5                  &  36.9                                                                \\
            \midrule
            \multirow{2}{*}{\textbf{Toluene}}
             & E & \textbf{0.3}    & \textbf{0.3}    & 1.1                                   & 1.0            & 1.7                               & 4.0                         & 7.7                                        & 4.2                                  & 6.5                               & 4.4                            & 4.1    & 5.2                                                                              \\
             & F & 1.9             & \textbf{1.6}    & 6.5                                   & 6.3            & 8.8                               & 17.8                        & 24.3                                       & 4.4                                  & 14.7                              & 9.4                            & 3.8   &           24.7                                                                    \\
            \midrule
            \multirow{2}{*}{\textbf{Uracil}}
             & E & \textbf{0.4}    & \textbf{0.4}    & 1.1                                   & 1.4            & 0.6                               & 3.0                         & 5.1                                        & 4.5                                  & 5.2                               & 5.0                            & 4.6  & 4.5                                                                                \\
             & F & 3.2             & \textbf{3.1}    & 6.6                                   & 10.4           & 4.2                               & 17.6                        & 21.4                                        & 6.1                                  & 14.3                              & 13.1                           & 6.4   & 3.3                                                                               \\
            \bottomrule
        \end{tabular}
    }
\end{table*}

\subsection{3BPA: Extrapolation to higher tempature}

\begin{figure*}
    \centering
    \includegraphics[width=\linewidth]{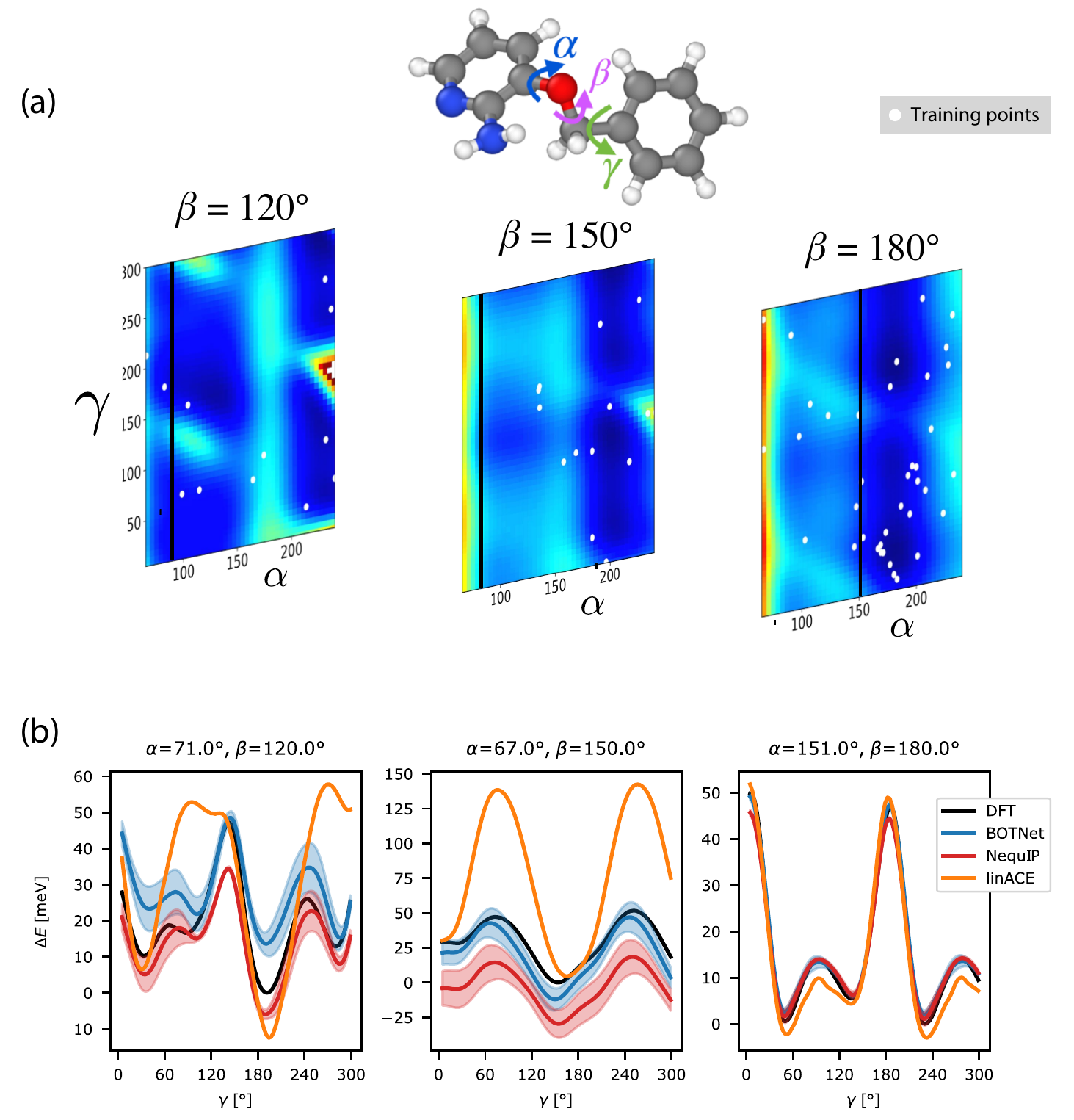}
    \caption{
        \textbf{(a)} Two dimensional slices of the potential energy surface of  3-(benzyloxy)pyridin-2-amine (3BPA). Two of the three freely rotating dihedral angles ($\gamma$ and $\alpha$) vary between 0 and 300 degrees while the third $\beta$ is kept fixed for each slice. The white dots corresponds to configuration in the training set at 300K that lie within $\pm 5^\circ$ of the fixed $\beta$. The black lines corresponds to the one dimensional cut of part (b) \protect\linebreak
        \textbf{(b)} Energy predictions on slices of the DFT potential energy landscape of 3-(benzyloxy)pyridin-2-amine (3BPA). The ground-truth (DFT) energy is shown in black. In all three panels, the energy scale is shifted so that the lowest point on the DFT curve is at zero.}
    \label{fig:3bpa}
\end{figure*}

To test the extrapolation capabilities of the different models to out-of-distribution input data we used the 3BPA dataset. This is well suited as the molecule is flexible with 3 rotating bonds as illustrated on Figure~\ref{fig:3bpa}. We use the training set collected at 300K because it samples only the pockets corresponding to the most stable dihedral angle combinations, whereas the higher temperature test sets contain geometries sampling the full dihedral profile~\cite{kovacs2021}.

\begin{table*}
    \caption{
        Root-mean-square error on 3BPA dataset.
        Energy (E, meV) and force (F, meV/\AA) errors of models trained and tested on configurations
        of the flexible drug-like molecule 3-(benzyloxy)pyridin-2-amine (3BPA) collected at 300K.
        Standard deviations are computed over three runs and shown (if available) in brackets.}
    \label{tab:3bpa}
    \centering
    \resizebox{0.8\textwidth}{!}{%
        \begin{tabular}{lllcccccccc}
            \toprule
             &                           &                      & BOTNet    & NequIP & Linear ACE & sGDML & GAP & FF & ANI & ANI-2x \\
            \midrule
            \multirow{8}{*}{}
             & \multirow{2}{*}{300 K}
             & E                         & 3.1 (0.13)           & \textbf{3.0} (0.2)   & 7.1             & 9.1                 & 22.8           & 60.8         & 23.5        & 38.6                           \\
             &                           & F                    & \textbf{11.0} (0.14) & 11.6 (0.2)      & 27.1                & 46.2           & 87.3         & 302.8       & 42.8         & 84.4            \\
            \cmidrule{2-11}
             & \multirow{2}{*}{600 K}
             & E                         & \textbf{11.5} (0.6)  & 11.9 (1.1)           & 24.0            & 484.8               & 61.4           & 136.8        & 37.8        & 54.5                           \\
             &                           & F                    & \textbf{26.7} (0.29) & 29.4 (0.8)      & 64.3                & 439.2          & 151.9        & 407.9       & 71.7         & 102.8           \\
            \cmidrule{2-11}
             & \multirow{2}{*}{1200 K}
             & E                         & \textbf{39.1} (1.1)  & 49.8 (4.0)           & 85.3            & 774.5               & 166.8          & 325.5        & 76.8        & 88.8                           \\
             &                           & F                    & \textbf{81.1} (1.5)  & 97.1 (5.6)      & 187.0               & 711.1          & 305.5        & 670.9       & 129.6        & 139.6           \\
            \cmidrule{2-11}
             & \multirow{2}{*}{Dihedral}
             & E                         & \textbf{16.3}  (1.5) & 27.0 (4.2)           & 22.2            & -                   & -              & -            & -           & -                              \\
             &                           & F                    & \textbf{20.0} (1.2)  & 23.8 (2.4)      & 39.2                & -              & -            & -           & -            & -               \\
            \bottomrule
        \end{tabular}
    }
\end{table*}

The root-mean-squared errors (RMSE) on energies and force components of several different models are shown in Table \ref{tab:3bpa}. We have included the RMSE-s on the three different temperature test sets and a 4th test set made up of the DFT dihedral scan geometries. The 300K test set measures the in-domain accuracy of the models. We observe similar accuracy between BOTNet and NequIP within the standard deviation, outperforming the closest model linear ACE by a factor of 2.
At 600K, we observe higher RMSE over all the models as the data are further from the training set and the magnitudes of the forces are also larger. Compared to the other models, BOTNet and NequIP are again more accurate by about a factor of 2.
The 1200K test set measures the most extreme extrapolation. In this case, the BOTNet model has the highest accuracy, performing around 20$\%$ better than NequIP and over two times better than all other models.
BOTNet is again the most accurate model on the dihedral scan, a different set of out-of-domain samples, proving its excellent extrapolation capabilities compared to the other models. 

\begin{table*}
    \caption{
        Root-mean-square error on the acetylacetone dataset.
        Energy (E, meV) and force (F, meV/\AA) errors of models trained on configurations
        of the acetylacetone molecule sampled at 300~K and tested on configurations sampled at 300~K and 600~K.
    }
    \label{tab:acac}
    \centering
    \begin{tabular}{lm{0.5cm}ccccc}
        \toprule
                      &   & BOTNet     & NequIP               & Linear ACE \\
        \midrule
        \multirow{2}{*}{300 K}
                      & E & 0.89 (0.0) & \textbf{0.81} (0.05) & 2.4        \\
                      & F & 6.3 (0.0)  & \textbf{5.90} (0.46) & 16.7       \\
        \multirow{2}{*}{600 K}
                      & E & 6.2 (1.1)  & \textbf{6.04} (1.54) & 8.3        \\
                      & F & 29.8 (1.0) & \textbf{27.8} (4.03) & 41.8       \\
        \midrule
        N° Parameters &   & 2,756,416  & 3,190,488            & 35,594     \\
        \bottomrule
    \end{tabular}
\end{table*}

We also inspect the shape of the potential energy surface by scanning along lines in the three-dimensional dihedral space, keeping $\alpha$ and $\beta$ fixed and varying $\gamma$. 
The energy along three such cuts is plotted in Figure~\ref{fig:3bpa} showing BOTNet, NequIP, and linear ACE predictions.
The three cuts present different degrees of prediction difficulty, primarily because they probe the PES at different energy levels above the equilibrium state.
The $\beta=120^\circ$ on the left and the $\beta=180^\circ$ on the right of Figure~\ref{fig:3bpa} (b) are easier,
because there are some training points in the dataset with similar combinations of dihedral angles,
whereas the third cut in the middle ($\beta=150^\circ$) is the most challenging of the three,
with no training data points near it. 

We can observe that all three models tested perform similarly well on the two easier cuts, with NequIP being the most accurate. On the most challenging cut in the middle of Figure~\ref{fig:3bpa} (b), we see that linear ACE smoothly reproduces the shape of the potential energy but overestimates the rotation barriers by about a factor of two. Both NequIP and BOTNet can predict the overall shape and barrier height with remarkable accuracy, with BOTNet even getting the overall energy shift right. 
Overall, all three models perform well on these tests, linear ACE is extrapolating smoothly, but very far from the training set, it can make more significant errors. In contrast, the nonlinear models are smooth and accurate even for input data far from the training distribution.

\subsection{Acetylacetone: flexibility and reactivity}

\begin{figure*}
    \centering
    \includegraphics[width=\linewidth]{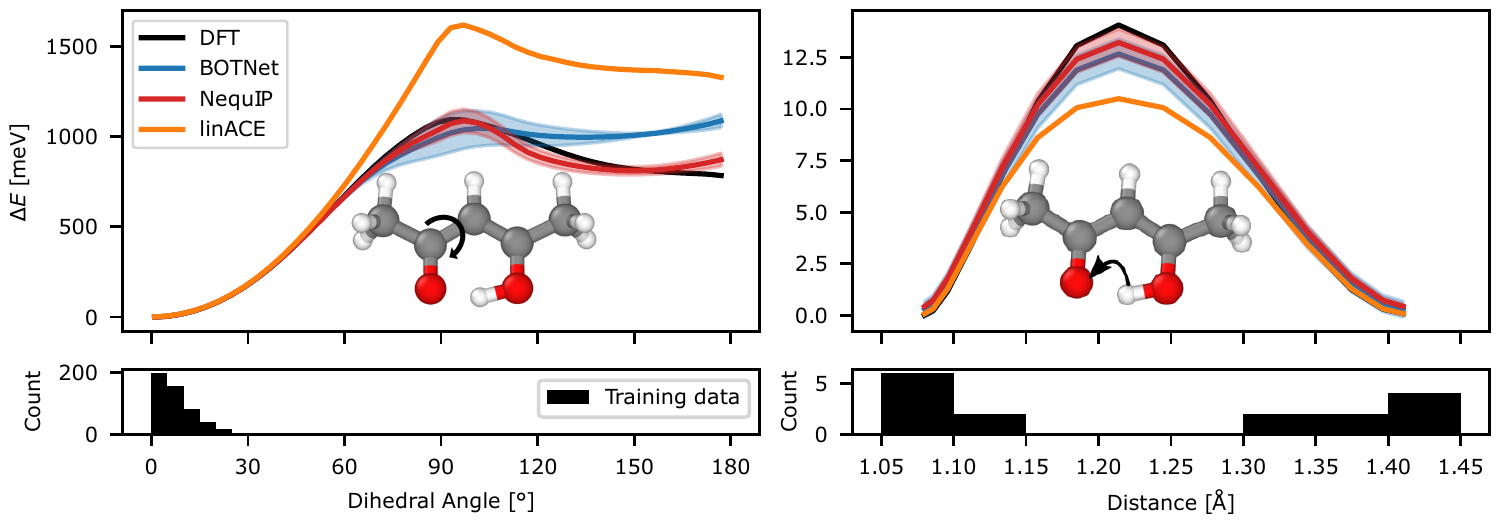}
    \caption{
        On the left, mean energy predictions of a dihedral scan of the DFT potential energy landscape of Acetyl-Acetone.
        On the right mean energy prediction of the proton transfer in Acetyl-Acetone as a function of O-H distance.
        Shaded areas indicate one standard deviations.}
    \label{fig:acac}
\end{figure*}

The potential energy surface of acetylacetone has been studied exhaustively in the past due to its many interesting properties, such as the tunneling splitting of the intramolecular hydrogen transfer~\cite{AcAcBowman}.
In this paper, we are not trying to create the most accurate PES of the molecule but deliberately use a small training set, making the task challenging for the inference methods. This helps us see the distinctions between the different models. To prepare the training set, we run a long molecular dynamics simulation at 300 K using a Langevin thermostat at the semi-empirical GFN2-xTB level of theory. We sampled several independent configurations and computed the energies and forces with density functional theory using the PBE exchange-correlation functional with D3 dispersion correction and def2-SVP basis set using the ORCA electronic structure package.

To test the models, we measure extrapolation both in temperature and along two internal coordinates of the molecule. The temperature extrapolation experiments show similar results to the 3BPA case, though NequIP performs slightly better than BOTNet. The results are shown in Table~\ref{tab:acac}.

Figure~\ref{fig:acac} shows the predictions of the three models along two different internal coordinates. The left panel shows the energy change as a function of one of the O-C-C-C dihedral angles. The training set only samples dihedral angles below $30^\circ$, and we test the models on angles up to $180^\circ$. This is a significant extrapolation in input space and energy space as the rotation barrier is about 1 eV, much more significant than the typical energy fluctuations in the training set. All models produce a smooth PES, reproducing the maxima at around 90 degrees, but remarkably NequIP and BOTNet also get the height of the barrier very accurately. We also observe that NequIP achieves better reproduction of the PES after the maxima.

On the right panel of Figure~\ref{fig:acac} we plot the energy along a reaction coordinate of the intramolecular hydrogen transfer found using the Nudged Elastic Band method~\cite{NEB1998}. This task probes how the models can cope with reactivity not too far from the training set. We can see that all models reproduce the shape of the barrier accurately, with BOTNet and NequIP getting the height of the barrier within 2 meV.

\section{Conclusion}
\label{sec:conclusion}

In this paper, we have introduced Multi-ACE, a framework in which many previously published \textit{E}(3)-equivariant (or invariant) machine-learning interatomic potentials can be understood. Using this framework, we have identified a large design space, and we have systematically studied how different choices made by the different models affect the accuracy, smoothness, and extrapolation of the fitted interatomic potentials.

Using this framework, we can identify the choices made by existing ML interatomic potentials: some use invariant 2 and 3-body features and nonlinear regression (SOAP-GAP, BPNN, etc.), and others use higher body-order features and linear regression (linear ACE, MTP) whereas most message-passing models use 2-body features locally but increase the body-order via nonlinear activations and applying multiple messages passing layers. A yet unexplored part of the design space is the use of locally many-body features in a message-passing model, and it is the subject of future investigations with a preliminary result in Ref \cite{MACE2022}.

We used NequIP as an example where we probed each of the design choices and have created a new model, BOTNet, which keeps the most crucial elements of NequIP: the equivariant tensor-product and the learnable residual architecture, but makes different choices on the radial basis, use of nonlinear activations and readouts, making it an explicitly body-ordered MPNN model. Our study also highlights the crucial importance of internal normalization and the effect of data normalization on both accuracy and extrapolation.

The design space set out in this paper and the systematic study of the different design choices provide the basis for the future development of new equivariant interatomic potentials.

\section{Author contributions}

IB, SB, GC and BK and planned the collaboration. IB and GS implemented the BOTNet software. Numerical experiments were performed by IB (BOTNet code), SB and AM (NequIP code) and  DK (Linear ACE). IB and DK produced the data sets. GC, RD and CO suggested the principles of multi-ACE and IB, DK and CO developed the equations. IB, DK, CO and GS drafted the manuscript text and figures. All authors edited the manuscript.

\section{Acknowledgments}

This work was performed using resources provided by the Cambridge Service for Data Driven Discovery (CSD3) operated by the University of Cambridge Research Computing Service (www.csd3.cam.ac.uk), provided by Dell EMC and Intel using Tier-2 funding from the Engineering and Physical Sciences Research Council (capital grant EP/T022159/1), and DiRAC funding from the Science and Technology Facilities Council (www.dirac.ac.uk). DPK acknowledges support from AstraZeneca and the Engineering and Physical Sciences Research Council. CO is supported by Leverhulme Research Project Grant RPG-2017-191 and by the Natural Sciences and Engineering Research Council of Canada (NSERC) [funding reference number IDGR019381].

Work at Harvard University was supported by Bosch Research, the US Department of Energy, Office of Basic Energy Sciences Award No. DE-SC0022199 and the Integrated Mesoscale Architectures for Sustainable Catalysis (IMASC), an Energy Frontier Research Center, Award No. DE-SC0012573 and by the NSF through the Harvard University Materials Research Science and Engineering Center Grant No. DMR-2011754. A.M is supported by U.S. Department of Energy, Office of Science, Office of Advanced Scientific Computing Research, Computational Science Graduate Fellowship under Award Number(s) DE-SC0021110. The authors acknowledge computing resources provided by the Harvard University FAS Division of Science Research Computing Group. \\

\bibliography{references}

\begin{thebibliography}{66}%
\makeatletter
\providecommand \@ifxundefined [1]{%
 \@ifx{#1\undefined}
}%
\providecommand \@ifnum [1]{%
 \ifnum #1\expandafter \@firstoftwo
 \else \expandafter \@secondoftwo
 \fi
}%
\providecommand \@ifx [1]{%
 \ifx #1\expandafter \@firstoftwo
 \else \expandafter \@secondoftwo
 \fi
}%
\providecommand \natexlab [1]{#1}%
\providecommand \enquote  [1]{``#1''}%
\providecommand \bibnamefont  [1]{#1}%
\providecommand \bibfnamefont [1]{#1}%
\providecommand \citenamefont [1]{#1}%
\providecommand \href@noop [0]{\@secondoftwo}%
\providecommand \href [0]{\begingroup \@sanitize@url \@href}%
\providecommand \@href[1]{\@@startlink{#1}\@@href}%
\providecommand \@@href[1]{\endgroup#1\@@endlink}%
\providecommand \@sanitize@url [0]{\catcode `\\12\catcode `\$12\catcode
  `\&12\catcode `\#12\catcode `\^12\catcode `\_12\catcode `\%12\relax}%
\providecommand \@@startlink[1]{}%
\providecommand \@@endlink[0]{}%
\providecommand \url  [0]{\begingroup\@sanitize@url \@url }%
\providecommand \@url [1]{\endgroup\@href {#1}{\urlprefix }}%
\providecommand \urlprefix  [0]{URL }%
\providecommand \Eprint [0]{\href }%
\providecommand \doibase [0]{http://dx.doi.org/}%
\providecommand \selectlanguage [0]{\@gobble}%
\providecommand \bibinfo  [0]{\@secondoftwo}%
\providecommand \bibfield  [0]{\@secondoftwo}%
\providecommand \translation [1]{[#1]}%
\providecommand \BibitemOpen [0]{}%
\providecommand \bibitemStop [0]{}%
\providecommand \bibitemNoStop [0]{.\EOS\space}%
\providecommand \EOS [0]{\spacefactor3000\relax}%
\providecommand \BibitemShut  [1]{\csname bibitem#1\endcsname}%
\let\auto@bib@innerbib\@empty
\bibitem [{\citenamefont {Behler}\ and\ \citenamefont
  {Parrinello}(2007)}]{Behler2007ACSF}%
  \BibitemOpen
  \bibfield  {author} {\bibinfo {author} {\bibfnamefont {J.}~\bibnamefont
  {Behler}}\ and\ \bibinfo {author} {\bibfnamefont {M.}~\bibnamefont
  {Parrinello}},\ }\href {\doibase 10.1103/PhysRevLett.98.146401} {\bibfield
  {journal} {\bibinfo  {journal} {Phys. Rev. Lett.}\ }\textbf {\bibinfo
  {volume} {98}},\ \bibinfo {pages} {146401} (\bibinfo {year}
  {2007})}\BibitemShut {NoStop}%
\bibitem [{\citenamefont {Bart\'ok}\ \emph {et~al.}(2013)\citenamefont
  {Bart\'ok}, \citenamefont {Kondor},\ and\ \citenamefont
  {Cs\'anyi}}]{Csanyi2013SOAP}%
  \BibitemOpen
  \bibfield  {author} {\bibinfo {author} {\bibfnamefont {A.~P.}\ \bibnamefont
  {Bart\'ok}}, \bibinfo {author} {\bibfnamefont {R.}~\bibnamefont {Kondor}}, \
  and\ \bibinfo {author} {\bibfnamefont {G.}~\bibnamefont {Cs\'anyi}},\ }\href
  {\doibase 10.1103/PhysRevB.87.184115} {\bibfield  {journal} {\bibinfo
  {journal} {Phys. Rev. B}\ }\textbf {\bibinfo {volume} {87}},\ \bibinfo
  {pages} {184115} (\bibinfo {year} {2013})}\BibitemShut {NoStop}%
\bibitem [{\citenamefont {Behler}(2021)}]{Behler2021ChemRev}%
  \BibitemOpen
  \bibfield  {author} {\bibinfo {author} {\bibfnamefont {J.}~\bibnamefont
  {Behler}},\ }\href {\doibase 10.1021/acs.chemrev.0c00868} {\bibfield
  {journal} {\bibinfo  {journal} {Chemical Reviews}\ }\textbf {\bibinfo
  {volume} {121}},\ \bibinfo {pages} {10037} (\bibinfo {year} {2021})},\
  \bibinfo {note} {pMID: 33779150},\ \Eprint
  {http://arxiv.org/abs/https://doi.org/10.1021/acs.chemrev.0c00868}
  {https://doi.org/10.1021/acs.chemrev.0c00868} \BibitemShut {NoStop}%
\bibitem [{\citenamefont {Deringer}\ \emph {et~al.}(2021)\citenamefont
  {Deringer}, \citenamefont {Bartók}, \citenamefont {Bernstein}, \citenamefont
  {Wilkins}, \citenamefont {Ceriotti},\ and\ \citenamefont
  {Csányi}}]{DeringerCsanyi2021ChemRev}%
  \BibitemOpen
  \bibfield  {author} {\bibinfo {author} {\bibfnamefont {V.~L.}\ \bibnamefont
  {Deringer}}, \bibinfo {author} {\bibfnamefont {A.~P.}\ \bibnamefont
  {Bartók}}, \bibinfo {author} {\bibfnamefont {N.}~\bibnamefont {Bernstein}},
  \bibinfo {author} {\bibfnamefont {D.~M.}\ \bibnamefont {Wilkins}}, \bibinfo
  {author} {\bibfnamefont {M.}~\bibnamefont {Ceriotti}}, \ and\ \bibinfo
  {author} {\bibfnamefont {G.}~\bibnamefont {Csányi}},\ }\href {\doibase
  10.1021/acs.chemrev.1c00022} {\bibfield  {journal} {\bibinfo  {journal}
  {Chemical Reviews}\ }\textbf {\bibinfo {volume} {121}},\ \bibinfo {pages}
  {10073} (\bibinfo {year} {2021})},\ \bibinfo {note} {pMID: 34398616},\
  \Eprint {http://arxiv.org/abs/https://doi.org/10.1021/acs.chemrev.1c00022}
  {https://doi.org/10.1021/acs.chemrev.1c00022} \BibitemShut {NoStop}%
\bibitem [{\citenamefont {Drautz}(2019)}]{ACE_ralf}%
  \BibitemOpen
  \bibfield  {author} {\bibinfo {author} {\bibfnamefont {R.}~\bibnamefont
  {Drautz}},\ }\href {\doibase 10.1103/PhysRevB.99.014104} {\bibfield
  {journal} {\bibinfo  {journal} {Phys. Rev. B}\ }\textbf {\bibinfo {volume}
  {99}},\ \bibinfo {pages} {014104} (\bibinfo {year} {2019})}\BibitemShut
  {NoStop}%
\bibitem [{\citenamefont {Dusson}\ \emph {et~al.}(2022)\citenamefont {Dusson},
  \citenamefont {Bachmayr}, \citenamefont {Csányi}, \citenamefont {Drautz},
  \citenamefont {Etter}, \citenamefont {{van der Oord}},\ and\ \citenamefont
  {Ortner}}]{DUSSON2022}%
  \BibitemOpen
  \bibfield  {author} {\bibinfo {author} {\bibfnamefont {G.}~\bibnamefont
  {Dusson}}, \bibinfo {author} {\bibfnamefont {M.}~\bibnamefont {Bachmayr}},
  \bibinfo {author} {\bibfnamefont {G.}~\bibnamefont {Csányi}}, \bibinfo
  {author} {\bibfnamefont {R.}~\bibnamefont {Drautz}}, \bibinfo {author}
  {\bibfnamefont {S.}~\bibnamefont {Etter}}, \bibinfo {author} {\bibfnamefont
  {C.}~\bibnamefont {{van der Oord}}}, \ and\ \bibinfo {author} {\bibfnamefont
  {C.}~\bibnamefont {Ortner}},\ }\href {\doibase
  https://doi.org/10.1016/j.jcp.2022.110946} {\bibfield  {journal} {\bibinfo
  {journal} {Journal of Computational Physics}\ }\textbf {\bibinfo {volume}
  {454}},\ \bibinfo {pages} {110946} (\bibinfo {year} {2022})}\BibitemShut
  {NoStop}%
\bibitem [{\citenamefont {Bartók}\ \emph {et~al.}(2010)\citenamefont
  {Bartók}, \citenamefont {Payne}, \citenamefont {Kondor},\ and\ \citenamefont
  {Csányi}}]{GAP2010}%
  \BibitemOpen
  \bibfield  {author} {\bibinfo {author} {\bibfnamefont {A.~P.}\ \bibnamefont
  {Bartók}}, \bibinfo {author} {\bibfnamefont {M.~C.}\ \bibnamefont {Payne}},
  \bibinfo {author} {\bibfnamefont {R.}~\bibnamefont {Kondor}}, \ and\ \bibinfo
  {author} {\bibfnamefont {G.}~\bibnamefont {Csányi}},\ }\href {\doibase
  10.1103/physrevlett.104.136403} {\bibfield  {journal} {\bibinfo  {journal}
  {Physical Review Letters}\ }\textbf {\bibinfo {volume} {104}} (\bibinfo
  {year} {2010}),\ 10.1103/physrevlett.104.136403}\BibitemShut {NoStop}%
\bibitem [{\citenamefont {Musil}\ \emph
  {et~al.}(2021{\natexlab{a}})\citenamefont {Musil}, \citenamefont {Grisafi},
  \citenamefont {Bartók}, \citenamefont {Ortner}, \citenamefont {Csányi},\
  and\ \citenamefont {Ceriotti}}]{Ceriotti2021Representation}%
  \BibitemOpen
  \bibfield  {author} {\bibinfo {author} {\bibfnamefont {F.}~\bibnamefont
  {Musil}}, \bibinfo {author} {\bibfnamefont {A.}~\bibnamefont {Grisafi}},
  \bibinfo {author} {\bibfnamefont {A.~P.}\ \bibnamefont {Bartók}}, \bibinfo
  {author} {\bibfnamefont {C.}~\bibnamefont {Ortner}}, \bibinfo {author}
  {\bibfnamefont {G.}~\bibnamefont {Csányi}}, \ and\ \bibinfo {author}
  {\bibfnamefont {M.}~\bibnamefont {Ceriotti}},\ }\href {\doibase
  10.1021/acs.chemrev.1c00021} {\bibfield  {journal} {\bibinfo  {journal}
  {Chemical Reviews}\ }\textbf {\bibinfo {volume} {121}},\ \bibinfo {pages}
  {9759} (\bibinfo {year} {2021}{\natexlab{a}})},\ \bibinfo {note} {pMID:
  34310133},\ \Eprint
  {http://arxiv.org/abs/https://doi.org/10.1021/acs.chemrev.1c00021}
  {https://doi.org/10.1021/acs.chemrev.1c00021} \BibitemShut {NoStop}%
\bibitem [{\citenamefont {Drautz}(2020)}]{ACE_equivariant_ralf}%
  \BibitemOpen
  \bibfield  {author} {\bibinfo {author} {\bibfnamefont {R.}~\bibnamefont
  {Drautz}},\ }\href {\doibase 10.1103/PhysRevB.102.024104} {\bibfield
  {journal} {\bibinfo  {journal} {Phys. Rev. B}\ }\textbf {\bibinfo {volume}
  {102}},\ \bibinfo {pages} {024104} (\bibinfo {year} {2020})}\BibitemShut
  {NoStop}%
\bibitem [{\citenamefont {Shapeev}(2016)}]{MTP}%
  \BibitemOpen
  \bibfield  {author} {\bibinfo {author} {\bibfnamefont {A.~V.}\ \bibnamefont
  {Shapeev}},\ }\href {\doibase 10.1137/15M1054183} {\bibfield  {journal}
  {\bibinfo  {journal} {Multiscale Modeling Simulation}\ }\textbf {\bibinfo
  {volume} {14}},\ \bibinfo {pages} {1153} (\bibinfo {year} {2016})},\ \Eprint
  {http://arxiv.org/abs/https://doi.org/10.1137/15M1054183}
  {https://doi.org/10.1137/15M1054183} \BibitemShut {NoStop}%
\bibitem [{\citenamefont {Kovács}\ \emph {et~al.}(2021)\citenamefont
  {Kovács}, \citenamefont {Oord}, \citenamefont {Kucera}, \citenamefont
  {Allen}, \citenamefont {Cole}, \citenamefont {Ortner},\ and\ \citenamefont
  {Csányi}}]{kovacs2021}%
  \BibitemOpen
  \bibfield  {author} {\bibinfo {author} {\bibfnamefont {D.~P.}\ \bibnamefont
  {Kovács}}, \bibinfo {author} {\bibfnamefont {C.~v.~d.}\ \bibnamefont
  {Oord}}, \bibinfo {author} {\bibfnamefont {J.}~\bibnamefont {Kucera}},
  \bibinfo {author} {\bibfnamefont {A.~E.~A.}\ \bibnamefont {Allen}}, \bibinfo
  {author} {\bibfnamefont {D.~J.}\ \bibnamefont {Cole}}, \bibinfo {author}
  {\bibfnamefont {C.}~\bibnamefont {Ortner}}, \ and\ \bibinfo {author}
  {\bibfnamefont {G.}~\bibnamefont {Csányi}},\ }\href {\doibase
  10.1021/acs.jctc.1c00647} {\bibfield  {journal} {\bibinfo  {journal} {Journal
  of Chemical Theory and Computation}\ }\textbf {\bibinfo {volume} {17}},\
  \bibinfo {pages} {7696} (\bibinfo {year} {2021})},\ \bibinfo {note} {pMID:
  34735161},\ \Eprint
  {http://arxiv.org/abs/https://doi.org/10.1021/acs.jctc.1c00647}
  {https://doi.org/10.1021/acs.jctc.1c00647} \BibitemShut {NoStop}%
\bibitem [{\citenamefont {Keith}\ \emph {et~al.}(2021)\citenamefont {Keith},
  \citenamefont {Vassilev-Galindo}, \citenamefont {Cheng}, \citenamefont
  {Chmiela}, \citenamefont {Gastegger}, \citenamefont {Müller},\ and\
  \citenamefont {Tkatchenko}}]{Tkatchenko2021ChemRev}%
  \BibitemOpen
  \bibfield  {author} {\bibinfo {author} {\bibfnamefont {J.~A.}\ \bibnamefont
  {Keith}}, \bibinfo {author} {\bibfnamefont {V.}~\bibnamefont
  {Vassilev-Galindo}}, \bibinfo {author} {\bibfnamefont {B.}~\bibnamefont
  {Cheng}}, \bibinfo {author} {\bibfnamefont {S.}~\bibnamefont {Chmiela}},
  \bibinfo {author} {\bibfnamefont {M.}~\bibnamefont {Gastegger}}, \bibinfo
  {author} {\bibfnamefont {K.-R.}\ \bibnamefont {Müller}}, \ and\ \bibinfo
  {author} {\bibfnamefont {A.}~\bibnamefont {Tkatchenko}},\ }\href {\doibase
  10.1021/acs.chemrev.1c00107} {\bibfield  {journal} {\bibinfo  {journal}
  {Chemical Reviews}\ }\textbf {\bibinfo {volume} {121}},\ \bibinfo {pages}
  {9816} (\bibinfo {year} {2021})},\ \bibinfo {note} {pMID: 34232033},\ \Eprint
  {http://arxiv.org/abs/https://doi.org/10.1021/acs.chemrev.1c00107}
  {https://doi.org/10.1021/acs.chemrev.1c00107} \BibitemShut {NoStop}%
\bibitem [{\citenamefont {Faber}\ \emph {et~al.}(2018)\citenamefont {Faber},
  \citenamefont {Christensen}, \citenamefont {Huang},\ and\ \citenamefont {von
  Lilienfeld}}]{Faber2018FCHL}%
  \BibitemOpen
  \bibfield  {author} {\bibinfo {author} {\bibfnamefont {F.~A.}\ \bibnamefont
  {Faber}}, \bibinfo {author} {\bibfnamefont {A.~S.}\ \bibnamefont
  {Christensen}}, \bibinfo {author} {\bibfnamefont {B.}~\bibnamefont {Huang}},
  \ and\ \bibinfo {author} {\bibfnamefont {O.~A.}\ \bibnamefont {von
  Lilienfeld}},\ }\href {\doibase 10.1063/1.5020710} {\bibfield  {journal}
  {\bibinfo  {journal} {The Journal of Chemical Physics}\ }\textbf {\bibinfo
  {volume} {148}},\ \bibinfo {pages} {241717} (\bibinfo {year} {2018})},\
  \Eprint {http://arxiv.org/abs/https://doi.org/10.1063/1.5020710}
  {https://doi.org/10.1063/1.5020710} \BibitemShut {NoStop}%
\bibitem [{\citenamefont {Zhu}\ \emph {et~al.}(2016)\citenamefont {Zhu},
  \citenamefont {Amsler}, \citenamefont {Fuhrer}, \citenamefont {Schaefer},
  \citenamefont {Faraji}, \citenamefont {Rostami}, \citenamefont {Ghasemi},
  \citenamefont {Sadeghi}, \citenamefont {Grauzinyte}, \citenamefont
  {Wolverton},\ and\ \citenamefont {Goedecker}}]{Goedecker_OM_fingerprints}%
  \BibitemOpen
  \bibfield  {author} {\bibinfo {author} {\bibfnamefont {L.}~\bibnamefont
  {Zhu}}, \bibinfo {author} {\bibfnamefont {M.}~\bibnamefont {Amsler}},
  \bibinfo {author} {\bibfnamefont {T.}~\bibnamefont {Fuhrer}}, \bibinfo
  {author} {\bibfnamefont {B.}~\bibnamefont {Schaefer}}, \bibinfo {author}
  {\bibfnamefont {S.}~\bibnamefont {Faraji}}, \bibinfo {author} {\bibfnamefont
  {S.}~\bibnamefont {Rostami}}, \bibinfo {author} {\bibfnamefont {S.~A.}\
  \bibnamefont {Ghasemi}}, \bibinfo {author} {\bibfnamefont {A.}~\bibnamefont
  {Sadeghi}}, \bibinfo {author} {\bibfnamefont {M.}~\bibnamefont {Grauzinyte}},
  \bibinfo {author} {\bibfnamefont {C.}~\bibnamefont {Wolverton}}, \ and\
  \bibinfo {author} {\bibfnamefont {S.}~\bibnamefont {Goedecker}},\ }\href
  {\doibase 10.1063/1.4940026} {\bibfield  {journal} {\bibinfo  {journal} {The
  Journal of Chemical Physics}\ }\textbf {\bibinfo {volume} {144}},\ \bibinfo
  {pages} {034203} (\bibinfo {year} {2016})},\ \Eprint
  {http://arxiv.org/abs/https://doi.org/10.1063/1.4940026}
  {https://doi.org/10.1063/1.4940026} \BibitemShut {NoStop}%
\bibitem [{\citenamefont {Sch\"{u}tt}\ \emph {et~al.}(2017)\citenamefont
  {Sch\"{u}tt}, \citenamefont {Kindermans}, \citenamefont {Sauceda~Felix},
  \citenamefont {Chmiela}, \citenamefont {Tkatchenko},\ and\ \citenamefont
  {M\"{u}ller}}]{schnet}%
  \BibitemOpen
  \bibfield  {author} {\bibinfo {author} {\bibfnamefont {K.}~\bibnamefont
  {Sch\"{u}tt}}, \bibinfo {author} {\bibfnamefont {P.-J.}\ \bibnamefont
  {Kindermans}}, \bibinfo {author} {\bibfnamefont {H.~E.}\ \bibnamefont
  {Sauceda~Felix}}, \bibinfo {author} {\bibfnamefont {S.}~\bibnamefont
  {Chmiela}}, \bibinfo {author} {\bibfnamefont {A.}~\bibnamefont {Tkatchenko}},
  \ and\ \bibinfo {author} {\bibfnamefont {K.-R.}\ \bibnamefont {M\"{u}ller}},\
  }in\ \href
  {https://proceedings.neurips.cc/paper/2017/file/303ed4c69846ab36c2904d3ba8573050-Paper.pdf}
  {\emph {\bibinfo {booktitle} {Advances in Neural Information Processing
  Systems}}},\ Vol.~\bibinfo {volume} {30},\ \bibinfo {editor} {edited by\
  \bibinfo {editor} {\bibfnamefont {I.}~\bibnamefont {Guyon}}, \bibinfo
  {editor} {\bibfnamefont {U.~V.}\ \bibnamefont {Luxburg}}, \bibinfo {editor}
  {\bibfnamefont {S.}~\bibnamefont {Bengio}}, \bibinfo {editor} {\bibfnamefont
  {H.}~\bibnamefont {Wallach}}, \bibinfo {editor} {\bibfnamefont
  {R.}~\bibnamefont {Fergus}}, \bibinfo {editor} {\bibfnamefont
  {S.}~\bibnamefont {Vishwanathan}}, \ and\ \bibinfo {editor} {\bibfnamefont
  {R.}~\bibnamefont {Garnett}}}\ (\bibinfo  {publisher} {Curran Associates,
  Inc.},\ \bibinfo {year} {2017})\BibitemShut {NoStop}%
\bibitem [{\citenamefont {Unke}\ and\ \citenamefont {Meuwly}(2019)}]{Physnet}%
  \BibitemOpen
  \bibfield  {author} {\bibinfo {author} {\bibfnamefont {O.~T.}\ \bibnamefont
  {Unke}}\ and\ \bibinfo {author} {\bibfnamefont {M.}~\bibnamefont {Meuwly}},\
  }\href {\doibase 10.1021/acs.jctc.9b00181} {\bibfield  {journal} {\bibinfo
  {journal} {Journal of Chemical Theory and Computation}\ }\textbf {\bibinfo
  {volume} {15}},\ \bibinfo {pages} {3678–3693} (\bibinfo {year}
  {2019})}\BibitemShut {NoStop}%
\bibitem [{\citenamefont {Klicpera}\ \emph {et~al.}(2020)\citenamefont
  {Klicpera}, \citenamefont {Groß},\ and\ \citenamefont
  {Günnemann}}]{DimeNet}%
  \BibitemOpen
  \bibfield  {author} {\bibinfo {author} {\bibfnamefont {J.}~\bibnamefont
  {Klicpera}}, \bibinfo {author} {\bibfnamefont {J.}~\bibnamefont {Groß}}, \
  and\ \bibinfo {author} {\bibfnamefont {S.}~\bibnamefont {Günnemann}},\
  }\href@noop {} {\enquote {\bibinfo {title} {Directional message passing for
  molecular graphs},}\ } (\bibinfo {year} {2020}),\ \Eprint
  {http://arxiv.org/abs/2003.03123} {arXiv:2003.03123 [cs.LG]} \BibitemShut
  {NoStop}%
\bibitem [{\citenamefont {Anderson}\ \emph
  {et~al.}(2019{\natexlab{a}})\citenamefont {Anderson}, \citenamefont {Hy},\
  and\ \citenamefont {Kondor}}]{Cormorant}%
  \BibitemOpen
  \bibfield  {author} {\bibinfo {author} {\bibfnamefont {B.}~\bibnamefont
  {Anderson}}, \bibinfo {author} {\bibfnamefont {T.~S.}\ \bibnamefont {Hy}}, \
  and\ \bibinfo {author} {\bibfnamefont {R.}~\bibnamefont {Kondor}},\ }in\
  \href
  {https://proceedings.neurips.cc/paper/2019/file/03573b32b2746e6e8ca98b9123f2249b-Paper.pdf}
  {\emph {\bibinfo {booktitle} {Advances in Neural Information Processing
  Systems}}},\ Vol.~\bibinfo {volume} {32},\ \bibinfo {editor} {edited by\
  \bibinfo {editor} {\bibfnamefont {H.}~\bibnamefont {Wallach}}, \bibinfo
  {editor} {\bibfnamefont {H.}~\bibnamefont {Larochelle}}, \bibinfo {editor}
  {\bibfnamefont {A.}~\bibnamefont {Beygelzimer}}, \bibinfo {editor}
  {\bibfnamefont {F.}~\bibnamefont {d\textquotesingle Alch\'{e}-Buc}}, \bibinfo
  {editor} {\bibfnamefont {E.}~\bibnamefont {Fox}}, \ and\ \bibinfo {editor}
  {\bibfnamefont {R.}~\bibnamefont {Garnett}}}\ (\bibinfo  {publisher} {Curran
  Associates, Inc.},\ \bibinfo {year} {2019})\BibitemShut {NoStop}%
\bibitem [{\citenamefont {Thomas}\ \emph {et~al.}(2018)\citenamefont {Thomas},
  \citenamefont {Smidt}, \citenamefont {Kearnes}, \citenamefont {Yang},
  \citenamefont {Li}, \citenamefont {Kohlhoff},\ and\ \citenamefont
  {Riley}}]{TensorField}%
  \BibitemOpen
  \bibfield  {author} {\bibinfo {author} {\bibfnamefont {N.}~\bibnamefont
  {Thomas}}, \bibinfo {author} {\bibfnamefont {T.}~\bibnamefont {Smidt}},
  \bibinfo {author} {\bibfnamefont {S.~M.}\ \bibnamefont {Kearnes}}, \bibinfo
  {author} {\bibfnamefont {L.}~\bibnamefont {Yang}}, \bibinfo {author}
  {\bibfnamefont {L.}~\bibnamefont {Li}}, \bibinfo {author} {\bibfnamefont
  {K.}~\bibnamefont {Kohlhoff}}, \ and\ \bibinfo {author} {\bibfnamefont
  {P.}~\bibnamefont {Riley}},\ }\href {http://arxiv.org/abs/1802.08219}
  {\bibfield  {journal} {\bibinfo  {journal} {CoRR}\ }\textbf {\bibinfo
  {volume} {abs/1802.08219}} (\bibinfo {year} {2018})},\ \Eprint
  {http://arxiv.org/abs/1802.08219} {arXiv:1802.08219} \BibitemShut {NoStop}%
\bibitem [{\citenamefont {Weiler}\ \emph {et~al.}(2018)\citenamefont {Weiler},
  \citenamefont {Geiger}, \citenamefont {Welling}, \citenamefont {Boomsma},\
  and\ \citenamefont {Cohen}}]{WeilerGatedNonLinearities2018}%
  \BibitemOpen
  \bibfield  {author} {\bibinfo {author} {\bibfnamefont {M.}~\bibnamefont
  {Weiler}}, \bibinfo {author} {\bibfnamefont {M.}~\bibnamefont {Geiger}},
  \bibinfo {author} {\bibfnamefont {M.}~\bibnamefont {Welling}}, \bibinfo
  {author} {\bibfnamefont {W.}~\bibnamefont {Boomsma}}, \ and\ \bibinfo
  {author} {\bibfnamefont {T.}~\bibnamefont {Cohen}},\ }\href
  {http://arxiv.org/abs/1807.02547} {\bibfield  {journal} {\bibinfo  {journal}
  {CoRR}\ }\textbf {\bibinfo {volume} {abs/1807.02547}} (\bibinfo {year}
  {2018})},\ \Eprint {http://arxiv.org/abs/1807.02547} {1807.02547}
  \BibitemShut {NoStop}%
\bibitem [{\citenamefont {Batzner}\ \emph {et~al.}(2022)\citenamefont
  {Batzner}, \citenamefont {Musaelian}, \citenamefont {Sun}, \citenamefont
  {Geiger}, \citenamefont {Mailoa}, \citenamefont {Kornbluth}, \citenamefont
  {Molinari}, \citenamefont {Smidt},\ and\ \citenamefont {Kozinsky}}]{nequip}%
  \BibitemOpen
  \bibfield  {author} {\bibinfo {author} {\bibfnamefont {S.}~\bibnamefont
  {Batzner}}, \bibinfo {author} {\bibfnamefont {A.}~\bibnamefont {Musaelian}},
  \bibinfo {author} {\bibfnamefont {L.}~\bibnamefont {Sun}}, \bibinfo {author}
  {\bibfnamefont {M.}~\bibnamefont {Geiger}}, \bibinfo {author} {\bibfnamefont
  {J.~P.}\ \bibnamefont {Mailoa}}, \bibinfo {author} {\bibfnamefont
  {M.}~\bibnamefont {Kornbluth}}, \bibinfo {author} {\bibfnamefont
  {N.}~\bibnamefont {Molinari}}, \bibinfo {author} {\bibfnamefont {T.~E.}\
  \bibnamefont {Smidt}}, \ and\ \bibinfo {author} {\bibfnamefont
  {B.}~\bibnamefont {Kozinsky}},\ }\href@noop {} {\bibfield  {journal}
  {\bibinfo  {journal} {Nature Communications}\ }\textbf {\bibinfo {volume}
  {13}},\ \bibinfo {pages} {2453} (\bibinfo {year} {2022})}\BibitemShut
  {NoStop}%
\bibitem [{\citenamefont {Satorras}\ \emph {et~al.}(2021)\citenamefont
  {Satorras}, \citenamefont {Hoogeboom},\ and\ \citenamefont {Welling}}]{EGNN}%
  \BibitemOpen
  \bibfield  {author} {\bibinfo {author} {\bibfnamefont {V.~G.}\ \bibnamefont
  {Satorras}}, \bibinfo {author} {\bibfnamefont {E.}~\bibnamefont {Hoogeboom}},
  \ and\ \bibinfo {author} {\bibfnamefont {M.}~\bibnamefont {Welling}},\ }\href
  {https://arxiv.org/abs/2102.09844} {\bibfield  {journal} {\bibinfo  {journal}
  {CoRR}\ }\textbf {\bibinfo {volume} {abs/2102.09844}} (\bibinfo {year}
  {2021})},\ \Eprint {http://arxiv.org/abs/2102.09844} {arXiv:2102.09844}
  \BibitemShut {NoStop}%
\bibitem [{\citenamefont {Sch{\"{u}}tt}\ \emph {et~al.}(2021)\citenamefont
  {Sch{\"{u}}tt}, \citenamefont {Unke},\ and\ \citenamefont
  {Gastegger}}]{Schutt2021Painn}%
  \BibitemOpen
  \bibfield  {author} {\bibinfo {author} {\bibfnamefont {K.~T.}\ \bibnamefont
  {Sch{\"{u}}tt}}, \bibinfo {author} {\bibfnamefont {O.~T.}\ \bibnamefont
  {Unke}}, \ and\ \bibinfo {author} {\bibfnamefont {M.}~\bibnamefont
  {Gastegger}},\ }\href {https://arxiv.org/abs/2102.03150} {\bibfield
  {journal} {\bibinfo  {journal} {CoRR}\ }\textbf {\bibinfo {volume}
  {abs/2102.03150}} (\bibinfo {year} {2021})},\ \Eprint
  {http://arxiv.org/abs/2102.03150} {2102.03150} \BibitemShut {NoStop}%
\bibitem [{\citenamefont {Haghighatlari}\ \emph {et~al.}(2021)\citenamefont
  {Haghighatlari}, \citenamefont {Li}, \citenamefont {Guan}, \citenamefont
  {Zhang}, \citenamefont {Das}, \citenamefont {Stein}, \citenamefont
  {Heidar-Zadeh}, \citenamefont {Liu}, \citenamefont {Head-Gordon},
  \citenamefont {Bertels}, \citenamefont {Hao}, \citenamefont {Leven},\ and\
  \citenamefont {Head-Gordon}}]{haghighatlari2021newtonnet}%
  \BibitemOpen
  \bibfield  {author} {\bibinfo {author} {\bibfnamefont {M.}~\bibnamefont
  {Haghighatlari}}, \bibinfo {author} {\bibfnamefont {J.}~\bibnamefont {Li}},
  \bibinfo {author} {\bibfnamefont {X.}~\bibnamefont {Guan}}, \bibinfo {author}
  {\bibfnamefont {O.}~\bibnamefont {Zhang}}, \bibinfo {author} {\bibfnamefont
  {A.}~\bibnamefont {Das}}, \bibinfo {author} {\bibfnamefont {C.~J.}\
  \bibnamefont {Stein}}, \bibinfo {author} {\bibfnamefont {F.}~\bibnamefont
  {Heidar-Zadeh}}, \bibinfo {author} {\bibfnamefont {M.}~\bibnamefont {Liu}},
  \bibinfo {author} {\bibfnamefont {M.}~\bibnamefont {Head-Gordon}}, \bibinfo
  {author} {\bibfnamefont {L.}~\bibnamefont {Bertels}}, \bibinfo {author}
  {\bibfnamefont {H.}~\bibnamefont {Hao}}, \bibinfo {author} {\bibfnamefont
  {I.}~\bibnamefont {Leven}}, \ and\ \bibinfo {author} {\bibfnamefont
  {T.}~\bibnamefont {Head-Gordon}},\ }\href@noop {} {\enquote {\bibinfo {title}
  {Newtonnet: A newtonian message passing network for deep learning of
  interatomic potentials and forces},}\ } (\bibinfo {year} {2021}),\ \Eprint
  {http://arxiv.org/abs/2108.02913} {arXiv:2108.02913 [physics.chem-ph]}
  \BibitemShut {NoStop}%
\bibitem [{\citenamefont {Klicpera}\ \emph {et~al.}(2022)\citenamefont
  {Klicpera}, \citenamefont {Becker},\ and\ \citenamefont
  {Günnemann}}]{klicpera2022gemnet}%
  \BibitemOpen
  \bibfield  {author} {\bibinfo {author} {\bibfnamefont {J.}~\bibnamefont
  {Klicpera}}, \bibinfo {author} {\bibfnamefont {F.}~\bibnamefont {Becker}}, \
  and\ \bibinfo {author} {\bibfnamefont {S.}~\bibnamefont {Günnemann}},\
  }\href@noop {} {\enquote {\bibinfo {title} {Gemnet: Universal directional
  graph neural networks for molecules},}\ } (\bibinfo {year} {2022}),\ \Eprint
  {http://arxiv.org/abs/2106.08903} {arXiv:2106.08903 [physics.comp-ph]}
  \BibitemShut {NoStop}%
\bibitem [{\citenamefont {Th{\"o}lke}\ and\ \citenamefont
  {Fabritiis}(2022)}]{torchmd-net2022}%
  \BibitemOpen
  \bibfield  {author} {\bibinfo {author} {\bibfnamefont {P.}~\bibnamefont
  {Th{\"o}lke}}\ and\ \bibinfo {author} {\bibfnamefont {G.~D.}\ \bibnamefont
  {Fabritiis}},\ }in\ \href {https://openreview.net/forum?id=zNHzqZ9wrRB}
  {\emph {\bibinfo {booktitle} {International Conference on Learning
  Representations}}}\ (\bibinfo {year} {2022})\BibitemShut {NoStop}%
\bibitem [{\citenamefont {Brandstetter}\ \emph {et~al.}(2021)\citenamefont
  {Brandstetter}, \citenamefont {Hesselink}, \citenamefont {van~der Pol},
  \citenamefont {Bekkers},\ and\ \citenamefont
  {Welling}}]{Brandstetter2021Geometric}%
  \BibitemOpen
  \bibfield  {author} {\bibinfo {author} {\bibfnamefont {J.}~\bibnamefont
  {Brandstetter}}, \bibinfo {author} {\bibfnamefont {R.}~\bibnamefont
  {Hesselink}}, \bibinfo {author} {\bibfnamefont {E.}~\bibnamefont {van~der
  Pol}}, \bibinfo {author} {\bibfnamefont {E.~J.}\ \bibnamefont {Bekkers}}, \
  and\ \bibinfo {author} {\bibfnamefont {M.}~\bibnamefont {Welling}},\ }in\
  \href@noop {} {\emph {\bibinfo {booktitle} {International {{Conference}} on
  {{Learning Representations}}}}}\ (\bibinfo {year} {2021})\BibitemShut
  {NoStop}%
\bibitem [{\citenamefont {Musaelian}\ \emph {et~al.}(2022)\citenamefont
  {Musaelian}, \citenamefont {Batzner}, \citenamefont {Johansson},
  \citenamefont {Sun}, \citenamefont {Owen}, \citenamefont {Kornbluth},\ and\
  \citenamefont {Kozinsky}}]{Allegro2022}%
  \BibitemOpen
  \bibfield  {author} {\bibinfo {author} {\bibfnamefont {A.}~\bibnamefont
  {Musaelian}}, \bibinfo {author} {\bibfnamefont {S.}~\bibnamefont {Batzner}},
  \bibinfo {author} {\bibfnamefont {A.}~\bibnamefont {Johansson}}, \bibinfo
  {author} {\bibfnamefont {L.}~\bibnamefont {Sun}}, \bibinfo {author}
  {\bibfnamefont {C.~J.}\ \bibnamefont {Owen}}, \bibinfo {author}
  {\bibfnamefont {M.}~\bibnamefont {Kornbluth}}, \ and\ \bibinfo {author}
  {\bibfnamefont {B.}~\bibnamefont {Kozinsky}},\ }\href {\doibase
  10.48550/ARXIV.2204.05249} {\enquote {\bibinfo {title} {Learning local
  equivariant representations for large-scale atomistic dynamics},}\ }
  (\bibinfo {year} {2022})\BibitemShut {NoStop}%
\bibitem [{\citenamefont {Nigam}\ \emph {et~al.}(2022)\citenamefont {Nigam},
  \citenamefont {Fraux},\ and\ \citenamefont {Ceriotti}}]{nigam2022unified}%
  \BibitemOpen
  \bibfield  {author} {\bibinfo {author} {\bibfnamefont {J.}~\bibnamefont
  {Nigam}}, \bibinfo {author} {\bibfnamefont {G.}~\bibnamefont {Fraux}}, \ and\
  \bibinfo {author} {\bibfnamefont {M.}~\bibnamefont {Ceriotti}},\ }\href@noop
  {} {\enquote {\bibinfo {title} {Unified theory of atom-centered
  representations and graph convolutional machine-learning schemes},}\ }
  (\bibinfo {year} {2022}),\ \Eprint {http://arxiv.org/abs/2202.01566}
  {arXiv:2202.01566 [stat.ML]} \BibitemShut {NoStop}%
\bibitem [{\citenamefont {Bochkarev}\ \emph
  {et~al.}(2022{\natexlab{a}})\citenamefont {Bochkarev}, \citenamefont
  {Lysogorskiy}, \citenamefont {Ortner}, \citenamefont {Csányi},\ and\
  \citenamefont {Drautz}}]{Bochkarev_2022}%
  \BibitemOpen
  \bibfield  {author} {\bibinfo {author} {\bibfnamefont {A.}~\bibnamefont
  {Bochkarev}}, \bibinfo {author} {\bibfnamefont {Y.}~\bibnamefont
  {Lysogorskiy}}, \bibinfo {author} {\bibfnamefont {C.}~\bibnamefont {Ortner}},
  \bibinfo {author} {\bibfnamefont {G.}~\bibnamefont {Csányi}}, \ and\
  \bibinfo {author} {\bibfnamefont {R.}~\bibnamefont {Drautz}},\ }\href
  {\doibase 10.48550/ARXIV.2205.08177} {\enquote {\bibinfo {title} {Multilayer
  atomic cluster expansion for semi-local interactions},}\ } (\bibinfo {year}
  {2022}{\natexlab{a}})\BibitemShut {NoStop}%
\bibitem [{\citenamefont {Kondor}(2018)}]{Risi_N_bodyNet2018}%
  \BibitemOpen
  \bibfield  {author} {\bibinfo {author} {\bibfnamefont {R.}~\bibnamefont
  {Kondor}},\ }\href {\doibase 10.48550/ARXIV.1803.01588} {\enquote {\bibinfo
  {title} {N-body networks: a covariant hierarchical neural network
  architecture for learning atomic potentials},}\ } (\bibinfo {year}
  {2018})\BibitemShut {NoStop}%
\bibitem [{\citenamefont {Battaglia}\ \emph {et~al.}(2018)\citenamefont
  {Battaglia}, \citenamefont {Hamrick}, \citenamefont {Bapst}, \citenamefont
  {Sanchez-Gonzalez}, \citenamefont {Zambaldi}, \citenamefont {Malinowski},
  \citenamefont {Tacchetti}, \citenamefont {Raposo}, \citenamefont {Santoro},
  \citenamefont {Faulkner}, \citenamefont {Gulcehre}, \citenamefont {Song},
  \citenamefont {Ballard}, \citenamefont {Gilmer}, \citenamefont {Dahl},
  \citenamefont {Vaswani}, \citenamefont {Allen}, \citenamefont {Nash},
  \citenamefont {Langston}, \citenamefont {Dyer}, \citenamefont {Heess},
  \citenamefont {Wierstra}, \citenamefont {Kohli}, \citenamefont {Botvinick},
  \citenamefont {Vinyals}, \citenamefont {Li},\ and\ \citenamefont
  {Pascanu}}]{GNNBattagial2018}%
  \BibitemOpen
  \bibfield  {author} {\bibinfo {author} {\bibfnamefont {P.~W.}\ \bibnamefont
  {Battaglia}}, \bibinfo {author} {\bibfnamefont {J.~B.}\ \bibnamefont
  {Hamrick}}, \bibinfo {author} {\bibfnamefont {V.}~\bibnamefont {Bapst}},
  \bibinfo {author} {\bibfnamefont {A.}~\bibnamefont {Sanchez-Gonzalez}},
  \bibinfo {author} {\bibfnamefont {V.}~\bibnamefont {Zambaldi}}, \bibinfo
  {author} {\bibfnamefont {M.}~\bibnamefont {Malinowski}}, \bibinfo {author}
  {\bibfnamefont {A.}~\bibnamefont {Tacchetti}}, \bibinfo {author}
  {\bibfnamefont {D.}~\bibnamefont {Raposo}}, \bibinfo {author} {\bibfnamefont
  {A.}~\bibnamefont {Santoro}}, \bibinfo {author} {\bibfnamefont
  {R.}~\bibnamefont {Faulkner}}, \bibinfo {author} {\bibfnamefont
  {C.}~\bibnamefont {Gulcehre}}, \bibinfo {author} {\bibfnamefont
  {F.}~\bibnamefont {Song}}, \bibinfo {author} {\bibfnamefont {A.}~\bibnamefont
  {Ballard}}, \bibinfo {author} {\bibfnamefont {J.}~\bibnamefont {Gilmer}},
  \bibinfo {author} {\bibfnamefont {G.}~\bibnamefont {Dahl}}, \bibinfo {author}
  {\bibfnamefont {A.}~\bibnamefont {Vaswani}}, \bibinfo {author} {\bibfnamefont
  {K.}~\bibnamefont {Allen}}, \bibinfo {author} {\bibfnamefont
  {C.}~\bibnamefont {Nash}}, \bibinfo {author} {\bibfnamefont {V.}~\bibnamefont
  {Langston}}, \bibinfo {author} {\bibfnamefont {C.}~\bibnamefont {Dyer}},
  \bibinfo {author} {\bibfnamefont {N.}~\bibnamefont {Heess}}, \bibinfo
  {author} {\bibfnamefont {D.}~\bibnamefont {Wierstra}}, \bibinfo {author}
  {\bibfnamefont {P.}~\bibnamefont {Kohli}}, \bibinfo {author} {\bibfnamefont
  {M.}~\bibnamefont {Botvinick}}, \bibinfo {author} {\bibfnamefont
  {O.}~\bibnamefont {Vinyals}}, \bibinfo {author} {\bibfnamefont
  {Y.}~\bibnamefont {Li}}, \ and\ \bibinfo {author} {\bibfnamefont
  {R.}~\bibnamefont {Pascanu}},\ }\href {\doibase 10.48550/ARXIV.1806.01261}
  {\enquote {\bibinfo {title} {Relational inductive biases, deep learning, and
  graph networks},}\ } (\bibinfo {year} {2018})\BibitemShut {NoStop}%
\bibitem [{\citenamefont {Gilmer}\ \emph {et~al.}(2017)\citenamefont {Gilmer},
  \citenamefont {Schoenholz}, \citenamefont {Riley}, \citenamefont {Vinyals},\
  and\ \citenamefont {Dahl}}]{MPNNGilmer2017}%
  \BibitemOpen
  \bibfield  {author} {\bibinfo {author} {\bibfnamefont {J.}~\bibnamefont
  {Gilmer}}, \bibinfo {author} {\bibfnamefont {S.~S.}\ \bibnamefont
  {Schoenholz}}, \bibinfo {author} {\bibfnamefont {P.~F.}\ \bibnamefont
  {Riley}}, \bibinfo {author} {\bibfnamefont {O.}~\bibnamefont {Vinyals}}, \
  and\ \bibinfo {author} {\bibfnamefont {G.~E.}\ \bibnamefont {Dahl}},\ }\href
  {\doibase 10.48550/ARXIV.1704.01212} {\enquote {\bibinfo {title} {Neural
  message passing for quantum chemistry},}\ } (\bibinfo {year}
  {2017})\BibitemShut {NoStop}%
\bibitem [{\citenamefont {Bronstein}\ \emph {et~al.}(2021)\citenamefont
  {Bronstein}, \citenamefont {Bruna}, \citenamefont {Cohen},\ and\
  \citenamefont {Veličković}}]{bronstein2021geometric}%
  \BibitemOpen
  \bibfield  {author} {\bibinfo {author} {\bibfnamefont {M.~M.}\ \bibnamefont
  {Bronstein}}, \bibinfo {author} {\bibfnamefont {J.}~\bibnamefont {Bruna}},
  \bibinfo {author} {\bibfnamefont {T.}~\bibnamefont {Cohen}}, \ and\ \bibinfo
  {author} {\bibfnamefont {P.}~\bibnamefont {Veličković}},\ }\href@noop {}
  {\enquote {\bibinfo {title} {Geometric deep learning: Grids, groups, graphs,
  geodesics, and gauges},}\ } (\bibinfo {year} {2021}),\ \Eprint
  {http://arxiv.org/abs/2104.13478} {arXiv:2104.13478 [cs.LG]} \BibitemShut
  {NoStop}%
\bibitem [{\citenamefont {Thompson}\ \emph {et~al.}(2015)\citenamefont
  {Thompson}, \citenamefont {Swiler}, \citenamefont {Trott}, \citenamefont
  {Foiles},\ and\ \citenamefont {Tucker}}]{THOMPSON2015SNAP}%
  \BibitemOpen
  \bibfield  {author} {\bibinfo {author} {\bibfnamefont {A.}~\bibnamefont
  {Thompson}}, \bibinfo {author} {\bibfnamefont {L.}~\bibnamefont {Swiler}},
  \bibinfo {author} {\bibfnamefont {C.}~\bibnamefont {Trott}}, \bibinfo
  {author} {\bibfnamefont {S.}~\bibnamefont {Foiles}}, \ and\ \bibinfo {author}
  {\bibfnamefont {G.}~\bibnamefont {Tucker}},\ }\href {\doibase
  https://doi.org/10.1016/j.jcp.2014.12.018} {\bibfield  {journal} {\bibinfo
  {journal} {Journal of Computational Physics}\ }\textbf {\bibinfo {volume}
  {285}},\ \bibinfo {pages} {316} (\bibinfo {year} {2015})}\BibitemShut
  {NoStop}%
\bibitem [{\citenamefont {Lysogorskiy}\ \emph {et~al.}(2021)\citenamefont
  {Lysogorskiy}, \citenamefont {Oord}, \citenamefont {Bochkarev}, \citenamefont
  {Menon}, \citenamefont {Rinaldi}, \citenamefont {Hammerschmidt},
  \citenamefont {Mrovec}, \citenamefont {Thompson}, \citenamefont {Csányi},
  \citenamefont {Ortner},\ and\ \citenamefont
  {Drautz}}]{lysogorskiy_performant_2021}%
  \BibitemOpen
  \bibfield  {author} {\bibinfo {author} {\bibfnamefont {Y.}~\bibnamefont
  {Lysogorskiy}}, \bibinfo {author} {\bibfnamefont {C.~v.~d.}\ \bibnamefont
  {Oord}}, \bibinfo {author} {\bibfnamefont {A.}~\bibnamefont {Bochkarev}},
  \bibinfo {author} {\bibfnamefont {S.}~\bibnamefont {Menon}}, \bibinfo
  {author} {\bibfnamefont {M.}~\bibnamefont {Rinaldi}}, \bibinfo {author}
  {\bibfnamefont {T.}~\bibnamefont {Hammerschmidt}}, \bibinfo {author}
  {\bibfnamefont {M.}~\bibnamefont {Mrovec}}, \bibinfo {author} {\bibfnamefont
  {A.}~\bibnamefont {Thompson}}, \bibinfo {author} {\bibfnamefont
  {G.}~\bibnamefont {Csányi}}, \bibinfo {author} {\bibfnamefont
  {C.}~\bibnamefont {Ortner}}, \ and\ \bibinfo {author} {\bibfnamefont
  {R.}~\bibnamefont {Drautz}},\ }\href {\doibase 10.1038/s41524-021-00559-9}
  {\bibfield  {journal} {\bibinfo  {journal} {npj Comput Mater}\ }\textbf
  {\bibinfo {volume} {7}} (\bibinfo {year} {2021}),\
  10.1038/s41524-021-00559-9}\BibitemShut {NoStop}%
\bibitem [{\citenamefont {Nigam}\ \emph {et~al.}(2020)\citenamefont {Nigam},
  \citenamefont {Pozdnyakov},\ and\ \citenamefont {Ceriotti}}]{NICE}%
  \BibitemOpen
  \bibfield  {author} {\bibinfo {author} {\bibfnamefont {J.}~\bibnamefont
  {Nigam}}, \bibinfo {author} {\bibfnamefont {S.}~\bibnamefont {Pozdnyakov}}, \
  and\ \bibinfo {author} {\bibfnamefont {M.}~\bibnamefont {Ceriotti}},\ }\href
  {\doibase 10.1063/5.0021116} {\bibfield  {journal} {\bibinfo  {journal} {The
  Journal of Chemical Physics}\ }\textbf {\bibinfo {volume} {153}},\ \bibinfo
  {pages} {121101} (\bibinfo {year} {2020})},\ \Eprint
  {http://arxiv.org/abs/https://doi.org/10.1063/5.0021116}
  {https://doi.org/10.1063/5.0021116} \BibitemShut {NoStop}%
\bibitem [{\citenamefont {Thomas}\ \emph {et~al.}(2021)\citenamefont {Thomas},
  \citenamefont {Chen},\ and\ \citenamefont {Ortner}}]{thomas2021rigorous}%
  \BibitemOpen
  \bibfield  {author} {\bibinfo {author} {\bibfnamefont {J.}~\bibnamefont
  {Thomas}}, \bibinfo {author} {\bibfnamefont {H.}~\bibnamefont {Chen}}, \ and\
  \bibinfo {author} {\bibfnamefont {C.}~\bibnamefont {Ortner}},\ }\href@noop {}
  {\enquote {\bibinfo {title} {Rigorous body-order approximations of an
  electronic structure potential energy landscape},}\ } (\bibinfo {year}
  {2021}),\ \Eprint {http://arxiv.org/abs/2106.12572} {arXiv:2106.12572
  [math-ph]} \BibitemShut {NoStop}%
\bibitem [{\citenamefont {Drautz}\ and\ \citenamefont
  {Pettifor}(2006)}]{Drautz06}%
  \BibitemOpen
  \bibfield  {author} {\bibinfo {author} {\bibfnamefont {R.}~\bibnamefont
  {Drautz}}\ and\ \bibinfo {author} {\bibfnamefont {D.~G.}\ \bibnamefont
  {Pettifor}},\ }\href {\doibase 10.1103/physrevb.74.174117} {\bibfield
  {journal} {\bibinfo  {journal} {Phys. Rev. B}\ }\textbf {\bibinfo {volume}
  {74}} (\bibinfo {year} {2006}),\ 10.1103/physrevb.74.174117}\BibitemShut
  {NoStop}%
\bibitem [{\citenamefont {van~der Oord}\ \emph {et~al.}(2020)\citenamefont
  {van~der Oord}, \citenamefont {Dusson}, \citenamefont {Cs{\'{a}}nyi},\ and\
  \citenamefont {Ortner}}]{aPIPs2020}%
  \BibitemOpen
  \bibfield  {author} {\bibinfo {author} {\bibfnamefont {C.}~\bibnamefont
  {van~der Oord}}, \bibinfo {author} {\bibfnamefont {G.}~\bibnamefont
  {Dusson}}, \bibinfo {author} {\bibfnamefont {G.}~\bibnamefont
  {Cs{\'{a}}nyi}}, \ and\ \bibinfo {author} {\bibfnamefont {C.}~\bibnamefont
  {Ortner}},\ }\href {\doibase 10.1088/2632-2153/ab527c} {\bibfield  {journal}
  {\bibinfo  {journal} {Machine Learning: Science and Technology}\ }\textbf
  {\bibinfo {volume} {1}},\ \bibinfo {pages} {015004} (\bibinfo {year}
  {2020})}\BibitemShut {NoStop}%
\bibitem [{\citenamefont {{Drautz}}\ \emph {et~al.}(2004)\citenamefont
  {{Drautz}}, \citenamefont {{F{\"a}hnle}},\ and\ \citenamefont
  {{Sanchez}}}]{Drautz04}%
  \BibitemOpen
  \bibfield  {author} {\bibinfo {author} {\bibfnamefont {R.}~\bibnamefont
  {{Drautz}}}, \bibinfo {author} {\bibfnamefont {M.}~\bibnamefont
  {{F{\"a}hnle}}}, \ and\ \bibinfo {author} {\bibfnamefont {J.~M.}\
  \bibnamefont {{Sanchez}}},\ }\href {\doibase 10.1088/0953-8984/16/23/005}
  {\bibfield  {journal} {\bibinfo  {journal} {J. Phys.: Condens. Matter}\
  }\textbf {\bibinfo {volume} {16}},\ \bibinfo {pages} {3843} (\bibinfo {year}
  {2004})}\BibitemShut {NoStop}%
\bibitem [{\citenamefont {Musil}\ \emph
  {et~al.}(2021{\natexlab{b}})\citenamefont {Musil}, \citenamefont {Grisafi},
  \citenamefont {Bartók}, \citenamefont {Ortner}, \citenamefont {Csányi},\
  and\ \citenamefont {Ceriotti}}]{Ceriotti2021Review}%
  \BibitemOpen
  \bibfield  {author} {\bibinfo {author} {\bibfnamefont {F.}~\bibnamefont
  {Musil}}, \bibinfo {author} {\bibfnamefont {A.}~\bibnamefont {Grisafi}},
  \bibinfo {author} {\bibfnamefont {A.~P.}\ \bibnamefont {Bartók}}, \bibinfo
  {author} {\bibfnamefont {C.}~\bibnamefont {Ortner}}, \bibinfo {author}
  {\bibfnamefont {G.}~\bibnamefont {Csányi}}, \ and\ \bibinfo {author}
  {\bibfnamefont {M.}~\bibnamefont {Ceriotti}},\ }\href {\doibase
  10.1021/acs.chemrev.1c00021} {\bibfield  {journal} {\bibinfo  {journal}
  {Chemical Reviews}\ }\textbf {\bibinfo {volume} {121}},\ \bibinfo {pages}
  {9759} (\bibinfo {year} {2021}{\natexlab{b}})},\ \bibinfo {note} {pMID:
  34310133},\ \Eprint
  {http://arxiv.org/abs/https://doi.org/10.1021/acs.chemrev.1c00021}
  {https://doi.org/10.1021/acs.chemrev.1c00021} \BibitemShut {NoStop}%
\bibitem [{\citenamefont {Kaliuzhnyi}\ and\ \citenamefont
  {Ortner}(2022)}]{2022-recursive}%
  \BibitemOpen
  \bibfield  {author} {\bibinfo {author} {\bibfnamefont {I.}~\bibnamefont
  {Kaliuzhnyi}}\ and\ \bibinfo {author} {\bibfnamefont {C.}~\bibnamefont
  {Ortner}},\ }\href {https://arxiv.org/abs/2202.04140} {\bibfield  {journal}
  {\bibinfo  {journal} {ArXiv e-prints}\ }\textbf {\bibinfo {volume}
  {2202.04140}} (\bibinfo {year} {2022})}\BibitemShut {NoStop}%
\bibitem [{\citenamefont {Zhang}\ \emph {et~al.}(2022)\citenamefont {Zhang},
  \citenamefont {Onat}, \citenamefont {Dusson}, \citenamefont {Anand},
  \citenamefont {Maurer}, \citenamefont {Ortner},\ and\ \citenamefont
  {Kermode}}]{zhang2022equivariantACE}%
  \BibitemOpen
  \bibfield  {author} {\bibinfo {author} {\bibfnamefont {L.}~\bibnamefont
  {Zhang}}, \bibinfo {author} {\bibfnamefont {B.}~\bibnamefont {Onat}},
  \bibinfo {author} {\bibfnamefont {G.}~\bibnamefont {Dusson}}, \bibinfo
  {author} {\bibfnamefont {G.}~\bibnamefont {Anand}}, \bibinfo {author}
  {\bibfnamefont {R.~J.}\ \bibnamefont {Maurer}}, \bibinfo {author}
  {\bibfnamefont {C.}~\bibnamefont {Ortner}}, \ and\ \bibinfo {author}
  {\bibfnamefont {J.~R.}\ \bibnamefont {Kermode}},\ }\href@noop {} {\enquote
  {\bibinfo {title} {Equivariant analytical mapping of first principles
  hamiltonians to accurate and transferable materials models},}\ } (\bibinfo
  {year} {2022}),\ \Eprint {http://arxiv.org/abs/2111.13736} {arXiv:2111.13736
  [cond-mat.mtrl-sci]} \BibitemShut {NoStop}%
\bibitem [{\citenamefont {Batatia}\ \emph {et~al.}(2022)\citenamefont
  {Batatia}, \citenamefont {Kovács}, \citenamefont {Simm}, \citenamefont
  {Ortner},\ and\ \citenamefont {Csányi}}]{MACE2022}%
  \BibitemOpen
  \bibfield  {author} {\bibinfo {author} {\bibfnamefont {I.}~\bibnamefont
  {Batatia}}, \bibinfo {author} {\bibfnamefont {D.~P.}\ \bibnamefont
  {Kovács}}, \bibinfo {author} {\bibfnamefont {G.~N.~C.}\ \bibnamefont
  {Simm}}, \bibinfo {author} {\bibfnamefont {C.}~\bibnamefont {Ortner}}, \ and\
  \bibinfo {author} {\bibfnamefont {G.}~\bibnamefont {Csányi}},\ }\href
  {\doibase 10.48550/ARXIV.2206.07697} {\enquote {\bibinfo {title} {Mace:
  Higher order equivariant message passing neural networks for fast and
  accurate force fields},}\ } (\bibinfo {year} {2022})\BibitemShut {NoStop}%
\bibitem [{\citenamefont {Anderson}\ \emph
  {et~al.}(2019{\natexlab{b}})\citenamefont {Anderson}, \citenamefont {Hy},\
  and\ \citenamefont {Kondor}}]{Anderson2019CormorantCM}%
  \BibitemOpen
  \bibfield  {author} {\bibinfo {author} {\bibfnamefont {B.}~\bibnamefont
  {Anderson}}, \bibinfo {author} {\bibfnamefont {T.~S.}\ \bibnamefont {Hy}}, \
  and\ \bibinfo {author} {\bibfnamefont {R.}~\bibnamefont {Kondor}},\ }in\
  \href
  {https://proceedings.neurips.cc/paper/2019/file/03573b32b2746e6e8ca98b9123f2249b-Paper.pdf}
  {\emph {\bibinfo {booktitle} {Advances in Neural Information Processing
  Systems}}},\ Vol.~\bibinfo {volume} {32},\ \bibinfo {editor} {edited by\
  \bibinfo {editor} {\bibfnamefont {H.}~\bibnamefont {Wallach}}, \bibinfo
  {editor} {\bibfnamefont {H.}~\bibnamefont {Larochelle}}, \bibinfo {editor}
  {\bibfnamefont {A.}~\bibnamefont {Beygelzimer}}, \bibinfo {editor}
  {\bibfnamefont {F.}~\bibnamefont {d\textquotesingle Alch\'{e}-Buc}}, \bibinfo
  {editor} {\bibfnamefont {E.}~\bibnamefont {Fox}}, \ and\ \bibinfo {editor}
  {\bibfnamefont {R.}~\bibnamefont {Garnett}}}\ (\bibinfo  {publisher} {Curran
  Associates, Inc.},\ \bibinfo {year} {2019})\BibitemShut {NoStop}%
\bibitem [{\citenamefont {Christensen}\ and\ \citenamefont {Anatole~von
  Lilienfeld}(2020)}]{Christensen2020}%
  \BibitemOpen
  \bibfield  {author} {\bibinfo {author} {\bibfnamefont {A.~S.}\ \bibnamefont
  {Christensen}}\ and\ \bibinfo {author} {\bibfnamefont {O.}~\bibnamefont
  {Anatole~von Lilienfeld}},\ }\href {\doibase 10.1088/2632-2153/abba6f}
  {\bibfield  {journal} {\bibinfo  {journal} {Machine Learning: Science and
  Technology}\ }\textbf {\bibinfo {volume} {1}} (\bibinfo {year} {2020}),\
  10.1088/2632-2153/abba6f}\BibitemShut {NoStop}%
\bibitem [{\citenamefont {Qu}\ \emph {et~al.}(2021{\natexlab{a}})\citenamefont
  {Qu}, \citenamefont {Conte}, \citenamefont {Houston},\ and\ \citenamefont
  {Bowman}}]{BowmanAcAc2021}%
  \BibitemOpen
  \bibfield  {author} {\bibinfo {author} {\bibfnamefont {C.}~\bibnamefont
  {Qu}}, \bibinfo {author} {\bibfnamefont {R.}~\bibnamefont {Conte}}, \bibinfo
  {author} {\bibfnamefont {P.~L.}\ \bibnamefont {Houston}}, \ and\ \bibinfo
  {author} {\bibfnamefont {J.~M.}\ \bibnamefont {Bowman}},\ }\href {\doibase
  10.1039/D0CP04221H} {\bibfield  {journal} {\bibinfo  {journal} {Phys. Chem.
  Chem. Phys.}\ }\textbf {\bibinfo {volume} {23}},\ \bibinfo {pages} {7758}
  (\bibinfo {year} {2021}{\natexlab{a}})}\BibitemShut {NoStop}%
\bibitem [{\citenamefont {Bannwarth}\ \emph {et~al.}(2019)\citenamefont
  {Bannwarth}, \citenamefont {Ehlert},\ and\ \citenamefont
  {Grimme}}]{GFN2-xTB}%
  \BibitemOpen
  \bibfield  {author} {\bibinfo {author} {\bibfnamefont {C.}~\bibnamefont
  {Bannwarth}}, \bibinfo {author} {\bibfnamefont {S.}~\bibnamefont {Ehlert}}, \
  and\ \bibinfo {author} {\bibfnamefont {S.}~\bibnamefont {Grimme}},\ }\href
  {\doibase 10.1021/acs.jctc.8b01176} {\bibfield  {journal} {\bibinfo
  {journal} {Journal of Chemical Theory and Computation}\ }\textbf {\bibinfo
  {volume} {15}},\ \bibinfo {pages} {1652} (\bibinfo {year} {2019})},\ \bibinfo
  {note} {pMID: 30741547},\ \Eprint
  {http://arxiv.org/abs/https://doi.org/10.1021/acs.jctc.8b01176}
  {https://doi.org/10.1021/acs.jctc.8b01176} \BibitemShut {NoStop}%
\bibitem [{\citenamefont {Geiger}\ \emph {et~al.}(2020)\citenamefont {Geiger},
  \citenamefont {Smidt}, \citenamefont {M.}, \citenamefont {Miller},
  \citenamefont {Boomsma}, \citenamefont {Dice}, \citenamefont {Lapchevskyi},
  \citenamefont {Weiler}, \citenamefont {Tyszkiewicz}, \citenamefont {Batzner},
  \citenamefont {Uhrin}, \citenamefont {Frellsen}, \citenamefont {Jung},
  \citenamefont {Sanborn}, \citenamefont {Rackers},\ and\ \citenamefont
  {Bailey}}]{e3nn}%
  \BibitemOpen
  \bibfield  {author} {\bibinfo {author} {\bibfnamefont {M.}~\bibnamefont
  {Geiger}}, \bibinfo {author} {\bibfnamefont {T.}~\bibnamefont {Smidt}},
  \bibinfo {author} {\bibfnamefont {A.}~\bibnamefont {M.}}, \bibinfo {author}
  {\bibfnamefont {B.~K.}\ \bibnamefont {Miller}}, \bibinfo {author}
  {\bibfnamefont {W.}~\bibnamefont {Boomsma}}, \bibinfo {author} {\bibfnamefont
  {B.}~\bibnamefont {Dice}}, \bibinfo {author} {\bibfnamefont {K.}~\bibnamefont
  {Lapchevskyi}}, \bibinfo {author} {\bibfnamefont {M.}~\bibnamefont {Weiler}},
  \bibinfo {author} {\bibfnamefont {M.}~\bibnamefont {Tyszkiewicz}}, \bibinfo
  {author} {\bibfnamefont {S.}~\bibnamefont {Batzner}}, \bibinfo {author}
  {\bibfnamefont {M.}~\bibnamefont {Uhrin}}, \bibinfo {author} {\bibfnamefont
  {J.}~\bibnamefont {Frellsen}}, \bibinfo {author} {\bibfnamefont
  {N.}~\bibnamefont {Jung}}, \bibinfo {author} {\bibfnamefont {S.}~\bibnamefont
  {Sanborn}}, \bibinfo {author} {\bibfnamefont {J.}~\bibnamefont {Rackers}}, \
  and\ \bibinfo {author} {\bibfnamefont {M.}~\bibnamefont {Bailey}},\ }\href
  {\doibase 10.5281/zenodo.5292912} {\enquote {\bibinfo {title} {Euclidean
  neural networks: e3nn},}\ } (\bibinfo {year} {2020})\BibitemShut {NoStop}%
\bibitem [{\citenamefont {Paszke}\ \emph {et~al.}(2019)\citenamefont {Paszke},
  \citenamefont {Gross}, \citenamefont {Massa}, \citenamefont {Lerer},
  \citenamefont {Bradbury}, \citenamefont {Chanan}, \citenamefont {Killeen},
  \citenamefont {Lin}, \citenamefont {Gimelshein}, \citenamefont {Antiga} \emph
  {et~al.}}]{paszke2019pytorch}%
  \BibitemOpen
  \bibfield  {author} {\bibinfo {author} {\bibfnamefont {A.}~\bibnamefont
  {Paszke}}, \bibinfo {author} {\bibfnamefont {S.}~\bibnamefont {Gross}},
  \bibinfo {author} {\bibfnamefont {F.}~\bibnamefont {Massa}}, \bibinfo
  {author} {\bibfnamefont {A.}~\bibnamefont {Lerer}}, \bibinfo {author}
  {\bibfnamefont {J.}~\bibnamefont {Bradbury}}, \bibinfo {author}
  {\bibfnamefont {G.}~\bibnamefont {Chanan}}, \bibinfo {author} {\bibfnamefont
  {T.}~\bibnamefont {Killeen}}, \bibinfo {author} {\bibfnamefont
  {Z.}~\bibnamefont {Lin}}, \bibinfo {author} {\bibfnamefont {N.}~\bibnamefont
  {Gimelshein}}, \bibinfo {author} {\bibfnamefont {L.}~\bibnamefont {Antiga}},
  \emph {et~al.},\ }in\ \href@noop {} {\emph {\bibinfo {booktitle} {Advances in
  neural information processing systems}}}\ (\bibinfo {year} {2019})\ pp.\
  \bibinfo {pages} {8026--8037}\BibitemShut {NoStop}%
\bibitem [{\citenamefont {Allen-Zhu}\ \emph {et~al.}(2018)\citenamefont
  {Allen-Zhu}, \citenamefont {Li},\ and\ \citenamefont
  {Liang}}]{OverparametrisedNN2020}%
  \BibitemOpen
  \bibfield  {author} {\bibinfo {author} {\bibfnamefont {Z.}~\bibnamefont
  {Allen-Zhu}}, \bibinfo {author} {\bibfnamefont {Y.}~\bibnamefont {Li}}, \
  and\ \bibinfo {author} {\bibfnamefont {Y.}~\bibnamefont {Liang}},\ }\href
  {\doibase 10.48550/ARXIV.1811.04918} {\enquote {\bibinfo {title} {Learning
  and generalization in overparameterized neural networks, going beyond two
  layers},}\ } (\bibinfo {year} {2018})\BibitemShut {NoStop}%
\bibitem [{\citenamefont {Caro}(2019)}]{Miguel2019TurboSOAP}%
  \BibitemOpen
  \bibfield  {author} {\bibinfo {author} {\bibfnamefont {M.~A.}\ \bibnamefont
  {Caro}},\ }\href {\doibase 10.1103/PhysRevB.100.024112} {\bibfield  {journal}
  {\bibinfo  {journal} {Phys. Rev. B}\ }\textbf {\bibinfo {volume} {100}},\
  \bibinfo {pages} {024112} (\bibinfo {year} {2019})}\BibitemShut {NoStop}%
\bibitem [{\citenamefont {Musil}\ \emph
  {et~al.}(2021{\natexlab{c}})\citenamefont {Musil}, \citenamefont {Veit},
  \citenamefont {Goscinski}, \citenamefont {Fraux}, \citenamefont {Willatt},
  \citenamefont {Stricker}, \citenamefont {Junge},\ and\ \citenamefont
  {Ceriotti}}]{Ceriotti2021Librascal}%
  \BibitemOpen
  \bibfield  {author} {\bibinfo {author} {\bibfnamefont {F.}~\bibnamefont
  {Musil}}, \bibinfo {author} {\bibfnamefont {M.}~\bibnamefont {Veit}},
  \bibinfo {author} {\bibfnamefont {A.}~\bibnamefont {Goscinski}}, \bibinfo
  {author} {\bibfnamefont {G.}~\bibnamefont {Fraux}}, \bibinfo {author}
  {\bibfnamefont {M.~J.}\ \bibnamefont {Willatt}}, \bibinfo {author}
  {\bibfnamefont {M.}~\bibnamefont {Stricker}}, \bibinfo {author}
  {\bibfnamefont {T.}~\bibnamefont {Junge}}, \ and\ \bibinfo {author}
  {\bibfnamefont {M.}~\bibnamefont {Ceriotti}},\ }\href {\doibase
  10.1063/5.0044689} {\bibfield  {journal} {\bibinfo  {journal} {The Journal of
  Chemical Physics}\ }\textbf {\bibinfo {volume} {154}},\ \bibinfo {pages}
  {114109} (\bibinfo {year} {2021}{\natexlab{c}})},\ \Eprint
  {http://arxiv.org/abs/https://doi.org/10.1063/5.0044689}
  {https://doi.org/10.1063/5.0044689} \BibitemShut {NoStop}%
\bibitem [{\citenamefont {Himanen}\ \emph {et~al.}(2020)\citenamefont
  {Himanen}, \citenamefont {Jäger}, \citenamefont {Morooka}, \citenamefont
  {{Federici Canova}}, \citenamefont {Ranawat}, \citenamefont {Gao},
  \citenamefont {Rinke},\ and\ \citenamefont {Foster}}]{Foster2020Dscribe}%
  \BibitemOpen
  \bibfield  {author} {\bibinfo {author} {\bibfnamefont {L.}~\bibnamefont
  {Himanen}}, \bibinfo {author} {\bibfnamefont {M.~O.}\ \bibnamefont {Jäger}},
  \bibinfo {author} {\bibfnamefont {E.~V.}\ \bibnamefont {Morooka}}, \bibinfo
  {author} {\bibfnamefont {F.}~\bibnamefont {{Federici Canova}}}, \bibinfo
  {author} {\bibfnamefont {Y.~S.}\ \bibnamefont {Ranawat}}, \bibinfo {author}
  {\bibfnamefont {D.~Z.}\ \bibnamefont {Gao}}, \bibinfo {author} {\bibfnamefont
  {P.}~\bibnamefont {Rinke}}, \ and\ \bibinfo {author} {\bibfnamefont {A.~S.}\
  \bibnamefont {Foster}},\ }\href {\doibase
  https://doi.org/10.1016/j.cpc.2019.106949} {\bibfield  {journal} {\bibinfo
  {journal} {Computer Physics Communications}\ }\textbf {\bibinfo {volume}
  {247}},\ \bibinfo {pages} {106949} (\bibinfo {year} {2020})}\BibitemShut
  {NoStop}%
\bibitem [{\citenamefont {Goscinski}\ \emph {et~al.}(2021)\citenamefont
  {Goscinski}, \citenamefont {Musil}, \citenamefont {Pozdnyakov}, \citenamefont
  {Nigam},\ and\ \citenamefont {Ceriotti}}]{CeriottiOptRadBasis2021}%
  \BibitemOpen
  \bibfield  {author} {\bibinfo {author} {\bibfnamefont {A.}~\bibnamefont
  {Goscinski}}, \bibinfo {author} {\bibfnamefont {F.}~\bibnamefont {Musil}},
  \bibinfo {author} {\bibfnamefont {S.}~\bibnamefont {Pozdnyakov}}, \bibinfo
  {author} {\bibfnamefont {J.}~\bibnamefont {Nigam}}, \ and\ \bibinfo {author}
  {\bibfnamefont {M.}~\bibnamefont {Ceriotti}},\ }\href {\doibase
  10.1063/5.0057229} {\bibfield  {journal} {\bibinfo  {journal} {The Journal of
  Chemical Physics}\ }\textbf {\bibinfo {volume} {155}},\ \bibinfo {pages}
  {104106} (\bibinfo {year} {2021})},\ \Eprint
  {http://arxiv.org/abs/https://doi.org/10.1063/5.0057229}
  {https://doi.org/10.1063/5.0057229} \BibitemShut {NoStop}%
\bibitem [{\citenamefont {Bochkarev}\ \emph
  {et~al.}(2022{\natexlab{b}})\citenamefont {Bochkarev}, \citenamefont
  {Lysogorskiy}, \citenamefont {Menon}, \citenamefont {Qamar}, \citenamefont
  {Mrovec},\ and\ \citenamefont {Drautz}}]{bochkarev_efficient_2022}%
  \BibitemOpen
  \bibfield  {author} {\bibinfo {author} {\bibfnamefont {A.}~\bibnamefont
  {Bochkarev}}, \bibinfo {author} {\bibfnamefont {Y.}~\bibnamefont
  {Lysogorskiy}}, \bibinfo {author} {\bibfnamefont {S.}~\bibnamefont {Menon}},
  \bibinfo {author} {\bibfnamefont {M.}~\bibnamefont {Qamar}}, \bibinfo
  {author} {\bibfnamefont {M.}~\bibnamefont {Mrovec}}, \ and\ \bibinfo {author}
  {\bibfnamefont {R.}~\bibnamefont {Drautz}},\ }\href {\doibase
  10.1103/physrevmaterials.6.013804} {\bibfield  {journal} {\bibinfo  {journal}
  {Phys. Rev. Materials}\ }\textbf {\bibinfo {volume} {6}} (\bibinfo {year}
  {2022}{\natexlab{b}}),\ 10.1103/physrevmaterials.6.013804}\BibitemShut
  {NoStop}%
\bibitem [{\citenamefont {Elfwing}\ \emph {et~al.}(2017)\citenamefont
  {Elfwing}, \citenamefont {Uchibe},\ and\ \citenamefont {Doya}}]{SiLU2017}%
  \BibitemOpen
  \bibfield  {author} {\bibinfo {author} {\bibfnamefont {S.}~\bibnamefont
  {Elfwing}}, \bibinfo {author} {\bibfnamefont {E.}~\bibnamefont {Uchibe}}, \
  and\ \bibinfo {author} {\bibfnamefont {K.}~\bibnamefont {Doya}},\ }\href
  {http://arxiv.org/abs/1702.03118} {\bibfield  {journal} {\bibinfo  {journal}
  {CoRR}\ }\textbf {\bibinfo {volume} {abs/1702.03118}} (\bibinfo {year}
  {2017})},\ \Eprint {http://arxiv.org/abs/1702.03118} {1702.03118}
  \BibitemShut {NoStop}%
\bibitem [{\citenamefont {He}\ \emph {et~al.}(2015)\citenamefont {He},
  \citenamefont {Zhang}, \citenamefont {Ren},\ and\ \citenamefont
  {Sun}}]{resnet2015}%
  \BibitemOpen
  \bibfield  {author} {\bibinfo {author} {\bibfnamefont {K.}~\bibnamefont
  {He}}, \bibinfo {author} {\bibfnamefont {X.}~\bibnamefont {Zhang}}, \bibinfo
  {author} {\bibfnamefont {S.}~\bibnamefont {Ren}}, \ and\ \bibinfo {author}
  {\bibfnamefont {J.}~\bibnamefont {Sun}},\ }\href@noop {} {\enquote {\bibinfo
  {title} {Deep residual learning for image recognition},}\ } (\bibinfo {year}
  {2015}),\ \Eprint {http://arxiv.org/abs/1512.03385} {arXiv:1512.03385
  [cs.CV]} \BibitemShut {NoStop}%
\bibitem [{\citenamefont {Ioffe}\ and\ \citenamefont
  {Szegedy}(2015)}]{ioffe2015batch}%
  \BibitemOpen
  \bibfield  {author} {\bibinfo {author} {\bibfnamefont {S.}~\bibnamefont
  {Ioffe}}\ and\ \bibinfo {author} {\bibfnamefont {C.}~\bibnamefont
  {Szegedy}},\ }\href@noop {} {\enquote {\bibinfo {title} {Batch normalization:
  Accelerating deep network training by reducing internal covariate shift},}\ }
  (\bibinfo {year} {2015}),\ \Eprint {http://arxiv.org/abs/1502.03167}
  {arXiv:1502.03167 [cs.LG]} \BibitemShut {NoStop}%
\bibitem [{\citenamefont {Lecun}\ \emph {et~al.}(1998)\citenamefont {Lecun},
  \citenamefont {Bottou}, \citenamefont {Orr},\ and\ \citenamefont
  {Müller}}]{Lecun98efficientbackprop}%
  \BibitemOpen
  \bibfield  {author} {\bibinfo {author} {\bibfnamefont {Y.}~\bibnamefont
  {Lecun}}, \bibinfo {author} {\bibfnamefont {L.}~\bibnamefont {Bottou}},
  \bibinfo {author} {\bibfnamefont {G.~B.}\ \bibnamefont {Orr}}, \ and\
  \bibinfo {author} {\bibfnamefont {K.-R.}\ \bibnamefont {Müller}},\
  }\href@noop {} {\enquote {\bibinfo {title} {Efficient backprop},}\ }
  (\bibinfo {year} {1998})\BibitemShut {NoStop}%
\bibitem [{\citenamefont {Chmiela}\ \emph {et~al.}(2017)\citenamefont
  {Chmiela}, \citenamefont {Tkatchenko}, \citenamefont {Sauceda}, \citenamefont
  {Poltavsky}, \citenamefont {Sch{\"u}tt},\ and\ \citenamefont
  {M{\"u}ller}}]{chmiela2017machine}%
  \BibitemOpen
  \bibfield  {author} {\bibinfo {author} {\bibfnamefont {S.}~\bibnamefont
  {Chmiela}}, \bibinfo {author} {\bibfnamefont {A.}~\bibnamefont {Tkatchenko}},
  \bibinfo {author} {\bibfnamefont {H.~E.}\ \bibnamefont {Sauceda}}, \bibinfo
  {author} {\bibfnamefont {I.}~\bibnamefont {Poltavsky}}, \bibinfo {author}
  {\bibfnamefont {K.~T.}\ \bibnamefont {Sch{\"u}tt}}, \ and\ \bibinfo {author}
  {\bibfnamefont {K.-R.}\ \bibnamefont {M{\"u}ller}},\ }\href@noop {}
  {\bibfield  {journal} {\bibinfo  {journal} {Science advances}\ }\textbf
  {\bibinfo {volume} {3}},\ \bibinfo {pages} {e1603015} (\bibinfo {year}
  {2017})}\BibitemShut {NoStop}%
\bibitem [{\citenamefont {Gao}\ \emph {et~al.}(2020)\citenamefont {Gao},
  \citenamefont {Ramezanghorbani}, \citenamefont {Isayev}, \citenamefont
  {Smith},\ and\ \citenamefont {Roitberg}}]{Isayev2020ANI}%
  \BibitemOpen
  \bibfield  {author} {\bibinfo {author} {\bibfnamefont {X.}~\bibnamefont
  {Gao}}, \bibinfo {author} {\bibfnamefont {F.}~\bibnamefont
  {Ramezanghorbani}}, \bibinfo {author} {\bibfnamefont {O.}~\bibnamefont
  {Isayev}}, \bibinfo {author} {\bibfnamefont {J.~S.}\ \bibnamefont {Smith}}, \
  and\ \bibinfo {author} {\bibfnamefont {A.~E.}\ \bibnamefont {Roitberg}},\
  }\href {\doibase 10.1021/acs.jcim.0c00451} {\bibfield  {journal} {\bibinfo
  {journal} {Journal of Chemical Information and Modeling}\ }\textbf {\bibinfo
  {volume} {60}},\ \bibinfo {pages} {3408} (\bibinfo {year} {2020})},\ \bibinfo
  {note} {pMID: 32568524},\ \Eprint
  {http://arxiv.org/abs/https://doi.org/10.1021/acs.jcim.0c00451}
  {https://doi.org/10.1021/acs.jcim.0c00451} \BibitemShut {NoStop}%
\bibitem [{\citenamefont {Zaverkin}\ and\ \citenamefont
  {Kästner}(2020)}]{zaverkin2020}%
  \BibitemOpen
  \bibfield  {author} {\bibinfo {author} {\bibfnamefont {V.}~\bibnamefont
  {Zaverkin}}\ and\ \bibinfo {author} {\bibfnamefont {J.}~\bibnamefont
  {Kästner}},\ }\href {\doibase 10.1021/acs.jctc.0c00347} {\bibfield
  {journal} {\bibinfo  {journal} {Journal of Chemical Theory and Computation}\
  }\textbf {\bibinfo {volume} {16}},\ \bibinfo {pages} {5410} (\bibinfo {year}
  {2020})}\BibitemShut {NoStop}%
\bibitem [{\citenamefont {Qu}\ \emph {et~al.}(2021{\natexlab{b}})\citenamefont
  {Qu}, \citenamefont {Houston}, \citenamefont {Conte}, \citenamefont {Nandi},\
  and\ \citenamefont {Bowman}}]{AcAcBowman}%
  \BibitemOpen
  \bibfield  {author} {\bibinfo {author} {\bibfnamefont {C.}~\bibnamefont
  {Qu}}, \bibinfo {author} {\bibfnamefont {P.~L.}\ \bibnamefont {Houston}},
  \bibinfo {author} {\bibfnamefont {R.}~\bibnamefont {Conte}}, \bibinfo
  {author} {\bibfnamefont {A.}~\bibnamefont {Nandi}}, \ and\ \bibinfo {author}
  {\bibfnamefont {J.~M.}\ \bibnamefont {Bowman}},\ }\href {\doibase
  10.1021/acs.jpclett.1c01142} {\bibfield  {journal} {\bibinfo  {journal} {The
  Journal of Physical Chemistry Letters}\ }\textbf {\bibinfo {volume} {12}},\
  \bibinfo {pages} {4902} (\bibinfo {year} {2021}{\natexlab{b}})},\ \bibinfo
  {note} {pMID: 34006096},\ \Eprint
  {http://arxiv.org/abs/https://doi.org/10.1021/acs.jpclett.1c01142}
  {https://doi.org/10.1021/acs.jpclett.1c01142} \BibitemShut {NoStop}%
\bibitem [{\citenamefont {Berne}\ \emph {et~al.}(1998)\citenamefont {Berne},
  \citenamefont {Ciccotti},\ and\ \citenamefont {Coker}}]{NEB1998}%
  \BibitemOpen
  \bibfield  {author} {\bibinfo {author} {\bibfnamefont {B.~J.}\ \bibnamefont
  {Berne}}, \bibinfo {author} {\bibfnamefont {G.}~\bibnamefont {Ciccotti}}, \
  and\ \bibinfo {author} {\bibfnamefont {D.~F.}\ \bibnamefont {Coker}},\ }\href
  {\doibase 10.1142/3816} {\emph {\bibinfo {title} {Classical and Quantum
  Dynamics in Condensed Phase Simulations}}}\ (\bibinfo  {publisher} {WORLD
  SCIENTIFIC},\ \bibinfo {year} {1998})\ \Eprint
  {http://arxiv.org/abs/https://www.worldscientific.com/doi/pdf/10.1142/3816}
  {https://www.worldscientific.com/doi/pdf/10.1142/3816} \BibitemShut {NoStop}%
\end{thebibliography}%

\section{Appendix}

\subsection{Generalized Clebsch-Gordan Coefficients and Spherical Coordinates}
\label{annex:spherical_vectors}

The generalised Clebsch-Gordan coefficients are defined as product of Clebsch-Gordan coefficients:
\begin{equation}
    \label{eq:generalized-clebsch-gordan}
    C^{LM}_{l_{1}m_{1},..,l_{n}m_{n} } = C^{L_{2}M_{2}}_{l_{1}m_{1},l_{2}m_{2}}C^{L_{3}M_{3}}_{L_{2}M_{2},l_{3}m_{3}} ... C^{L_{N}M_{N}}_{L_{N-1}M_{N-1},l_{N}m_{N}}
\end{equation}
and
\begin{multline}
    L \equiv (L_{2},..,L_{N}), \quad |l_{1} - l_{2}| \leq L_{2} \leq l_{1} + l_{2} \quad \text{and} \\ \forall 3 \leq i |L_{i-1} - l_{i}| \leq L_{i} \leq L_{i-1} + l_{i}
\end{multline}

\begin{equation}
    M \in \{m_{i} | -l_{i}\leq m_{i} \leq l_{i}\}
\end{equation}

In equivariant networks it is usually more convenient to employ features in spherical coordinates.
A {\it spherical tensor} $t^{l_1 \cdots l_d}_{m_1 \cdots m_\nu}$ transforms as
\[
    D^{l_1}_{k_1 m_1} \cdots D^{l_d}_{k_d m_d} t^{l_1 \cdots l_d}_{m_1 \cdots m_d} \circ R
    = t^{l_1 \cdots l_d}_{k_1 \cdots k_d},
\]
where $D^l = D^l(R)$ are the Wigner D-matrices, and a symmetrisation analogous
to \eqref{eq:equivariant-overcomplete} may be performed, to enforce this equivariance,

\begin{equation}
    \label{eq:equivariant-overcomplete2}
    {\bf B}_{i, \bf v, {\bf lm}} = \int_{O(3)} \big({\bf D}^{\bf l}_{\bf mk} E^{\bf l}_{\bf k}\big) A_{i \bf v} \circ R \, dR,
\end{equation}
where  ${\bf D}^{\bf l}_{\bf mk} = D^{l_1}_{k_1 m_1} \cdots D^{l_d}_{k_d m_d}$ and
$E^{\bf l}_{\bf k}$ the canonical basis of $d$-dimensional tensors with indices $k_t = -k_t, \dots, l_t$.

\subsection{Equivariant non-linearities}
\label{sec:non_linear_equivariant}

The constraint of equivariance (Eq.~\eqref{eq:equivariant_const}) on non-linearities takes very different forms depending on the details of the group and can only be discussed case-by-case.
A general practice is to observe that the equivariance constraint is always satisfied for invariant messages.
Let $m^{(t)}_{00}$ be the channels of invariant messages.
Then by Eq.~\eqref{eq:equivariant_const} for $U(R) = I, \forall R \in SE(3)$:
\begin{equation}
    m^{(t)}_{00}(R[r_{i_{1}},...,r_{i_{n}}]) = m^{(t)}_{00}(r_{i_{1}},...,r_{i_{n}}), \forall R \in SE(3)
\end{equation}
So the application of any non-linearity $\mathcal{F}$ will be
\begin{equation}
    \label{invariant-non-lin}
    \mathcal{F}(m^{(t)}_{00}(R[r_{i_{1}},...,r_{i_{n}}])) =  \mathcal{F}(m^{(t)}_{00}(r_{i_{1}},...,r_{i_{n}}))
\end{equation}
For general equivariant channels $LM$ of a message, the trick is to use square-norm gated non-linearities~\cite{WeilerGatedNonLinearities2018} of the form
\begin{multline}
    \mathcal{F}\big(m^{(t)}_{LM}(r_{i_{1}},...,r_{i_{n}})\big) \: = \\
    \mathcal{F}\big(\lVert m^{(t)}_{LM}(r_{i_{1}},...,r_{i_{n}})\rVert^{2}\big) m^{(t)}_{LM}(r_{i_{1}},...,r_{i_{n}})
\end{multline}
As the non-linearity is only applied to the squared norm of a feature which is always an invariant scalar, this type of nonlinear functions preserve equivariance.

\subsection{Body-ordering of the SiLU non-linearity}
\label{sec:SiLU_body_order}

Assume that
\begin{equation}
    \mathcal{F}(x) = \text{SiLU}(x) = \frac{x}{1 + e^{-x}} .
\end{equation}
The Taylor expansion of SiLU can can be written as,
\begin{equation}
    \text{SiLU}(x) = \frac{x}{1 + \sum\limits_{k}^{+\infty} (-1)^{k}\frac{x^{k}}{k!}} = \sum_{k=0}^{+\infty} \frac{(-1)^{k}(2^{n} - 1) B_{n}}{n}x^{n}
\end{equation}
where $B_{n}$ corresponds to Bernoulli numbers.
One can immediately notice that
\begin{equation}
    \frac{\partial^{T} \text{SiLU}(x)}{\partial x^{T}} \neq 0 \quad \forall T .
\end{equation}
Thereby formally $T = +\infty$, and the resulting $h^{t+1}_{i}$ will admit infinite body order.

\end{document}